\documentclass[10pt,journal,compsoc]{IEEEtran}

\ifCLASSOPTIONcompsoc

  \usepackage[nocompress]{cite}
\else

  \usepackage{cite}
\fi

\ifCLASSINFOpdf
 
\else
 
\fi

\usepackage{tikz}
\usepackage{array}
\usepackage{times}
\usepackage{dsfont}
\usepackage{subfig}
\usepackage{helvet}
\usepackage{subfig}
\usepackage{amssymb}
\usepackage{amsmath}
\usepackage{amssymb}
\usepackage{courier}
\usepackage{ragged2e}
\usepackage{multirow}
\usepackage{mathrsfs}
\usepackage{graphicx}
\usepackage{enumitem}
\usepackage{graphicx}
\usepackage{blindtext}
\usepackage{algorithm}
\usepackage{algorithmic}
\usepackage{lineno,hyperref}
\usepackage[marginal]{footmisc}
\usepackage{bbm}

\usepackage[utf8]{inputenc}
\usepackage[english]{babel}
\usepackage{epstopdf}
\usepackage{array}
\usepackage{multirow}
\usepackage{booktabs}

\begin{document}

\title{Improved Dual Correlation Reduction Network}
\author{Yue~Liu,~Sihang~Zhou,~Xinwang~Liu,~Wenxuan~Tu,~Xihong~Yang
\IEEEcompsocitemizethanks{
\IEEEcompsocthanksitem Y. Liu, S. Zhou, X. Liu, W. Tu, X. Yang are with National University of Defense Technology, Changsha, 410073, China. (E-mail: yueliu@nudt.edu.cn)

}
}

\markboth{Under Review}%
{Y. Liu \MakeLowercase{\textit{et al.}}: Improved Dual Correlation Reduction Network}

\IEEEtitleabstractindextext{%
\begin{abstract}
\justifying
Deep graph clustering, which aims to reveal the underlying graph structure and divide the nodes into different clusters without human annotations, is a fundamental yet challenging task. However, we observed that the existing methods suffer from the representation collapse problem and easily tend to encode samples with different classes into the same latent embedding. Consequently, the discriminative capability of nodes is limited, resulting in sub-optimal clustering performance. To address this problem, we propose a novel deep graph clustering algorithm termed Improved Dual Correlation Reduction Network (IDCRN) through improving the discriminative capability of samples. Specifically, by approximating the cross-view feature correlation matrix to an identity matrix, we reduce the redundancy between different dimensions of features, thus improving the discriminative capability of the latent space explicitly. Meanwhile, the cross-view sample correlation matrix is forced to approximate the designed clustering-refined adjacency matrix to guide the learned latent representation to recover the affinity matrix even across views, thus enhancing the discriminative capability of features implicitly. Moreover, we avoid the collapsed representation caused by the over-smoothing issue in Graph Convolutional Networks (GCNs) through an introduced propagation regularization term, enabling IDCRN to capture the long-range information with the shallow network structure. Extensive experimental results on six benchmarks have demonstrated the effectiveness and the efficiency of IDCRN compared to the existing state-of-the-art deep graph clustering algorithms.
\end{abstract}

}

\maketitle

\IEEEdisplaynontitleabstractindextext

\IEEEpeerreviewmaketitle

\IEEEraisesectionheading{\section{Introduction}\label{sec:introduction}}
\IEEEPARstart{D}eep graph clustering is a fundamental yet challenging task that aims to reveal the underlying graph structure and divide the nodes into different clusters in a self-supervised manner. GCN \cite{GCN}, which possesses the powerful graph representation learning capability, has achieved promising performance in the attribute graph clustering task. Consequently, based on GCN, many deep graph clustering methods are proposed \cite{DAEGC,ARGA,AGAE,GMM_VGAE,GALA,AGE,O2MAC,SDCN,AGCN,DFCN,DAGC,R-GAE}.

\begin{figure}[!t]
\centering
\begin{minipage}{0.49\linewidth}
\centerline{\includegraphics[width=\textwidth]{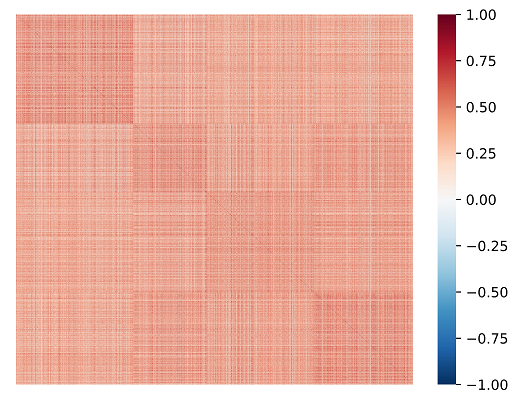}}
\vspace{5pt}
\centerline{(a) GAE}
\vspace{5pt}
\end{minipage}
\begin{minipage}{0.49\linewidth}
\centerline{\includegraphics[width=\textwidth]{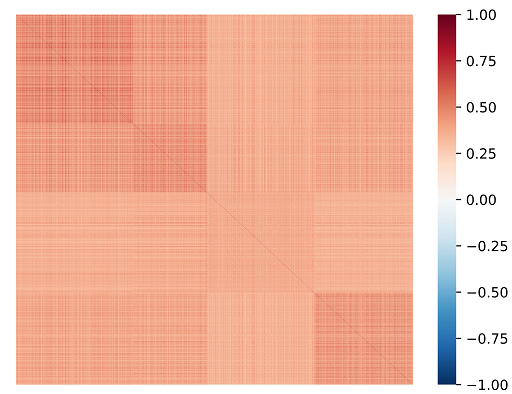}}
\vspace{5pt}
\centerline{(b) MVGRL}
\vspace{5pt}
\end{minipage}
\begin{minipage}{0.49\linewidth}
\centerline{\includegraphics[width=\textwidth]{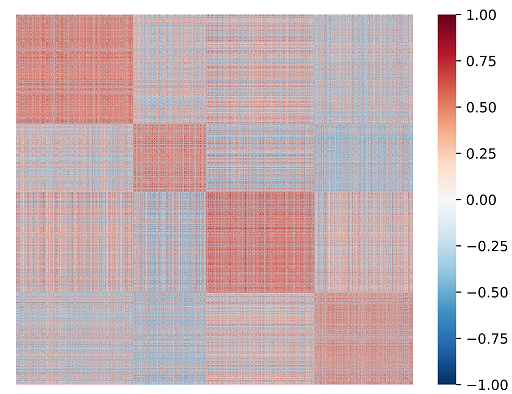}}
\vspace{5pt}
\centerline{(c) SDCN}
\vspace{5pt}
\end{minipage}
\begin{minipage}{0.49\linewidth}
\centerline{\includegraphics[width=\textwidth]{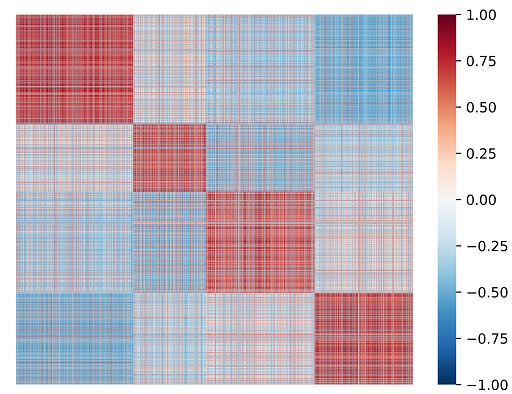}}
\vspace{5pt}
\centerline{(d) IDCRN}
\vspace{5pt}
\end{minipage}
\caption{Visualization of the sample similarity matrices in the latent space learned by the four compared algorithms, i.e., GAE \cite{GAE}, MVGRL \cite{MVGRL}, SDCN \cite{SDCN}, and our proposed method (IDCRN), on the DBLP dataset. The sample order is rearranged to make those from the same cluster beside each other.}
\label{MOTIVATION}
\end{figure}

Although promising performance has been achieved, we observed that the existing deep graph clustering algorithms \cite{GAE,SDCN,MVGRL} suffer from the representation collapse issue \cite{collapse} and easily embed the nodes from different classes into the same embedding. To solve this problem, many attempts have been made. The contrastive-strategy-enhanced method \cite{MVGRL} pushes negative sample pairs away while pulling positive sample pairs together, which to some extent alleviating the collapsed representations. However, since the definition of positive and negative sample pairs is not accurate enough, the learned latent features are still indiscriminative. SDCN \cite{SDCN} exploits both the structural information and attribute information with an information transport operation, which alleviates the over-smoothing problem. However, SDCN \cite{SDCN} neglects the correlation of the latent embeddings, thus leading to sub-optimal clustering performance. We conduct a simple experiment on the DBLP dataset to illustrate this issue. Specifically, in the experiment, we first extract the node embeddings of four well-trained deep graph clustering algorithms, i.e., GAE \cite{GAE}, MVGRL \cite{MVGRL}, SDCN \cite{SDCN}, and our proposed algorithm (IDCRN). Afterward, we calculate the cosine similarity between the embedded samples in the latent space and visualize the resultant similarity matrix for each algorithm. From Fig. \ref{MOTIVATION} (a-c), we observe that although the problem of over-smoothing is alleviated to different extents, the intrinsic four-dimensional cluster structure is not well revealed by the compared three algorithms \cite{GAE,SDCN,MVGRL}. This phenomenon illustrates that the representation collapse problem is still an open problem that limits the clustering performance of the existing deep graph clustering algorithms.

To address the representation collapse problem, we propose a novel contrastive-learning-based deep graph clustering network termed Improved Dual Correlation Reduction Network (IDCRN) by improving the discriminative capability of nodes. In our algorithm, the discriminative capability is improved in two aspects, i.e., the feature aspect and the sample aspect. Specifically, to the feature aspect, we first construct two augmented graph views and encode the nodes with a siamese network. Subsequently, we reduce the redundancy between different dimensions of features via approximating the cross-view feature correlation matrix to the identity matrix. With this setting, the discriminative capability of the latent space is enhanced explicitly, thus alleviating the collapsed representation. Moreover, in the sample aspect, we force the cross-view sample correlation matrix to approximate the high confident clustering results refined affinity matrix. In this manner, we guide the learned latent representation to recover the affinity matrix even across views, thus improving the feature discriminative capability implicitly. As shown in Fig. \ref{MOTIVATION} (d), we found that, with our improved dual correlation reduction module, IDCRN could better reveal the latent cluster structure among data than the other compared algorithms \cite{GAE,MVGRL,SDCN}. Besides, IDCRN saves more GPU space than the contrastive-learning-based algorithms since it eliminates the space-consuming negative sample generation operation. For instance, our proposed method saves about 54\% GPU memory on average compared to MVGRL \cite{MVGRL} during training on DBLP, CITE, and ACM datasets. Moreover, motivated by the Propagation-regularization (P-reg) \cite{p-reg}, we avoid the collapsed representations caused by over-smoothing in GCN by a propagation-regularization term, thus further improving the clustering performance of the proposed method.

This work is an extended version of our AAAI 2022 conference paper, i.e., dual correlation reduction network (DCRN) \cite{DCRN}. In our conference paper, both the sample-level and feature-level cross-view correlation matrices are forced to approximate the identity matrices. However, through our observation and experimental validation, we found that, to improve the discriminative capability, instead of just guiding the network to tell apart samples across views, guiding the network to reveal the underlying sample distribution would endow it with better discriminative capability. To this end, we approximate the cross-view sample correlation matrix to a designed clustering-refined affinity matrix instead of the identity matrix. In this manner, the discriminative capability of latent features wound be enhanced, thus further improving the clustering performance. Take the clustering results on DBLP dataset as an example, our proposed algorithm improves 5.21\% on the metric of ARI compared to DCRN \cite{DCRN}. Moreover, more sufficient experimental studies are conducted to demonstrate the effectiveness and efficiency of the proposed algorithm. Three contributions of this work are listed as follows.

\begin{itemize}
\item A novel contrastive-learning-based method termed IDCRN is proposed to solve the representation collapse problem in the existing deep graph clustering methods.

\item We propose two strategies to enhance the discriminative capability of samples implicitly and explicitly. Besides, compared to other contrastive-learning-based methods, IDCRN saves more GPU memory since it gets rid of the complicated negative sample generation operation.

\item More sufficient experimental results on six benchmarks have verified the superiority of IDCRN compared to the existing state-of-the-art deep graph clustering methods.

\end{itemize}

\section{Related Works}
\subsection{Deep Graph Clustering}
Graph Convolutional Networks (GCNs), which possess the powerful graph representation learning capability, have achieved impressive performance in the field of deep graph clustering. Specifically, the authors of GAE / VGAE \cite{GAE} firstly design a graph encoder to learn the node embeddings from both the attributes and the structure information and then reconstruct the adjacency matrix by an inner product decoder. Based on GAE / VGAE, recent researches, including DAEGC \cite{DAEGC} and GALA \cite{GALA} improve the clustering performance with the attention mechanism and the Laplacian sharpening, respectively. Moreover, other two methods termed ARGA / ARVGA \cite{ARGA} and AGAE \cite{AGAE} enhance the discriminative capability of samples by generative adversarial learning, thus achieving the promising clustering performance. Though ahead mentioned works have improved the clustering performance of early works, the over-smoothing problem is still not solved. SDCN / $\text{SDCN}_\text{Q}$ \cite{SDCN} and DFCN \cite{DFCN} are proposed to jointly train an AE \cite{AE_K_MEANS} and a GAE \cite{GAE} to avoid the over-smoothing issue through the designed information transport operation and the attribute-structure fusion strategy, respectively. 
Similar to DFCN \cite{DFCN}, AGCN \cite{AGCN} also demonstrates the effectiveness of the attribute-structure fusion module. More recently, the contrastive-learning-based methods including MCGC \cite{MCGC} and MVGRL \cite{MVGRL} aim to learn consensus node embeddings from different views of the graph by introducing the contrastive loss, thus further improving the clustering performance. Although recent works have proved the effectiveness of learning consensus information from different views of the graph data, we found that they suffer from the representation collapse problem and easily map samples from different classes into the same embedding, leading to the unsatisfied clustering performance. To avoid the collapsed representation, we propose a novel contrastive-learning-based deep graph clustering method by improving the discriminative capability of the learned embeddings in the sample and feature aspects.

\subsection{Representation Collapse}
Representation collapse is a common problem that the network tends to encode samples from different classes into the same representation in the field of the self-supervised learning \cite{collapse}. Contrastive learning \cite{SSL} is one possible way to solve this problem. Specifically, a pioneer termed MoCo \cite{MOCO} adopts the momentum encoder to keep the consistency of the negative pair embeddings from the designed memory bank. After that, another effective contrastive learning method termed SimCLR \cite{SIMCLR}, defines the ``negative'' and ``positive'' sample pairs and then pushes the ``negative'' samples away while pulling closer the ``positive'' samples. Subsequently, BYOL \cite{BYOL} introduces three strategies, including momentum encoder, predictor, and gradient stopping to address this problem. Different from them, a scalable online clustering method named SwAV \cite{SWAV} alleviates the collapsed representations by mapping the representations into different clusters. Moreover, SimSiam \cite{SIMSAIM} has demonstrated that the stop-gradient mechanism is crucial to alleviate the collapsed representations without negative samples. More recently, Barlow Twins \cite{BARLOW} and VICREG \cite{VICReg} propose the effective yet simple redundancy reduction mechanisms to alleviate the representation collapse issue even without the momentum encoder, the negative sample pairs, or the gradient stopping mechanism. Motivated by their success \cite{BARLOW,VICReg}, DCRN \cite{DCRN} is designed to solve the representation collapse problem through reducing the correlation in the sample and feature aspects. Following DCRN \cite{DCRN}, we further enhance the discriminative capability of learned node embeddings by guiding the network to reveal the underlying sample distribution, thus further improving the clustering performance.

\section{Proposed Method}
\begin{figure*}[!t]
\centering
\includegraphics[scale=0.54]{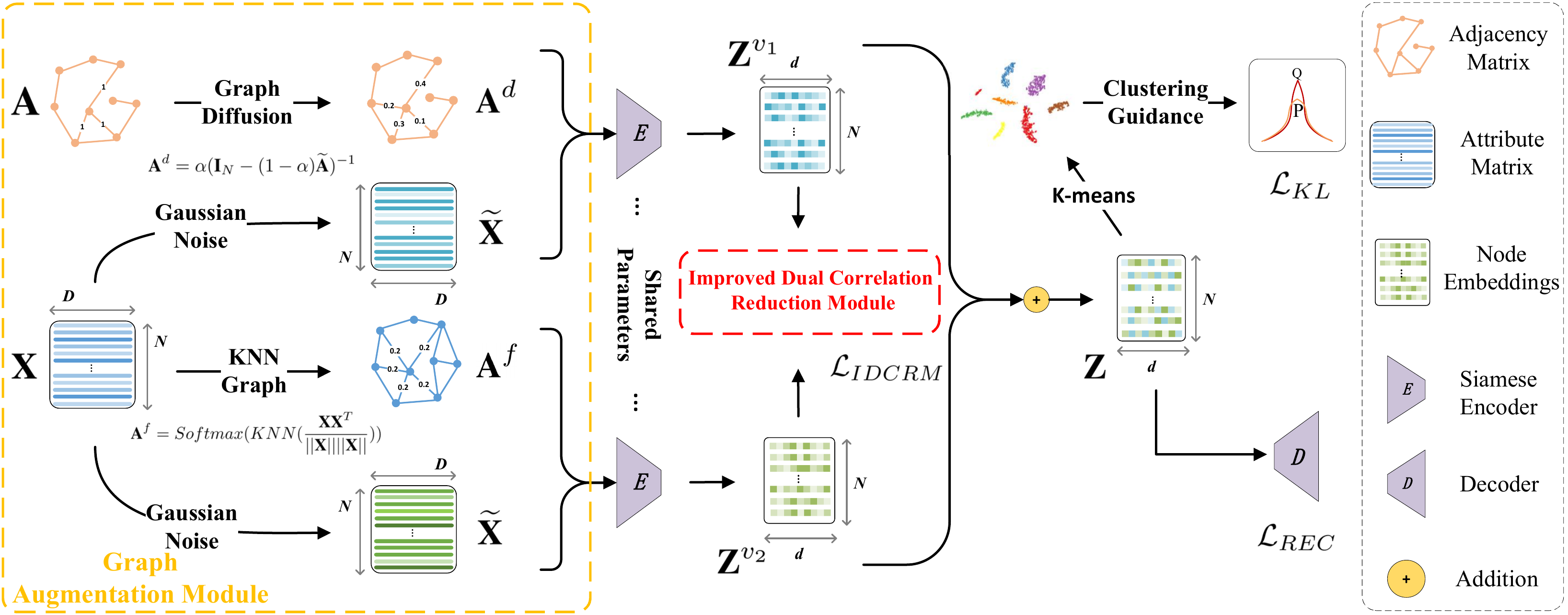} 
\caption{Illustration of the training process of the proposed IDCRN. In the graph augmentation module, two different minor Gaussian perturbations are added to the attribute matrix to generate two forms of the same matrix. The graph structure is strengthened with two manners, i.e., graph diffusion and KNN graph construction based on similarity to improve the graph quality. After the data augmentation, the generated attribute and graph structure pairs are embedded with a siamese network into the latent space. Then, by reducing the feature redundancy and correcting the embedded sample distribution with the improved dual correlation reduction module (IDCRM) in Fig. \ref{IDCRM}, we improve the discriminative capability of the network. Finally, the two embeddings are merged to perform sample reconstruction and K-means clustering \cite{K-means}, which is guided by the widely-used distribution alignment loss \cite{DEC,IDEC,DAEGC,SDCN,DFCN}.}
 \label{OVERRALL_FIGURE}  
\end{figure*}

In this section, we propose a novel contrastive-learning-based graph clustering method termed Improved Dual Correlation Reduction Network (IDCRN) to solve the representation collapse issue by improving the discriminative capability of the node embeddings. Fig. \ref{OVERRALL_FIGURE} illustrates that IDCRN mainly contains two components including graph augmentation module, and Improved Dual Correlation Reduction Module (IDCRM). In what follows, we first define the notations and formulate the problem. Subsequently, we will detail graph augmentation module, IDCRM, and the overall objective function.

\subsection{Notations Definition and Problem Formulation}

Let $\mathcal{V}=\{v_1, v_2, \dots, v_N\}$ be a set of $N$ nodes with $C$ classes and $\mathcal{E}$ be a set of edges. In the matrix form, $\textbf{X} \in \mathds{R}^{N\times D}$ and $\textbf{A} \in \mathds{R}^{N\times N}$ denote the node attribute matrix and the original adjacency matrix, respectively. Then $\mathcal{G}=\left \{\textbf{X}, \textbf{A} \right \}$ denotes an undirected graph. The degree matrix is formulated as $\textbf{D}=diag(d_1, d_2, \dots ,d_N)\in \mathds{R}^{N\times N}$ and $d_i=\sum_{(v_i,v_j)\in \mathcal{E}}a_{ij}$. The normalized adjacency matrix $\widetilde{\textbf{A}} \in \mathds{R}^{N\times N}$ is calculated by $\hat{\textbf{D}}^{-\frac{1}{2}}\hat{\textbf{A}}\hat{\textbf{D}}^{-\frac{1}{2}}$, where $\hat{\textbf{A}} \in \mathds{R}^{N \times N}$ denotes the self-looped adjacency matrix and $\hat{\textbf{D}}$ is the degree matrix of $\hat{\textbf{A}}$. Besides, $||\cdot||$ denotes the square norm. The notations are summarized in Table \ref{NOTATION_TABLE}.

In this paper, we aim to group the nodes into several disjoint groups in the unsupervised manner. To be specific, we first embed the nodes into the latent space without labels and then directly performance clustering algorithm K-means\cite{K-means} over the learned embeddings.

\begin{table}[!t]
\centering
\large
\scalebox{0.87}{
\begin{tabular}{ll}
\toprule
\textbf{Notations}                                        & \textbf{Meanings}                                \\ \midrule
$\textbf{X}\in \mathds{R}^{N\times D}$           & The attribute matrix                       \\
$\textbf{A}\in \mathds{R}^{N\times N}$           & The original adjacency matrix              \\
$\hat{\textbf{A}}\in \mathds{R}^{N\times N}$     & The self-looped adjacency matrix    \\
$\widetilde{\textbf{A}}\in \mathds{R}^{N\times N}$   & The normalized adjacency matrix            \\
$\textbf{A}^f\in \mathds{R}^{N\times N}$   &
The KNN graph adjacency matrix            \\
$\textbf{A}^d\in \mathds{R}^{N\times N}$   & 
The graph diffusion matrix            \\
$\textbf{Z}^{v_k} \in \mathds{R}^{N\times d}$   & The node embedding in $k$-th view      \\
$\widetilde{\textbf{Z}}^{v_k} \in \mathds{R}^{d\times K}$  & The cluster-level embedding in $k$-th view       \\
$\textbf{Z} \in \mathds{R}^{N\times d}$         & The clustering-oriented node embedding      \\
$\textbf{S}^\mathcal{N} \in \mathds{R}^{N\times N}$        & The cross-view sample correlation matrix  \\
$\textbf{S}^\mathcal{F} \in \mathds{R}^{d\times d}$        & The cross-view feature correlation matrix  \\
$\textbf{T} \in \mathds{R}^{N\times N}$          & The clustering-refined affinity matrix                    \\ 
$\textbf{Q} \in \mathds{R}^{N\times C}$          & The soft assignment distribution           \\
$\textbf{P} \in \mathds{R}^{N\times C}$          & The target distribution                    \\ 
\bottomrule
\end{tabular}
}
\caption{Notation summary table}
\label{NOTATION_TABLE} 
\end{table}

\subsection{Graph Augmentation Module}
In the field of self-supervised graph representation learning, several works \cite{MVGRL, aug_2, aug_3} have demonstrated that, the network would learn richer node representations from different augmented graphs. Motivated by their success, we adopt the augmentations on graphs to improve the discriminative capability of node representations. As illustrated the graph augmentation module in Fig. \ref{OVERRALL_FIGURE}, two types of augmentations are considered in our proposed method, i.e., feature perturbation and structure construction.

\subsubsection{Feature Perturbation}
First, we utilize an attribute-level augmentation, i.e., feature perturbation, which disturbs the attribute of nodes in the graph. To be specific, we firstly generate the random noise matrix $\textbf{N} \in \mathds{R}^{N \times D}$, which is sampled from a Gaussian distribution $\mathcal{N}(1, 0.1)$. Subsequently, we calculate the perturbated attribute matrix $\widetilde{\textbf{X}} \in \mathds{R}^{N \times D}$ as formulated:

\begin{equation} 
\widetilde{\textbf{X}}=\textbf{X} \odot \textbf{N},
\label{SCALING}
\end{equation}
where $\odot$ denotes the Hadamard product \cite{hadamard}.

\subsubsection{Structure Construction}
Besides, for the structure-level augmentations, we consider two structure construction strategies. The first strategy is the KNN graph adjacency matrix construction based on the feature similarity \cite{KNN_graph}. Specifically, we calculate the cosine similarity between nodes in the graph and then generate the KNN graph adjacency matrix $\textbf{A}^{f} \in \mathds{R}^{N \times D}$ by the K-Nearest Neighbors algorithm (KNN) \cite{KNN} as formulated:
\begin{equation}
\textbf{A}^{f} = Softmax(KNN(\frac{\textbf{X}\textbf{X}^T}{||\textbf{X}||||\textbf{X}||})),
\label{KNN}
\end{equation}
where $KNN(\cdot)$ denotes the KNN algorithm \cite{KNN} and $Softmax(\cdot)$ denotes the softmax function \cite{SOFTMAX} for normalization. For the number of nearest neighbors $\epsilon$ in the KNN algorithm \cite{KNN}, we set it to 5 in our model. To another strategy, we adopt the generalized graph diffusion utilized in MVGRL \cite{MVGRL}, which capture the local and global information of a graph structure. The generalized graph diffusion matrix $\textbf{G} \in \mathds{R}^{N \times N}$ could be formulated as:
\begin{equation}
\textbf{G} = \sum_{k=0}^{\infty}\theta_k\textbf{T}^k,
\label{DIFFUSION}
\end{equation}
where $\theta_k \in [0,1]$ is the coefficient $k$-order structure information and $\sum_{k=0}^{\infty} \theta_k = 1$. Besides, $\textbf{T} \in \mathds{R}^{N \times N}$ denotes the generalized transition matrix. Actually, we adopt a special case of the generalized graph diffusion, i.e., Personalized PageRank (PPR) algorithm \cite{PAGERANK}, which sets $\textbf{T} = \widetilde{\textbf{A}}$ and $\theta_k = \alpha(1-\alpha)^k$. Thus, the graph diffusion matrix could be formulated by the closed-form solution to PPR as follow:
\begin{equation}
\textbf{A}^{d}=\alpha(\textbf{I}_N-(1-\alpha)\widetilde{\textbf{A}})^{-1},
\label{PPR}
\end{equation}
where $\alpha$ is the teleport (or restart) probability and $\textbf{I}_N \in \mathds{R}^{N \times N}$ denotes an identity matrix.

Based on the feature perturbation and the structure construction, two augmented graphs $\mathcal{G}^{1}=\{\widetilde{\textbf{X}},\textbf{A}^f\}$ and $\mathcal{G}^{2}=\{\widetilde{\textbf{X}},\textbf{A}^d\}$ are generated. In what follows, we aim to guide our network to learn the more discriminative embeddings from two augmented views of the graph.

\subsection{Improved Dual Correlation Reduction Module} \label{sec:IDCRM}
In this section, we propose Improved Dual Correlation Reduction Module (IDCRM) to improve the discriminative capability of the node embeddings in two aspects, i.e., the sample aspect and the feature aspect, thus avoiding the representation collapse problem. Following the above ideas, two strategies termed Affinity Recovery Strategy (ARS) and Redundancy Reduction Strategy (RRS) are designed in IDCRM as shown in Fig. \ref{IDCRM}.

\subsubsection{Affinity Recovery Strategy} To the sample aspect, we design Affinity Recovery Strategy (ARS) to enhance the feature discriminative capability implicitly. Specifically, the proposed ARS contains the following three steps. 

First, we encode two graphs $\mathcal{G}^{1}$ and $\mathcal{G}^{2}$, which are generated by the graph augmentation module, into two-view node embeddings $\textbf{Z}^{v_1}$ and $\textbf{Z}^{v_2}$ with a siamese graph encoder \cite{DFCN}. 

Second, the cross-view sample correlation matrix $\textbf{S}^{\mathcal{N}} \in \mathds{R}^{N \times N}$, whose elements comprised between -1 and 1, could be formulated as: 
\begin{equation}
\textbf{S}_{ij}^{\mathcal{N}}= \frac{\left(\textbf{Z}^{v_1}_i\right) (\textbf{Z}^{v_2}_j)^{\text{T}}}{||\textbf{Z}^{v_1}_i|| || \textbf{Z}^{v_2}_j ||}, \,\,  \forall\,\,i, j \in [1, N],
\label{SAMPLE_CROSS}
\end{equation}
where the element $\textbf{S}_{ij}^{\mathcal{N}}$ is the cosine similarity between $\textbf{Z}^{v_1}_i$ and $\textbf{Z}^{v_2}_j$. Besides, $\textbf{Z}^{v_1}_i$ and $\textbf{Z}^{v_2}_j$ denote $i$-th node embedding of the first view and $j$-th node embedding of the second view, respectively. 

\begin{figure}[!t]
\centering
\includegraphics[scale=0.63]{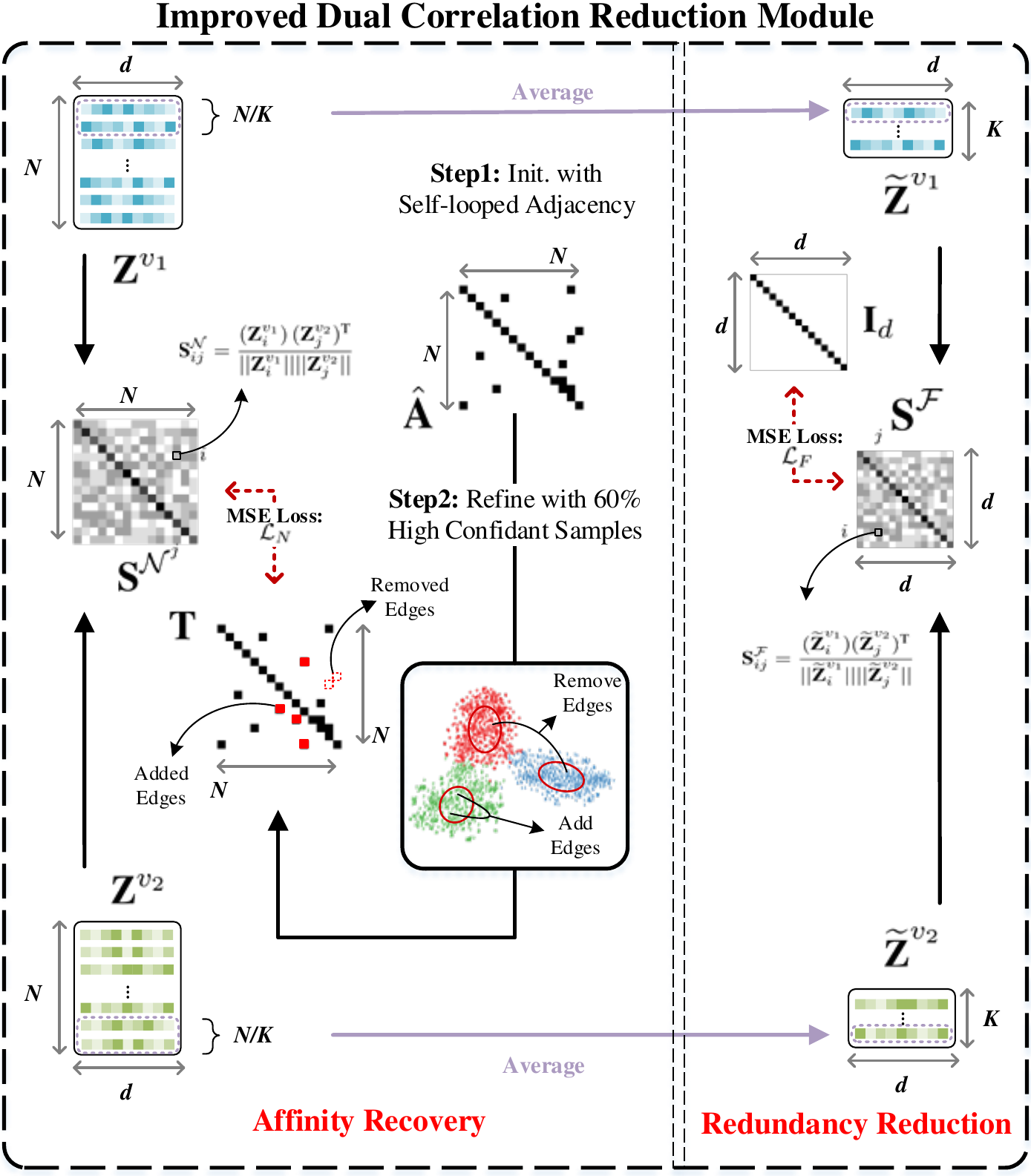} 
\caption{Illustration of the Improved Dual Correlation Reduction Module (IDCRM). Our proposed IDCRM aims to improve the discriminative capability of the embeddings in two aspects, i.e., the sample aspect and the feature aspect. Specifically, to the feature aspect, we reduce the redundancy between different dimensions of features via approximating the cross-view feature correlation matrix to the identity matrix, thus enhancing the discriminative capability of the latent space explicitly. Moreover, in the sample aspect, we force the cross-view sample correlation matrix to approximate the high confident clustering results refined affinity matrix. With this setting, we guide the learned latent representation to recover the affinity matrix even across views, thus improving the feature discriminative capability implicitly.}
\label{IDCRM}  
\end{figure}

Subsequently, we force the cross-view sample correlation matrix $\textbf{S}^{\mathcal{N}}$ to approximate the clustering-refined affinity matrix $\textbf{T}\in \mathds{R}^{N\times N}$ as formulated: 
\begin{equation}
\begin{aligned}
\mathcal{L}_N &= \frac{1}{N^2}\sum (\textbf{S}^{\mathcal{N}}-\textbf{T})^2 \\
&= \frac{1}{N^2}(\underbrace{\sum_i\sum_j\mathbbm{1}_{ij}^{1}(\textbf{S}^{\mathcal{N}}_{ij}-1)^2}_{homogeneity}+\underbrace{\sum_i\sum_j\mathbbm{1}_{ij}^{0}(\textbf{S}^{\mathcal{N}}_{ij})^2}_{heterogeneity}),
\end{aligned}
\label{SAMPLE_LOSS}
\end{equation}
where $\mathbbm{1}_{ij}^{1}$ denotes if $\textbf{T}_{ij}$ is equal to 1 while $\mathbbm{1}_{ij}^{0}$ denotes if $\textbf{T}_{ij}$ is equal to 0. We design $\textbf{T}$ in two steps as shown in the left part of Fig. \ref{IDCRM}. (1) Considering the homogeneity principle \cite{HOMO}, which indicates that nodes from the same class tend to form edges, we initialize $\textbf{T}$ with $\hat{\textbf{A}}$, i.e., the self-looped adjacency matrix. (2) $\textbf{T}$ is refined with the 60\% high confident clustering resultant samples. To be specific, we construct pseudo labels for these samples based on the cluster-ID and further add / remove edges when the paired samples have the same / different pseudo labels. It's worth mentioning that the samples, which are closer to the corresponding cluster centers, have higher confidence in K-means clustering algorithm \cite{K-means}. In this manner, the proposed clustering-refined affinity matrix $\textbf{T}$ could better reveal the homogeneity between the nodes from the same categories and the heterogeneity between the nodes from the different categories.

In Eq. \eqref{SAMPLE_LOSS}, the \textit{homogeneity} term pulls together the nodes from the same category across two views. Differently, the \textit{heterogeneity} term pushes away the samples from different categories across two views. In this manner, our proposed ARS guides the learned representation to recover the affinity matrix even across views, thus improving the feature discriminative capability implicitly.


\subsubsection{Redundancy Reduction Strategy}
In addition to the sample aspect, we further consider the feature aspect to reduce the redundancy between different dimensions of the latent features. Guided by this idea, anther effective strategy termed Redundancy Reduction Strategy (RRS) is designed as illustrated in the right part of Fig. \ref{IDCRM}. To be specific, our proposed RRS contains the following three steps.

We first utilize a readout function $\mathcal{R}(\cdot) : \mathds{R}^{d \times N} \to \mathds{R}^{d \times K}$ to obtain the cluster-level embeddings $\widetilde{\textbf{Z}}^{v_1}, \widetilde{\textbf{Z}}^{v_2} \in \mathds{R}^{d \times K}$ from the node embeddings $\textbf{Z}^{v_1}, \textbf{Z}^{v_2}$. Here, to the readout function $\mathcal{R}$, we first divide the samples into $K$ groups and then output the average value of each group.

Second, similar to Eq. \eqref{SAMPLE_CROSS}, we calculate the cross-view feature correlation matrix $\textbf{S}^{\mathcal{F}} \in \mathds{R}^{N \times N}$ as formulated:
\begin{equation}
\textbf{S}_{ij}^{\mathcal{F}}= \frac{(\widetilde{\textbf{Z}}^{v_1}_i) (\widetilde{\textbf{Z}}^{v_2}_j)^\text{T}}{||\widetilde{\textbf{Z}}^{v_1}_i|| ||\widetilde{\textbf{Z}}^{v_2}_j||} , \,\,  \forall\,\,i, j \in [1, d],
\label{FEATURE_CROSS}
\end{equation}
where $\textbf{S}_{ij}^{\mathcal{F}}$ actually denotes the cosine similarity between $i$-th dimension feature in the first view and $j$-th dimension feature in the second view. 

Subsequently, different from Eq. \eqref{SAMPLE_LOSS}, we force the cross-view feature correlation matrix $\textbf{S}^{\mathcal{F}}$ to approximate an identity matrix $\textbf{I}_d \in \mathds{R}^{d \times d}$ as formulated:
\begin{equation}
\begin{aligned}
\mathcal{L}_F &= \frac{1}{{d}^2}\sum (\textbf{S}^{\mathcal{F}}-\textbf{I}_d)^2 \\ &= \frac{1}{d^2}\sum\limits_{i=1}^{d} \left(\textbf{S}^{\mathcal{F}}_{ii}-1\right)^2
+
\frac{1}{d^2-d}\sum\limits_{i=1}^{d} \sum\limits_{j\ne i} \left(\textbf{S}_{ij}^{\mathcal{F}}\right)^2,
\end{aligned}
\label{FEATURE_LOSS}
\end{equation}
where $d$ denotes the dimension of learned embedding. 

We analyze Eq. \eqref{FEATURE_LOSS} that the first term indicates that the same dimensions of the learned features from two augmented views are enforced to agree with each other. On the contrary, the second term decorrelates the different dimensions of the latent representations. By this way, the redundant information in the learned features is reduced and then the discriminative capability of features is enhanced explicitly, thus avoiding the representation collapse problem.

During training process of the network, we adopt a propagation regularization \cite{p-reg} to alleviate over-smoothing as formulated:
\begin{equation}
\mathcal{L}_{R} = JSD(\textbf{Z}, \ \widetilde{\textbf{A}}\textbf{Z}), 
\label{P-REG}
\end{equation}
where $JSD(\cdot)$ is the Jensen-Shannon divergence \cite{JS}. In this manner, IDCRN is enabled to capture long-range information with the shallow network.

\subsubsection{Fusion and Clustering}
Under the constraints of ARS and RRS, we combination the two views of the node embeddings in a linear manner as formulated:
\begin{equation}
\textbf{Z} = \frac{1}{2}(\textbf{Z}^{v_1}+\textbf{Z}^{v_2}),
\label{FUSION}
\end{equation}
where $\textbf{Z} \in \mathds{R}^{N \times d}$ denotes the resultant clustering-oriented node embeddings. Subsequently, we directly perform K-means algorithm \cite{K-means} over $\textbf{Z}$ and obtain the clustering results.

In summary, the loss function of the proposed IDCRM could be formulated as follows:
\begin{equation}
\mathcal{L}_{IDCRM} = \mathcal{L}_N + \mathcal{L}_F + \gamma \mathcal{L}_{R}, 
\label{JOINT_IDCRM}
\end{equation}
where $\gamma$ is a trade-off hyper-parameter. Technically, in the proposed IDCRM, we consider to enhance the discriminative capability of the node embeddings from both the sample and feature perspective. Under the constraint of IDCRM, our network is guided to reveal the underlying sample distribution and meanwhile the redundancy in the learned features could be filtered out. In this manner, our model would learn more discriminative embeddings to avoid the collapsed representation and further improve the clustering performance.

\subsection{Overall Objective Function}
The overall objective function of IDCRN contains three parts, i.e., the reconstruction loss, the clustering loss, and the loss of IDCRM as follows:  
\begin{equation}
\mathcal{L} = \mathcal{L}_{IDCRM}+\mathcal{L}_{REC}+ \lambda \mathcal{L}_{KL}, 
\label{LOSS}
\end{equation}
where $\mathcal{L}_{REC}$ is the MSE reconstruction loss adopted in \cite{DFCN}. And $\mathcal{L}_{KL}$ denotes the KL divergence \cite{KL}, i.e., a widely-used clustering loss in \cite{DEC,IDEC,DAEGC,SDCN,DFCN}. Here, we first generate a soft assignment distribution $\textbf{Q} \in \mathds{R}^{N \times C}$ and a target distribution $\textbf{P} \in \mathds{R}^{N \times C}$ over the node embeddings $\textbf{Z}$. And then we align them by $\mathcal{L}_{KL}$ to guide the network. The detailed procedure of IDCRN is shown in Algorithm \ref{ALGORITHM}.

\begin{algorithm}[!t]
\small
\caption{IDCRN}
\label{ALGORITHM}
\textbf{Input}: An undircted graph:
$\mathcal{G}=\{\textbf{X},\textbf{A}\}$; The cluster number $C$; Iteration number $t$; Hyper-parameters $\gamma$ and $\lambda$. \\ \textbf{Output}: The clustering result \textbf{O}.
\begin{algorithmic}[1]

\STATE Utilize the proposed graph augmentation module to generate two augmented graph views $\mathcal{G}^{1}=\{\widetilde{\textbf{X}},\textbf{A}^f\}$ and $\mathcal{G}^{2}=\{\widetilde{\textbf{X}},\textbf{A}^d\}$;

\STATE Pre-train the feature extraction encoder to obtain $\textbf{Z}$;
\STATE Initialize the cluster centers by performing K-means over $\textbf{Z}$;

\FOR{$i=1$ to $t$}
\STATE Utilize the feature extraction encoder to obtain $\textbf{Z}^{v_1}$ and $\textbf{Z}^{v_2}$;
\STATE Obtain $\widetilde{\textbf{Z}}^{v_1}$ and $\widetilde{\textbf{Z}}^{v_2}$ by the readout function $\mathcal{R}$;
\STATE Calculate $\textbf{S}^{\mathcal{N}}$ and $\textbf{S}^{\mathcal{F}}$ by Eq. \eqref{SAMPLE_CROSS} and Eq. \eqref{FEATURE_CROSS}, respectively;
\STATE Conduct the Affinity Recovery Strategy and Redundancy Reduction Strategy by Eq. \ref{SAMPLE_LOSS} and Eq. \ref{FEATURE_LOSS}, respectively;
\STATE Obtain $\textbf{Z}$ by fusing $\textbf{Z}^{v_1}$ and $\textbf{Z}^{v_2}$ in Eq. \eqref{FUSION};
\STATE Calculate \textit{$\mathcal{L}_{IDCRM}$}, \textit{$\mathcal{L}_{REC}$}, and \textit{$\mathcal{L}_{KL}$}, respectively.
\STATE{Optimize the whole network by minimizing $\mathcal{L}$ in Eq. \eqref{LOSS}};
\ENDFOR
\STATE{Obtain \textbf{O} by performing K-means over $\textbf{Z}$.}
\STATE \textbf{return} \textbf{O}
\end{algorithmic}
\end{algorithm}

\section{Experiments}

\subsection{Datasets}
To verify the effectiveness and efficiency of IDCRN, abundant experimental studies are conducted on six graph clustering benchmarks, including ACM\cite{SDCN}, CITE\cite{SDCN}, DBLP\cite{SDCN}, AMAP \cite{AMAP}, PUBMED\cite{PUBMED}, and CORAFULL\cite{CORAFULL}. We list the statistics of these datasets in Table \ref{DATASET_INFO} and the detailed descriptions are summarized as follows:

\begin{itemize}
\item \textbf{ACM} \cite{SDCN}:
It is a network of the papers. An edge will be constructed between two papers if they are written by the same author. The features of the papers are the bag-of-words of the keywords. The papers published in MobiCOMM, SIGCOMM, SIGMOD, KDD are selected and divided into three classes, including data mining, wireless communication, and database.

\item \textbf{CITE} \cite{SDCN}:
This citation network consists of a set of citation links between different documents whose feature vectors are the sparse bag-of-words. The labels are divided into the six areas including HCI, machine language, information retrieval, database, artificial intelligence, and agents.

\item \textbf{DBLP} \cite{SDCN}: 
This author network contains authors from four areas including information retrieval, machine learning, data mining, and database. The edge in constructed between two authors if they are the co-author relationship. The features of the authors are the bag-of-words of keywords.

\item \textbf{AMAP} \cite{AMAP}:
This is a co-purchase graph from Amazon. The nodes in the graph denote the products and the features are the reviews encoded by the bag-of-words. The edges indicate whether two products are frequently co-purchased or not. The nodes are divided into eight classes.

\item \textbf{PUBMED} \cite{PUBMED}:
This is a citation network, which contains scientific publications from the PubMed database. The nodes are divided into three classes and links indicates the citation between different publications. The publications in the graph are described by a TF/IDF weighted word vector from a dictionary which consists of 500 unique words.

\item \textbf{CORAFULL} \cite{CORAFULL}:
The is a citation network consists of 19793 scientific publications classified into one of seventy classes. This citation network includes 65311 links.

\end{itemize}

\subsection{Experiment Setup}
\subsubsection{Training Procedure}
The deep learning platform and the GPU of all experiments are PyTorch and an NVIDIA 3090. The training process of our network consists of three steps. Following DFCN \cite{DFCN}, we independently pre-train the sub-networks for 30 epochs by minimizing the reconstruction loss $\mathcal{L}_{REC}$. Afterward, we obtain the initial clustering centers by integrating two sub-networks into a united framework and training another 100 epochs. Then we fine-tune our whole network with 400 epochs until convergence by minimizing the loss calculated in Eq. \ref{LOSS}. Consequently, we perform clustering on the embeddings $\textbf{Z}$ by K-means algorithm \cite{K-means}. In the compare experiments, we conduct ten runs for all methods and report the average values with standard deviations of four metrics to alleviate the random seed influence.

\subsubsection{Parameters Setting}
To ARGA / ARVGA \cite{ARGA}, MVGRL \cite{MVGRL}, and DFCN \cite{DFCN}, we reproduce the average values with standard deviations as results by adopting the corresponding source code with the original literature setting. To MCGC \cite{MCGC}, for fairness, we merely adopt and run their source code on the graph datasets in Table \ref{DATASET_INFO}. For other baselines, we list the corresponding results reported in DFCN \cite{DFCN}. In our proposed method, we utilize DFCN \cite{DFCN} as our feature extractor. Besides, our network is optimized with the Adam optimizer \cite{ADAM}. The learning rate of our IDCRN is set to 5e-5 for ACM, 1e-3 for AMAP, 1e-4 for DBLP, 1e-5 for CITE, PUBMED, and CORAFULL, respectively. The teleport probability $\alpha$ in the Personalized PageRank (PPR) algorithm \cite{PAGERANK} is set as 0.1 for PUBMED, 0.3 for ACM, and 0.2 for other datasets. Afterward, $\epsilon$ in KNN algorithm \cite{KNN} and $K$ in readout function $\mathcal{R}(\cdot)$ are fixed as 5 and the number of clusters $C$. For the trade-off hyper-parameters $\lambda$ and $\gamma$, we respectively set them as 10 and 1e3.

\begin{table}[!t]
\centering
\setlength{\tabcolsep}{6mm}{
\begin{tabular}{@{}cccc@{}}
\toprule
\textbf{Dataset}  & \textbf{Samples} & \textbf{Dimensions}  & \textbf{Classes} \\ \midrule
\textbf{DBLP}     & 4057    & 334        & 4       \\
\textbf{CITE}     & 3327    & 3703       & 6       \\
\textbf{ACM}      & 3025    & 1870       & 3       \\
\textbf{AMAP}     & 7650    & 745        & 8       \\
\textbf{PUBMED}   & 19717   & 500        & 3       \\
\textbf{CORAFULL} & 19793   & 8710       & 70      \\ \bottomrule
\end{tabular}
}
\caption{Dataset summary}
\label{DATASET_INFO} 
\end{table}

\subsubsection{Metrics} 
To comprehensively verify the superiority of the compared methods, the clustering performance is evaluated by four metrics \cite{suyuan_1,ZHOU_1,ZHOU_2,tiejian_1,siwei_1}, i.e., ACC, NMI, ARI, and F1. In more detail, ACC could be calculated as follow:

\begin{equation}
\text{ACC}=\frac{\sum_{i=1}^n\phi(l_i,map(c_i))}{n}, 
\label{ACC}
\end{equation}
where $c_i$ and $l_i$ respectively denote the predicted cluster ID and the label for the $i$-th sample. The$\phi(\cdot)$ is an indicator function as formulated:
\begin{equation}
\phi(l_i, map(c_i)) = 
\left \{
\begin{aligned}
\ 1 \ \ & if \ l_i=map(c_i), \\
\ 0 \ \ & otherwise.
\end{aligned}
\right .
\end{equation}
The best map from the predicted cluster ID $c_i$ to class ID could be constructed by the Kuhn-Munkres algorithm \cite{Kuhn-Munkres}, i.e., the $map(\cdot)$ function.

Another critical metric macro F1-score, which indicates a balance of precision and recall, cloud be calculated as formulated: 
\begin{equation}
\text{F1}=2 \cdot \frac{\text{P} \cdot \text{R}}{\text{P} + \text{R}}.
\label{F1}
\end{equation}
In detail, $\text{P}=\text{TP}/(\text{TP}+\text{FP})$ is the precision value and $\text{R}=\text{TP}/(\text{TP}+\text{FN})$ is the recall value. TP, FP, and FN denotes True Positive error, False Positive error, and False Negative error, respectively.

A mutual information score based metric named NMI is widely used in clustering tasks since it is robust to the unbalanced label distribution. It is defined as:
\begin{equation}
\text{NMI}=-\frac{2\sum_x\sum_y p(x,y) log \frac{p(x,y)}{p(x)p(y)}}{\sum_i p(x_i) log(p(x_i))+\sum_j p(y_j) log(p(y_j))},
\label{NMI}
\end{equation}
where $x,y$ denote the distribution of the predicted results and the ground truth, respectively.

Different from NMI, ARI is based on the similarity of pairwise labels between the ground truth and predicted results as formulated:
\begin{equation}
\text{ARI} = \frac{\overbrace{\sum_{i j}\tbinom{n_{i j}}{2}}^{\text {Index}} - \overbrace{\left[\sum_{i} \tbinom{a_{i}}{2}\sum_{j} \tbinom{b_{j}}{2}\right] / \tbinom{n}{2}}^{\text{Expected index}}}{\underbrace{\frac{1}{2}\left[\sum_{i} \tbinom{a_{i}}{2}+\sum_{j} \tbinom{b_{j}}{2} \right]}_{\text {Max index }}-\underbrace{\left[ \sum_{i} \tbinom{a_{i}}{2} \sum_{j} \tbinom{b_{j}}{2} \right] / \tbinom{n}{2}}_{\text{Expected index}}},
\end{equation}
where $n$ is the number of all pairs, $a$ is the number of pairs with the same cluster, and $b$ is the number of pairs with different clusters.

\begin{table*}[]
\centering
\large
\scalebox{0.43}{
\begin{tabular}{c|c|cccccccccccccc|cc}
\hline
{\color[HTML]{000000} }                                    & {\color[HTML]{000000} }                                  & {\color[HTML]{000000} \textbf{K-Means}} & {\color[HTML]{000000} \textbf{AE}}                   & {\color[HTML]{000000} \textbf{DEC}} & {\color[HTML]{000000} \textbf{IDEC}} & {\color[HTML]{000000} \textbf{GAE}} & {\color[HTML]{000000} \textbf{VGAE}} & {\color[HTML]{000000} \textbf{DAEGC}} & {\color[HTML]{000000} \textbf{ARGA}} & {\color[HTML]{000000} \textbf{ARVGA}} & {\color[HTML]{000000} \textbf{$\text{SDCN}_\text{Q}$}} & {\color[HTML]{000000} \textbf{SDCN}} & {\color[HTML]{000000} \textbf{MVGRL}} & {\color[HTML]{000000} \textbf{MCGC}} & {\color[HTML]{000000} \textbf{DFCN}} & {\color[HTML]{000000} \textbf{DCRN}} & {\color[HTML]{000000} \textbf{IDCRN}} \\ \cline{3-18} 
\multirow{-2}{*}{{\color[HTML]{000000} \textbf{Dataset}}}  & \multirow{-2}{*}{{\color[HTML]{000000} \textbf{Metric}}} & {\color[HTML]{000000} \cite{K-means}}                & {\color[HTML]{000000} \cite{AE_K_MEANS}}                             & {\color[HTML]{000000} \cite{DEC}}            & {\color[HTML]{000000} \cite{IDEC}}             & {\color[HTML]{000000} \cite{GAE}} & {\color[HTML]{000000} \cite{GAE}}                           & {\color[HTML]{000000} \cite{DAEGC}}              &{\color[HTML]{000000} \cite{ARGA}}            & {\color[HTML]{000000} \cite{ARGA}}                      & {\color[HTML]{000000} \cite{SDCN}}       & {\color[HTML]{000000} \cite{SDCN}}                           & {\color[HTML]{000000} \cite{MVGRL}}              & {\color[HTML]{000000} \cite{MCGC}}            & {\color[HTML]{000000} \cite{DFCN}}            & \multicolumn{2}{c}{{\color[HTML]{000000} \textbf{Our Proposed Methods}}}            \\ \hline
{\color[HTML]{000000} }                                    & {\color[HTML]{000000} ACC}                               & {\color[HTML]{000000} 38.65±0.65}       & {\color[HTML]{000000} 51.43±0.35}                    & {\color[HTML]{000000} 58.16±0.56}   & {\color[HTML]{000000} 60.31±0.62}    & {\color[HTML]{000000} 61.21±1.22}   & {\color[HTML]{000000} 58.59±0.06}    & {\color[HTML]{000000} 62.05±0.48}     & {\color[HTML]{000000} 64.83±0.59}    & {\color[HTML]{000000} 54.41±0.42}     & {\color[HTML]{000000} 65.74±1.34}       & {\color[HTML]{000000} 68.05±1.81}    & {\color[HTML]{000000} 42.73±1.02}     & {\color[HTML]{000000} 58.92±0.05}    & {\color[HTML]{000000} 76.00±0.80}    & {\color[HTML]{3531FF} 79.66±0.25}    & {\color[HTML]{FE0000} 82.08±0.18}    \\
{\color[HTML]{000000} }                                    & {\color[HTML]{000000} NMI}                               & {\color[HTML]{000000} 11.45±0.38}       & {\color[HTML]{000000} 25.40±0.16}                    & {\color[HTML]{000000} 29.51±0.28}   & {\color[HTML]{000000} 31.17±0.50}    & {\color[HTML]{000000} 30.80±0.91}   & {\color[HTML]{000000} 26.92±0.06}    & {\color[HTML]{000000} 32.49±0.45}     & {\color[HTML]{000000} 29.42±0.92}    & {\color[HTML]{000000} 25.90±0.33}     & {\color[HTML]{000000} 35.11±1.05}       & {\color[HTML]{000000} 39.50±1.34}    & {\color[HTML]{000000} 15.41±0.63}     & {\color[HTML]{000000} 33.69±0.06}    & {\color[HTML]{000000} 43.70±1.00}    & {\color[HTML]{3531FF} 48.95±0.44}    & {\color[HTML]{FE0000} 52.70±0.36}    \\
{\color[HTML]{000000} }                                    & {\color[HTML]{000000} ARI}                               & {\color[HTML]{000000} 6.97±0.39}        & {\color[HTML]{000000} 12.21±0.43}                    & {\color[HTML]{000000} 23.92±0.39}   & {\color[HTML]{000000} 25.37±0.60}    & {\color[HTML]{000000} 22.02±1.40}   & {\color[HTML]{000000} 17.92±0.07}    & {\color[HTML]{000000} 21.03±0.52}     & {\color[HTML]{000000} 27.99±0.91}    & {\color[HTML]{000000} 19.81±0.42}     & {\color[HTML]{000000} 34.00±1.76}       & {\color[HTML]{000000} 39.15±2.01}    & {\color[HTML]{000000} 8.22±0.50}      & {\color[HTML]{000000} 25.97±0.21}    & {\color[HTML]{000000} 47.00±1.50}    & {\color[HTML]{3531FF} 53.60±0.46}    & {\color[HTML]{FE0000} 58.81±0.37}    \\
\multirow{-4}{*}{{\color[HTML]{000000} \textbf{DBLP}}}     & {\color[HTML]{000000} F1}                                & {\color[HTML]{000000} 31.92±0.27}       & {\color[HTML]{000000} 52.53±0.36}                    & {\color[HTML]{000000} 59.38±0.51}   & {\color[HTML]{000000} 61.33±0.56}    & {\color[HTML]{000000} 61.41±2.23}   & {\color[HTML]{000000} 58.69±0.07}    & {\color[HTML]{000000} 61.75±0.67}     & {\color[HTML]{000000} 64.97±0.66}    & {\color[HTML]{000000} 55.37±0.40}     & {\color[HTML]{000000} 65.78±1.22}       & {\color[HTML]{000000} 67.71±1.51}    & {\color[HTML]{000000} 40.52±1.51}     & {\color[HTML]{000000} 50.39±0.09}    & {\color[HTML]{000000} 75.70±0.80}    & {\color[HTML]{3531FF} 79.28±0.26}    & {\color[HTML]{FE0000} 81.47±0.20}    \\ \hline
{\color[HTML]{000000} }                                    & {\color[HTML]{000000} ACC}                               & {\color[HTML]{000000} 39.32±3.17}       & {\color[HTML]{000000} 57.08±0.13}                    & {\color[HTML]{000000} 55.89±0.20}   & {\color[HTML]{000000} 60.49±1.42}    & {\color[HTML]{000000} 61.35±0.80}   & {\color[HTML]{000000} 60.97±0.36}    & {\color[HTML]{000000} 64.54±1.39}     & {\color[HTML]{000000} 61.07±0.49}    & {\color[HTML]{000000} 59.31±1.38}     & {\color[HTML]{000000} 61.67±1.05}       & {\color[HTML]{000000} 65.96±0.31}    & {\color[HTML]{000000} 68.66±0.36}     & {\color[HTML]{000000} 64.76±0.07}    & {\color[HTML]{000000} 69.50±0.20}    & {\color[HTML]{3531FF} 70.86±0.18}    & {\color[HTML]{FE0000} 71.40±0.08}    \\
{\color[HTML]{000000} }                                    & {\color[HTML]{000000} NMI}                               & {\color[HTML]{000000} 16.94±3.22}       & {\color[HTML]{000000} 27.64±0.08}                    & {\color[HTML]{000000} 28.34±0.30}   & {\color[HTML]{000000} 27.17±2.40}    & {\color[HTML]{000000} 34.63±0.65}   & {\color[HTML]{000000} 32.69±0.27}    & {\color[HTML]{000000} 36.41±0.86}     & {\color[HTML]{000000} 34.40±0.71}    & {\color[HTML]{000000} 31.80±0.81}     & {\color[HTML]{000000} 34.39±1.22}       & {\color[HTML]{000000} 38.71±0.32}    & {\color[HTML]{000000} 43.66±0.40}     & {\color[HTML]{000000} 39.11±0.06}    & {\color[HTML]{000000} 43.90±0.20}    & {\color[HTML]{3531FF} 45.86±0.35}    & {\color[HTML]{FE0000} 46.77±0.21}    \\
{\color[HTML]{000000} }                                    & {\color[HTML]{000000} ARI}                               & {\color[HTML]{000000} 13.43±3.02}       & {\color[HTML]{000000} 29.31±0.14}                    & {\color[HTML]{000000} 28.12±0.36}   & {\color[HTML]{000000} 25.70±2.65}    & {\color[HTML]{000000} 33.55±1.18}   & {\color[HTML]{000000} 33.13±0.53}    & {\color[HTML]{000000} 37.78±1.24}     & {\color[HTML]{000000} 34.32±0.70}    & {\color[HTML]{000000} 31.28±1.22}     & {\color[HTML]{000000} 35.50±1.49}       & {\color[HTML]{000000} 40.17±0.43}    & {\color[HTML]{000000} 44.27±0.73}     & {\color[HTML]{000000} 37.54±0.12}    & {\color[HTML]{000000} 45.50±0.30}    & {\color[HTML]{3531FF} 47.64±0.30}    & {\color[HTML]{FE0000} 48.67±0.20}    \\
\multirow{-4}{*}{{\color[HTML]{000000} \textbf{CITE}}}     & {\color[HTML]{000000} F1}                                & {\color[HTML]{000000} 36.08±3.53}       & {\color[HTML]{000000} 53.80±0.11}                    & {\color[HTML]{000000} 52.62±0.17}   & {\color[HTML]{000000} 61.62±1.39}    & {\color[HTML]{000000} 57.36±0.82}   & {\color[HTML]{000000} 57.70±0.49}    & {\color[HTML]{000000} 62.20±1.32}     & {\color[HTML]{000000} 58.23±0.31}    & {\color[HTML]{000000} 56.05±1.13}     & {\color[HTML]{000000} 57.82±0.98}       & {\color[HTML]{000000} 63.62±0.24}    & {\color[HTML]{000000} 63.71±0.39}     & {\color[HTML]{000000} 59.64±0.05}    & {\color[HTML]{000000} 64.30±0.20}    & {\color[HTML]{3531FF} 65.83±0.21}    & {\color[HTML]{FE0000} 66.27±0.21}    \\ \hline
{\color[HTML]{000000} }                                    & {\color[HTML]{000000} ACC}                               & {\color[HTML]{000000} 67.31±0.71}       & {\color[HTML]{000000} 81.83±0.08}                    & {\color[HTML]{000000} 84.33±0.76}   & {\color[HTML]{000000} 85.12±0.52}    & {\color[HTML]{000000} 84.52±1.44}   & {\color[HTML]{000000} 84.13±0.22}    & {\color[HTML]{000000} 86.94±2.83}     & {\color[HTML]{000000} 86.29±0.36}    & {\color[HTML]{000000} 83.89±0.54}     & {\color[HTML]{000000} 86.95±0.08}       & {\color[HTML]{000000} 90.45±0.18}    & {\color[HTML]{000000} 86.73±0.76}     & {\color[HTML]{000000} 91.64±0.00}    & {\color[HTML]{000000} 90.90±0.20}    & {\color[HTML]{3531FF} 91.93±0.20}    & {\color[HTML]{FE0000} 92.58±0.08}    \\
{\color[HTML]{000000} }                                    & {\color[HTML]{000000} NMI}                               & {\color[HTML]{000000} 32.44±0.46}       & {\color[HTML]{000000} 49.30±0.16}                    & {\color[HTML]{000000} 54.54±1.51}   & {\color[HTML]{000000} 56.61±1.16}    & {\color[HTML]{000000} 55.38±1.92}   & {\color[HTML]{000000} 53.20±0.52}    & {\color[HTML]{000000} 56.18±4.15}     & {\color[HTML]{000000} 56.21±0.82}    & {\color[HTML]{000000} 51.88±1.04}     & {\color[HTML]{000000} 58.90±0.17}       & {\color[HTML]{000000} 68.31±0.25}    & {\color[HTML]{000000} 60.87±1.40}     & {\color[HTML]{000000} 70.71±0.00}    & {\color[HTML]{000000} 69.40±0.40}    & {\color[HTML]{3531FF} 71.56±0.61}    & {\color[HTML]{FE0000} 73.17±0.32}    \\
{\color[HTML]{000000} }                                    & {\color[HTML]{000000} ARI}                               & {\color[HTML]{000000} 30.60±0.69}       & {\color[HTML]{000000} 54.64±0.16}                    & {\color[HTML]{000000} 60.64±1.87}   & {\color[HTML]{000000} 62.16±1.50}    & {\color[HTML]{000000} 59.46±3.10}   & {\color[HTML]{000000} 57.72±0.67}    & {\color[HTML]{000000} 59.35±3.89}     & {\color[HTML]{000000} 63.37±0.86}    & {\color[HTML]{000000} 57.77±1.17}     & {\color[HTML]{000000} 65.25±0.19}       & {\color[HTML]{000000} 73.91±0.40}    & {\color[HTML]{000000} 65.07±1.76}     & {\color[HTML]{000000} 76.63±0.00}    & {\color[HTML]{000000} 74.90±0.40}    & {\color[HTML]{3531FF} 77.56±0.52}    & {\color[HTML]{FE0000} 79.18±0.22}    \\
\multirow{-4}{*}{{\color[HTML]{000000} \textbf{ACM}}}      & {\color[HTML]{000000} F1}                                & {\color[HTML]{000000} 67.57±0.74}       & {\color[HTML]{000000} 82.01±0.08}                    & {\color[HTML]{000000} 84.51±0.74}   & {\color[HTML]{000000} 85.11±0.48}    & {\color[HTML]{000000} 84.65±1.33}   & {\color[HTML]{000000} 84.17±0.23}    & {\color[HTML]{000000} 87.07±2.79}     & {\color[HTML]{000000} 86.31±0.35}    & {\color[HTML]{000000} 83.87±0.55}     & {\color[HTML]{000000} 86.84±0.09}       & {\color[HTML]{000000} 90.42±0.19}    & {\color[HTML]{000000} 86.85±0.72}     & {\color[HTML]{000000} 91.70±0.00}    & {\color[HTML]{000000} 90.80±0.20}    & {\color[HTML]{3531FF} 91.94±0.20}    & {\color[HTML]{FE0000} 92.60±0.08}    \\ \hline
{\color[HTML]{000000} }                                    & {\color[HTML]{000000} ACC}                               & {\color[HTML]{000000} 27.22±0.76}       & {\color[HTML]{000000} 48.25±0.08}                    & {\color[HTML]{000000} 47.22±0.08}   & {\color[HTML]{000000} 47.62±0.08}    & {\color[HTML]{000000} 71.57±2.48}   & {\color[HTML]{000000} 74.26±3.63}    & {\color[HTML]{000000} 76.44±0.01}     & {\color[HTML]{000000} 69.28±2.30}    & {\color[HTML]{000000} 61.46±2.71}     & {\color[HTML]{000000} 35.53±0.39}       & {\color[HTML]{000000} 53.44±0.81}    & {\color[HTML]{000000} 45.19±1.79}     & {\color[HTML]{000000} 71.64±0.00}           & {\color[HTML]{000000} 76.88±0.80}    & {\color[HTML]{3531FF} 79.94±0.13}    & {\color[HTML]{FE0000} 80.17±0.04}    \\
{\color[HTML]{000000} }                                    & {\color[HTML]{000000} NMI}                               & {\color[HTML]{000000} 13.23±1.33}       & {\color[HTML]{000000} 38.76±0.30}                    & {\color[HTML]{000000} 37.35±0.05}   & {\color[HTML]{000000} 37.83±0.08}    & {\color[HTML]{000000} 62.13±2.79}   & {\color[HTML]{000000} 66.01±3.40}    & {\color[HTML]{000000} 65.57±0.03}     & {\color[HTML]{000000} 58.36±2.76}    & {\color[HTML]{000000} 53.25±1.91}     & {\color[HTML]{000000} 27.90±0.40}       & {\color[HTML]{000000} 44.85±0.83}    & {\color[HTML]{000000} 36.89±1.31}     & {\color[HTML]{000000} 61.54±0.00}           & {\color[HTML]{000000} 69.21±1.00}    & {\color[HTML]{3531FF} 73.70±0.24}    & {\color[HTML]{FE0000} 74.32±0.07}    \\
{\color[HTML]{000000} }                                    & {\color[HTML]{000000} ARI}                               & {\color[HTML]{000000} 5.50±0.44}        & {\color[HTML]{000000} 20.80±0.47}                    & {\color[HTML]{000000} 18.59±0.04}   & {\color[HTML]{000000} 19.24±0.07}    & {\color[HTML]{000000} 48.82±4.57}   & {\color[HTML]{000000} 56.24±4.66}    & {\color[HTML]{000000} 59.39±0.02}     & {\color[HTML]{000000} 44.18±4.41}    & {\color[HTML]{000000} 38.44±4.69}     & {\color[HTML]{000000} 15.27±0.37}       & {\color[HTML]{000000} 31.21±1.23}    & {\color[HTML]{000000} 18.79±0.47}     & {\color[HTML]{000000} 43.23±0.00}           & {\color[HTML]{000000} 58.98±0.84}    & {\color[HTML]{3531FF} 63.69±0.20}    & {\color[HTML]{FE0000} 64.10±0.10}    \\
\multirow{-4}{*}{{\color[HTML]{000000} \textbf{AMAP}}}     & {\color[HTML]{000000} F1}                                & {\color[HTML]{000000} 23.96±0.51}       & {\color[HTML]{000000} 47.87±0.20}                    & {\color[HTML]{000000} 46.71±0.12}   & {\color[HTML]{000000} 47.20±0.11}    & {\color[HTML]{000000} 68.08±1.76}   & {\color[HTML]{000000} 70.38±2.98}    & {\color[HTML]{000000} 69.97±0.02}     & {\color[HTML]{000000} 64.30±1.95}    & {\color[HTML]{000000} 58.5±1.70}      & {\color[HTML]{000000} 34.25±0.44}       & {\color[HTML]{000000} 50.66±1.49}    & {\color[HTML]{000000} 39.65±2.39}     & {\color[HTML]{000000} 68.64±0.00}           & {\color[HTML]{000000} 71.58±0.31}    & {\color[HTML]{3531FF} 73.82±0.12}    & {\color[HTML]{FE0000} 74.01±0.04}    \\ \hline
{\color[HTML]{000000} }                                    & {\color[HTML]{000000} ACC}                               & {\color[HTML]{000000} 59.83±0.01}       & {\color[HTML]{000000} 63.07±0.31}                    & {\color[HTML]{000000} 60.14±0.09}   & {\color[HTML]{000000} 60.70±0.34}    & {\color[HTML]{000000} 62.09±0.81}   & {\color[HTML]{000000} 68.48±0.77}    & {\color[HTML]{000000} 68.73±0.03}     & {\color[HTML]{000000} 65.26±0.12}    & {\color[HTML]{000000} 64.25±1.24}     & {\color[HTML]{000000} 64.39±0.30}       & {\color[HTML]{000000} 64.20±1.30}    & {\color[HTML]{000000} 67.01±0.52}     & {\color[HTML]{000000} 60.97±0.01}    & {\color[HTML]{000000} 68.89±0.07}    & {\color[HTML]{3531FF} 69.87±0.07}    & {\color[HTML]{FE0000} 70.02±0.03}    \\
{\color[HTML]{000000} }                                    & {\color[HTML]{000000} NMI}                               & {\color[HTML]{000000} 31.05±0.02}       & {\color[HTML]{000000} 26.32±0.57}                    & {\color[HTML]{000000} 22.44±0.14}   & {\color[HTML]{000000} 23.67±0.29}    & {\color[HTML]{000000} 23.84±3.54}   & {\color[HTML]{000000} 30.61±1.71}    & {\color[HTML]{000000} 28.26±0.03}     & {\color[HTML]{000000} 24.80±0.17}    & {\color[HTML]{000000} 23.88±1.05}     & {\color[HTML]{000000} 26.67±1.31}       & {\color[HTML]{000000} 22.87±2.04}    & {\color[HTML]{000000} 31.59±1.45}     & {\color[HTML]{FE0000} 33.39±0.02}    & {\color[HTML]{000000} 31.43±0.13}    & {\color[HTML]{000000} 32.20±0.08}    & {\color[HTML]{3531FF} 33.29±0.07}    \\
{\color[HTML]{000000} }                                    & {\color[HTML]{000000} ARI}                               & {\color[HTML]{000000} 51.43±0.35}       & {\color[HTML]{000000} 23.86±0.67}                    & {\color[HTML]{000000} 19.55±0.13}   & {\color[HTML]{000000} 20.58±0.39}    & {\color[HTML]{000000} 20.62±1.39}   & {\color[HTML]{000000} 30.15±1.23}    & {\color[HTML]{000000} 29.84±0.04}     & {\color[HTML]{000000} 24.35±0.17}    & {\color[HTML]{000000} 22.82±1.52}     & {\color[HTML]{000000} 24.61±1.46}       & {\color[HTML]{000000} 22.30±2.07}    & {\color[HTML]{000000} 29.42±1.06}     & {\color[HTML]{000000} 29.25±0.01}    & {\color[HTML]{000000} 30.64±0.11}    & {\color[HTML]{3531FF} 31.41±0.12}    & {\color[HTML]{FE0000} 32.67±0.05}    \\
\multirow{-4}{*}{{\color[HTML]{000000} \textbf{PUBMED}}}   & {\color[HTML]{000000} F1}                                & {\color[HTML]{000000} 58.88±0.01}       & {\color[HTML]{000000} 64.01±0.29}                    & {\color[HTML]{000000} 61.49±0.10}   & {\color[HTML]{000000} 62.41±0.32}    & {\color[HTML]{000000} 61.37±0.85}   & {\color[HTML]{000000} 67.68±0.89}    & {\color[HTML]{000000} 68.23±0.02}     & {\color[HTML]{000000} 65.69±0.13}    & {\color[HTML]{000000} 64.51±1.32}     & {\color[HTML]{000000} 65.46±0.39}       & {\color[HTML]{000000} 65.01±1.21}    & {\color[HTML]{000000} 67.07±0.36}     & {\color[HTML]{000000} 59.84±0.01}    & {\color[HTML]{000000} 68.10±0.07}    & {\color[HTML]{3531FF} 68.94±0.08}    & {\color[HTML]{FE0000} 69.19±0.03}    \\ \hline
{\color[HTML]{000000} }                                    & {\color[HTML]{000000} ACC}                               & {\color[HTML]{000000} 26.27±1.10}       & {\color[HTML]{000000} 33.12±0.19}                    & {\color[HTML]{000000} 31.92±0.45}   & {\color[HTML]{000000} 32.19±0.31}    & {\color[HTML]{000000} 29.60±0.81}   & {\color[HTML]{000000} 32.66±1.29}    & {\color[HTML]{000000} 34.35±1.00}     & {\color[HTML]{000000} 22.07±0.43}    & {\color[HTML]{000000} 29.57±0.59}     & {\color[HTML]{000000} 29.75±0.69}       & {\color[HTML]{000000} 26.67±0.40}    & {\color[HTML]{000000} 31.52±2.95}     & {\color[HTML]{000000} 29.08±0.58}    & {\color[HTML]{000000} 37.51±0.81}    & {\color[HTML]{3531FF} 38.80±0.60}    & {\color[HTML]{FE0000} 39.45±0.50}    \\
{\color[HTML]{000000} }                                    & {\color[HTML]{000000} NMI}                               & {\color[HTML]{000000} 34.68±0.84}       & {\color[HTML]{000000} 41.53±0.25}                    & {\color[HTML]{000000} 41.67±0.24}   & {\color[HTML]{000000} 41.64±0.28}    & {\color[HTML]{000000} 45.82±0.75}   & {\color[HTML]{000000} 47.38±1.59}    & {\color[HTML]{000000} 49.16±0.73}     & {\color[HTML]{000000} 41.28±0.25}    & {\color[HTML]{000000} 48.77±0.44}     & {\color[HTML]{000000} 40.10±0.22}       & {\color[HTML]{000000} 37.38±0.39}    & {\color[HTML]{000000} 48.99±3.95}     & {\color[HTML]{000000} 36.86±0.56}    & {\color[HTML]{000000} 51.30±0.41}    & {\color[HTML]{3531FF} 51.91±0.35}    & {\color[HTML]{FE0000} 52.83±0.39}    \\
{\color[HTML]{000000} }                                    & {\color[HTML]{000000} ARI}                               & {\color[HTML]{000000} 9.35±0.57}        & {\color[HTML]{000000} 18.13±0.27}                    & {\color[HTML]{000000} 16.98±0.29}   & {\color[HTML]{000000} 17.17±0.22}    & {\color[HTML]{000000} 17.84±0.86}   & {\color[HTML]{000000} 20.01±1.38}    & {\color[HTML]{000000} 22.60±0.47}     & {\color[HTML]{000000} 12.38±0.24}    & {\color[HTML]{000000} 18.80±0.57}     & {\color[HTML]{000000} 16.47±0.38}       & {\color[HTML]{000000} 13.63±0.27}    & {\color[HTML]{000000} 19.11±2.63}     & {\color[HTML]{000000} 13.15±0.48}    & {\color[HTML]{000000} 24.46±0.48}    & {\color[HTML]{3531FF} 25.25±0.49}    & {\color[HTML]{FE0000} 25.97±0.54}    \\
\multirow{-4}{*}{{\color[HTML]{000000} \textbf{CORAFULL}}} & {\color[HTML]{000000} F1}                                & {\color[HTML]{000000} 22.57±1.09}       & \multicolumn{1}{l}{{\color[HTML]{000000} 28.4±0.30}} & {\color[HTML]{000000} 27.71±0.58}   & {\color[HTML]{000000} 27.72±0.41}    & {\color[HTML]{000000} 25.95±0.75}   & {\color[HTML]{000000} 29.06±1.15}    & {\color[HTML]{000000} 26.96±1.33}     & {\color[HTML]{000000} 18.85±0.41}    & {\color[HTML]{000000} 25.43±0.62}     & {\color[HTML]{000000} 24.62±0.53}       & {\color[HTML]{000000} 22.14±0.43}    & {\color[HTML]{000000} 26.51±2.87}     & {\color[HTML]{000000} 22.90±0.52}    & {\color[HTML]{000000} 31.22±0.87}    & {\color[HTML]{3531FF} 31.68±0.76}    & {\color[HTML]{FE0000} 32.58±0.72}    \\ \hline
\end{tabular}}
\caption{The clustering performance of 14 state-of-the-art algorithms and our proposed method with mean values ± standard deviations (mean ± std) on six datasets.
The values with {\color[HTML]{FE0000} red} and {\color[HTML]{3531FF} blue} correspond to the best and the runner-up results.}
\label{COMPARE_RESULT}
\end{table*}


\begin{figure*}[!t]
\begin{minipage}{0.139\linewidth}
\centerline{\includegraphics[width=\textwidth]{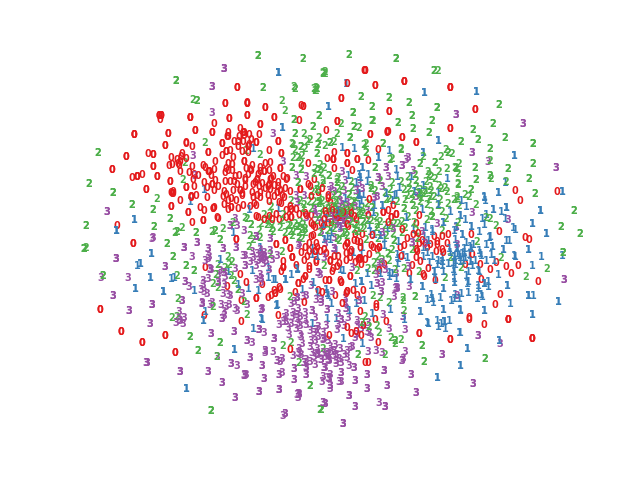}}
\vspace{5pt}
\centerline{\includegraphics[width=\textwidth]{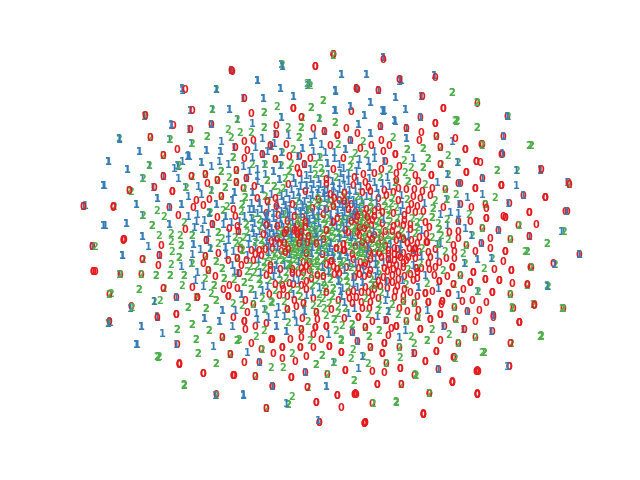}}
\vspace{5pt}
\centerline{(a) Raw Data}
\end{minipage}
\begin{minipage}{0.139\linewidth}
\centerline{\includegraphics[width=\textwidth]{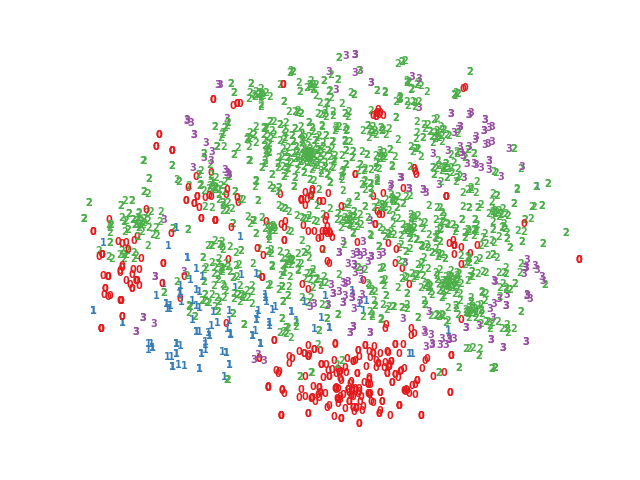}}
\vspace{5pt}
\centerline{\includegraphics[width=\textwidth]{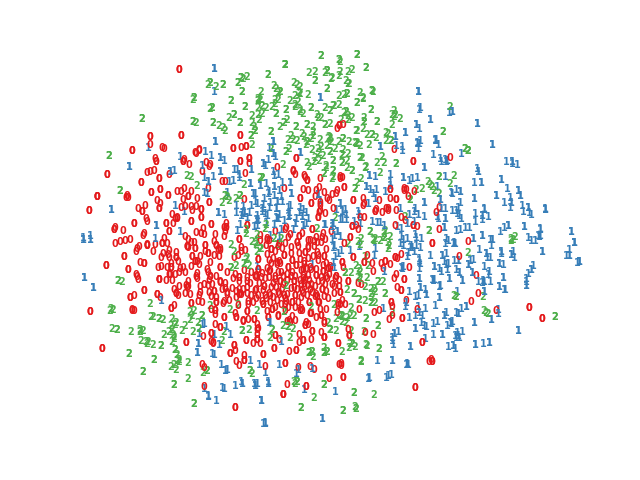}}
\vspace{5pt}
\centerline{(b) AE}
\end{minipage}
\begin{minipage}{0.139\linewidth}
\centerline{\includegraphics[width=\textwidth]{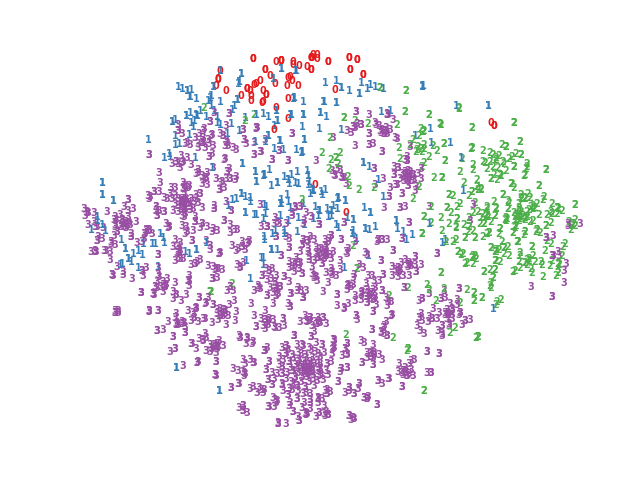}}
\vspace{5pt}
\centerline{\includegraphics[width=\textwidth]{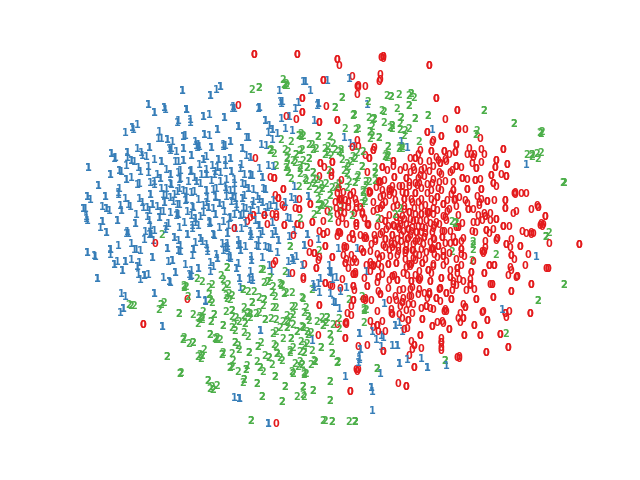}}
\vspace{5pt}
\centerline{(c) DEC}
\end{minipage}
\begin{minipage}{0.139\linewidth}
\centerline{\includegraphics[width=\textwidth]{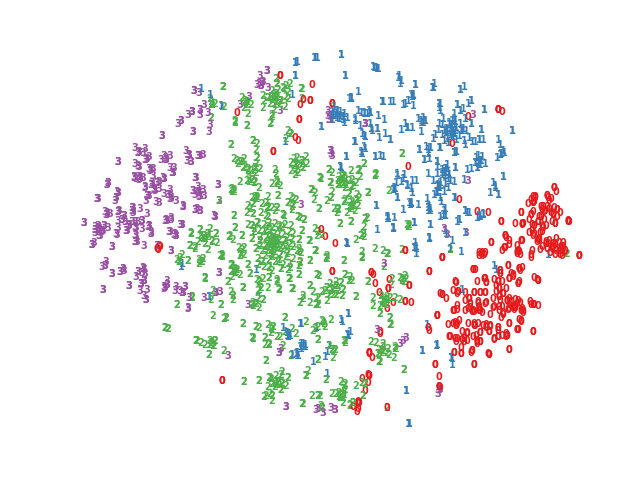}}
\vspace{5pt}
\centerline{\includegraphics[width=\textwidth]{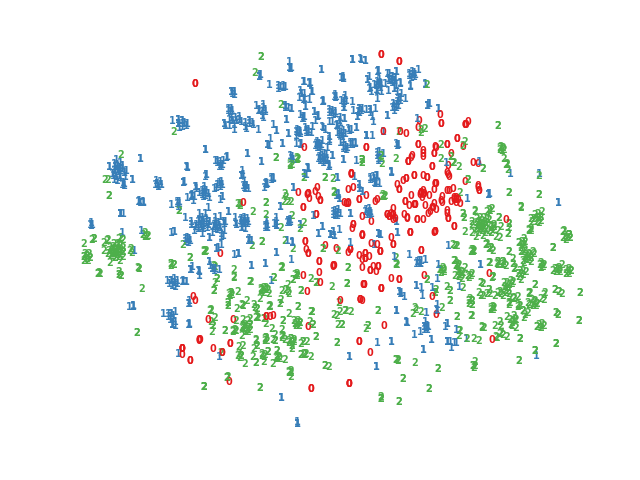}}
\vspace{5pt}
\centerline{(d) GAE}
\end{minipage}
\begin{minipage}{0.139\linewidth}
\centerline{\includegraphics[width=\textwidth]{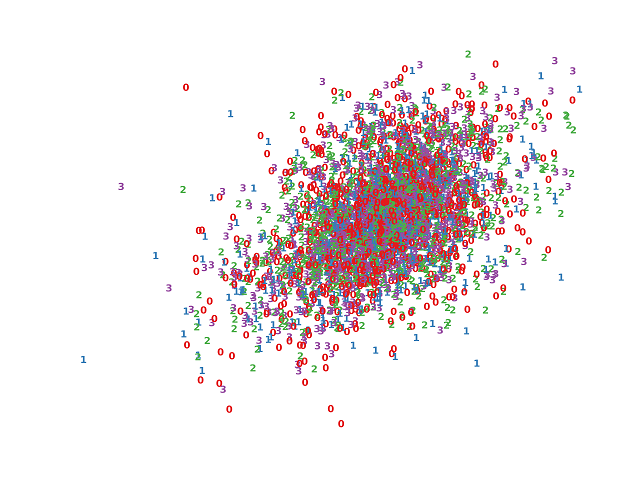}}
\vspace{5pt}
\centerline{\includegraphics[width=\textwidth]{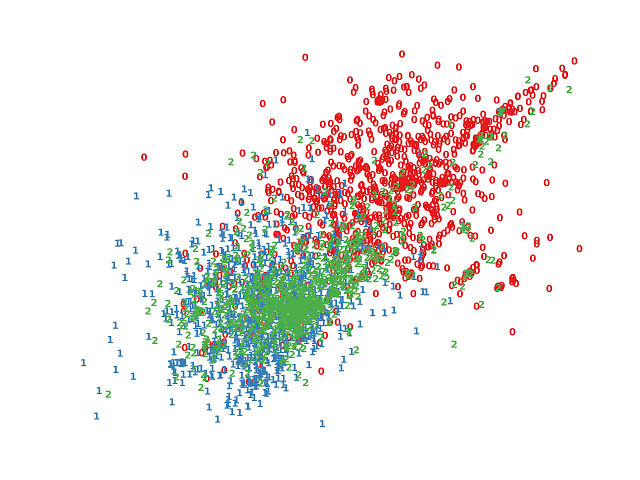}}
\vspace{5pt}
\centerline{(e) ARGA}
\end{minipage}
\begin{minipage}{0.139\linewidth}
\centerline{\includegraphics[width=\textwidth]{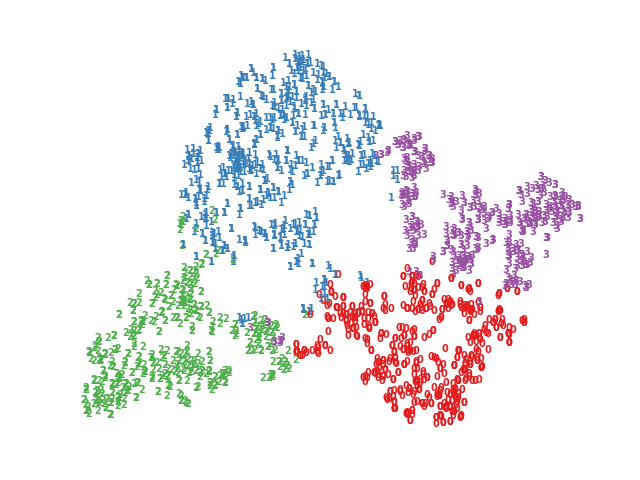}}
\vspace{5pt}
\centerline{\includegraphics[width=\textwidth]{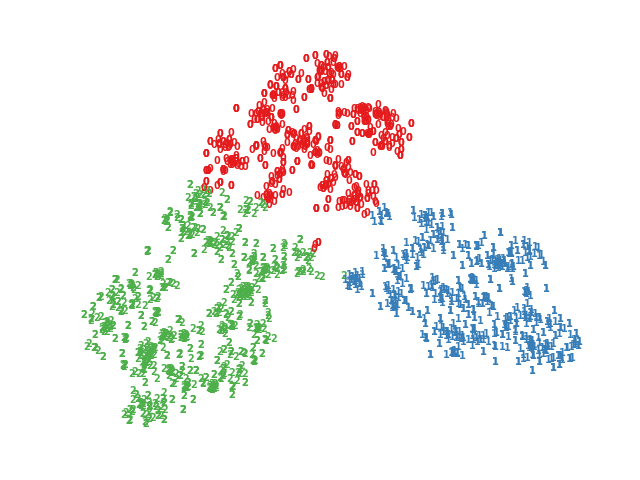}}
\vspace{5pt}
\centerline{(f) DFCN}
\end{minipage}
\begin{minipage}{0.139\linewidth}
\centerline{\includegraphics[width=\textwidth]{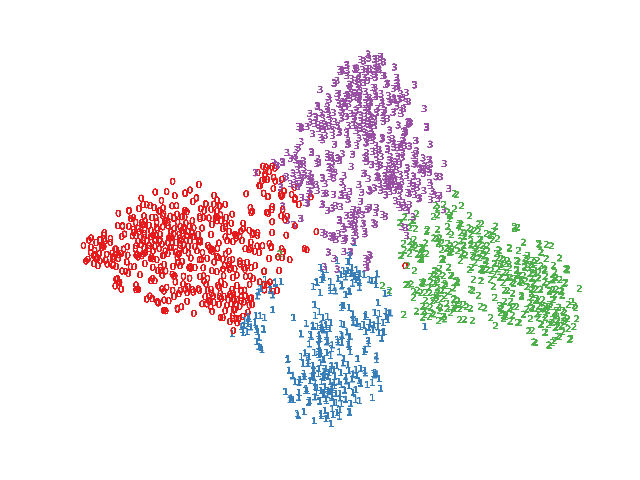}}
\vspace{5pt}
\centerline{\includegraphics[width=\textwidth]{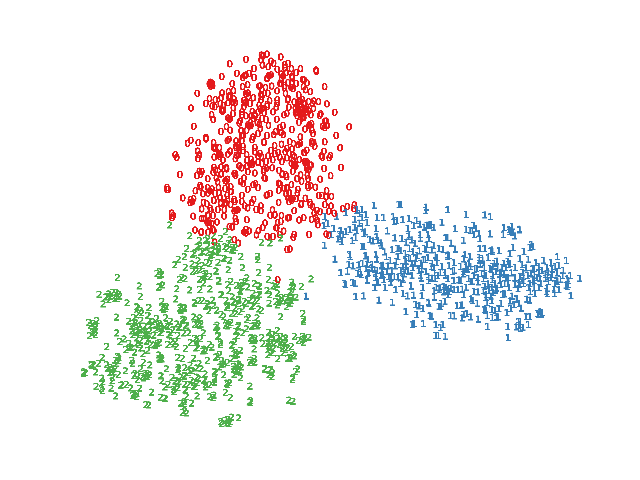}}
\vspace{5pt}
\centerline{(g) IDCRN}
\end{minipage}

\caption{$t$-SNE \cite{T_SNE} visualization of the representation of raw data, AE \cite{AE_K_MEANS}, DEC \cite{DEC}, GAE \cite{GAE}, ARGA \cite{ARGA}, DFCN \cite{DFCN}, and our proposed method 
on two datasets. 
The first and second row indicate the results on ACM and DBLP dataset, respectively.}
\label{VIS}  
\end{figure*}

\subsection{Performance Comparison}
In this section, we conduct comparison experiments of IDCRN and the other 14 baselines to show the superiority of IDCRN. Specifically, K-means algorithm \cite{K-means} is a classic clustering method with the idea of EM algorithm \cite{EM}. Besides, three generative deep clustering methods, including AE \cite{AE_K_MEANS}, DEC \cite{DEC}, and IDEC \cite{IDEC}, first train an auto encoder to embed the samples into the latent space and then perform K-means \cite{K-means} over the learned embeddings. Different from them, three typical GCN-based frameworks, i.e., GAE / VGAE \cite{GAE}, DAEGC \cite{DAEGC}, and ARGA / ARVGA \cite{ARGA} aim to learn representations for clustering by exploiting from both the structure and attribute of the graph. More recently, four state-of-the-art deep graph clustering methods, i.e., SDCN / SDCN$_{Q}$ \cite{SDCN}, DFCN \cite{DFCN}, MVGRL \cite{MVGRL}, and MCGC \cite{MCGC} have achieved the promising clustering performance through learning the consensus representations from different views of the graph.

Table \ref{COMPARE_RESULT} reports the clustering performance of IDCRN and other 14 compared baselines on six benchmarks. Based on these results, we analyze and conclude as follows. 
\begin{itemize}
\item Our proposed IDCRN almost exceeds all other baselines in four metrics on six datasets except the NMI metric on PUBMED dataset. 

\item Specifically, our proposed method achieves better clustering performance than the strongest deep graph clustering frameworks, including SDCN/$\text{SDCN}_Q$ \cite{SDCN}, MVGRL \cite{MVGRL}, MCGC \cite{MCGC}, and DFCN \cite{DFCN}. For instance, IDCRN exceeds DFCN by 6.08\% 9.00\%, 11.81\% 5.77\% increment concerning ACC, NMI, ARI, and F1 metrics on DBLP dataset. The reason is that they all aim to learn latent embeddings from multi-view graph data with redundant information, thus easily suffering from representation collapse. Different from them, by reducing the redundancy and recovering the affinity matrix, IDCRN is guided to learn more discriminative representation, thus avoiding the collapsed representation.

\item Other GCN-based graph clustering methods, including ARGA \cite{ARGA}, DAEGC \cite{DAEGC}, and GAE/VGAE \cite{GAE} achieve unsatisfactory performance compared to ours since these methods fail to exploit different views of the graph.

\item The auto-encoder-based clustering methods, including AE \cite{AE_K_MEANS}, DEC \cite{AE_K_MEANS}, and IDEC \cite{IDEC}, achieve unpromising clustering performance. This verifies that these methods, which are merely based on attribute of the samples, can not learn discriminative features from the graph data. 


\item The classical clustering method K-means algorithm \cite{K-means} achieve unpromising results since it is directly performed on the raw attributes.

\end{itemize}
Overall, the above observations and conclusions have verified that our proposed method effectively alleviates representation collapse and achieves superior clustering performance.

\begin{table}[!t]
\centering
\scalebox{0.89}{
\begin{tabular}{c|c|llll}
\hline
\textbf{Dataset}                   & \textbf{Metric} & \multicolumn{1}{c}{\textbf{B}} & \multicolumn{1}{c}{\textbf{B-P}} & \multicolumn{1}{c}{\textbf{B-I}} & \multicolumn{1}{c}{\textbf{B-P-I}} \\ \hline
\multirow{4}{*}{\textbf{DBLP}}     & ACC    & 76.00±0.80                   & 77.00±0.41                     & 82.04±0.22                     & 82.08±0.18                       \\
                          & NMI    & 43.70±1.00                   & 44.98±0.56                     & 52.72±0.42                     & 52.70±0.36                       \\
                          & ARI    & 47.00±1.50                  & 48.51±0.84                     & 58.74±0.42                     & 58.81±0.37                       \\
                          & F1     & 75.70±0.80                   & 76.77±0.38                     & 81.40±0.24                     & 81.47±0.20                       \\ \hline
\multirow{4}{*}{\textbf{CITE}}         & ACC    & 69.50±0.20                   & 70.07±0.21                     & 71.12±0.14                     & 71.40±0.08                       \\
                          & NMI    & 43.90±0.20                   & 44.75±0.40                     & 46.30±0.35                     & 46.77±0.21                       \\
                          & ARI    & 45.50±0.30                   & 46.52±0.36                     & 48.29±0.32                     & 48.67±0.20                       \\
                          & F1     & 64.30±0.20                   & 65.03±0.23                     & 66.23±0.22                     & 66.27±0.21                       \\ \hline
\multirow{4}{*}{\textbf{ACM}}      & ACC    & 90.90±0.20                   & 91.57±0.12                     & 92.30±0.19                     & 92.58±0.08                       \\
                          & NMI    & 69.40±0.40                   & 70.82±0.25                     & 72.32±0.53                     & 73.17±0.32                       \\
                          & ARI    & 74.90±0.40                   & 76.68±0.28                     & 78.46±0.49                     & 79.18±0.22                       \\
                          & F1     & 90.80±0.20                   & 91.53±0.12                     & 92.31±0.20                     & 92.60±0.08                       \\ \hline
\multirow{4}{*}{\textbf{AMAP}}     & ACC    & 76.88±0.80                   & 79.01±0.01                     & 80.02±0.24                     & 80.17±0.04                       \\
                          & NMI    & 69.21±1.00                   & 72.29±0.01                     & 73.81±0.21                     & 74.32±0.07                       \\
                          & ARI    & 58.98±0.84                   & 62.10±0.01                     & 63.95±0.39                     & 64.10±0.10                       \\
                          & F1     & 71.58±0.31                   & 73.09±0.00                     & 73.92±0.20                     & 74.01±0.04                       \\ \hline
\multirow{4}{*}{\textbf{PUBMED}}   & ACC    & 68.89±0.07                   & 69.43±0.05                     & 69.80±0.06                     & 70.02±0.03                       \\
                          & NMI    & 31.43±0.13                   & 31.98±0.12                     & 32.05±0.06                     & 33.29±0.07                       \\
                          & ARI    & 30.64±0.11                   & 31.35±0.12                     & 31.34±0.11                     & 32.67±0.05                       \\
                          & F1     & 68.10±0.07                   & 68.54±0.06                     & 68.83±0.07                     & 69.19±0.03                       \\ \hline
\multirow{4}{*}{\textbf{CORAFULL}} & ACC    & 37.51±0.81                   & 37.04±0.71                     & 38.45±0.27                     & 39.45±0.50                       \\
                          & NMI    & 51.30±0.41                   & 51.90±0.26                     & 51.04±0.23                     & 52.83±0.39                       \\
                          & ARI    & 24.46±0.48                   & 24.13±0.51                     & 24.96±0.20                     & 25.97±0.54                       \\
                          & F1     & 31.22±0.87                   & 30.35±0.87                     & 31.87±0.75                     & 32.58±0.72                       \\ \hline
\end{tabular}
}

\caption{Ablation study results of IDCRM and the propagated regularization on six benchmarks.}
\label{ICR_ABLATION} 
\end{table}

\subsection{Ablation Studies}
In this section, we conduct ablation studies to verify the effectiveness of our proposed IDCRM and further two proposed strategies including Affinity Recovery Strategy (ARS) and Redundancy Reduction Strategy (RRS) in IDCRM.

\subsubsection{Effectiveness of Improved Dual Correlation Reduction Module}
In order to verify the effectiveness of Improved Dual Correlation Reduction Module (IDCRM) clearly, extensive ablation studies are conducted in Table \ref{ICR_ABLATION}. Here, we adopt DFCN \cite{DFCN} as the baseline. Besides, the baseline with the propagated regularization (P-reg) \cite{p-reg}, the IDCRM, and both of them is denoted as B-P, B-I, and, B-P-I, respectively. From these results, we could observe and conclude as follows.

\begin{itemize}
\item To B-P, P-reg could improve the baseline by about 0.79\% on DBLP dataset. From these results, we conclude that P-reg could some extent alleviate the over-smoothing problem and improve our model's generalization capacity. 

\item Our proposed IDCRM improves the baseline by a large margin. For instance, the baseline with IDCRM, i.e., B-I, exceeds the baseline by 6.04\%, 9.02\%, 11.74\%, 5.70\% performance increment in terms of ACC, NMI, ARI, and F1 on DBLP dataset. Based on these results, we analyze and conclude that the discriminative capacity of the latent features is enhanced by our proposed IDCRM, thus improving the clustering performance.

\item Compared to other variants, B-P-I achieves the best results, verifying the effectiveness of the both components, i.e., IDCRM and P-reg.

\end{itemize}

\subsubsection{Effectiveness of ARS and RRS}
\begin{figure}[!t]
\centering
\small
\begin{minipage}{0.49\linewidth}
\centerline{\includegraphics[width=0.9\textwidth]{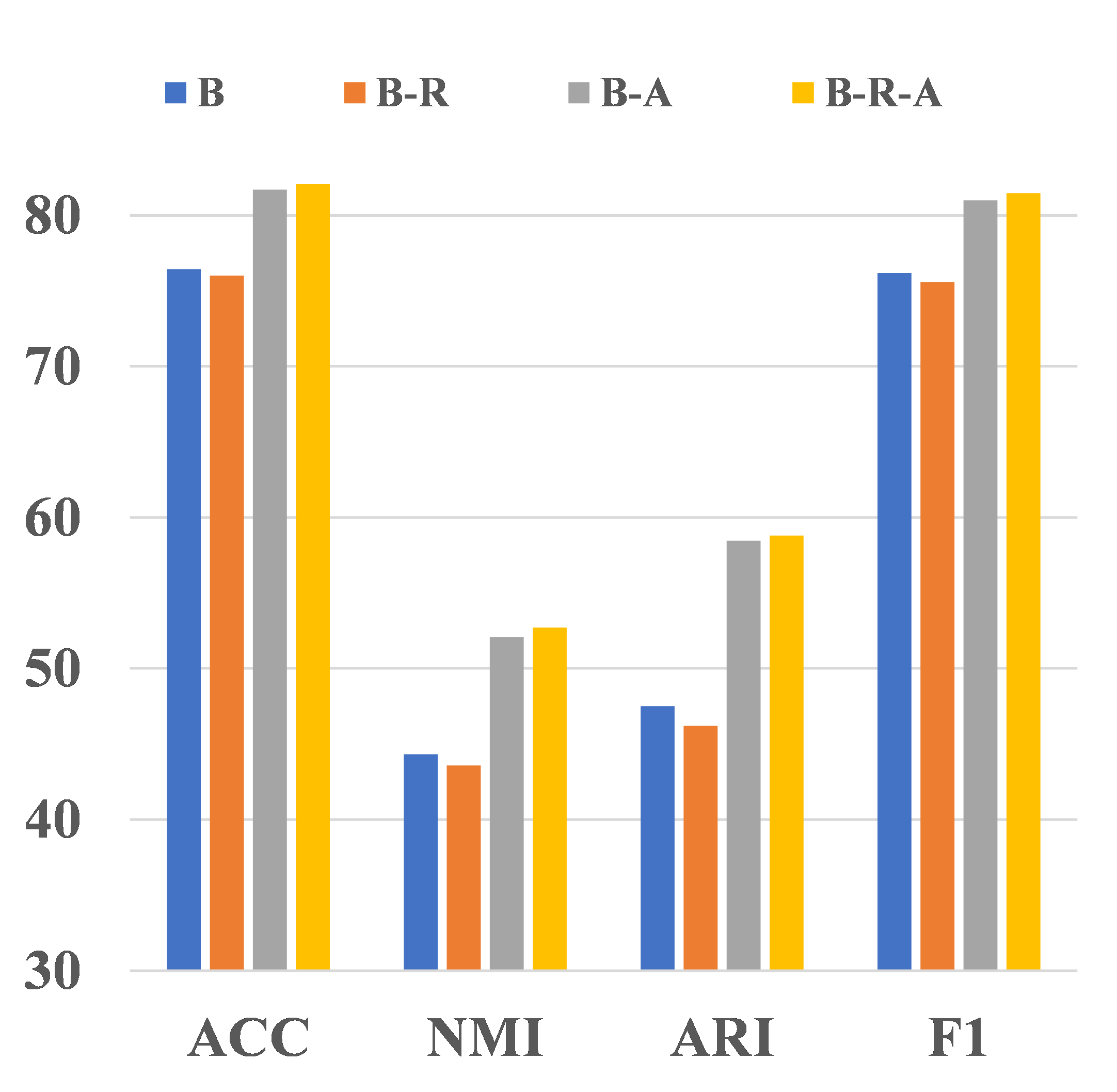}}
\vspace{3pt}
\centerline{(a) DBLP}
\vspace{3pt}
\centerline{\includegraphics[width=0.9\textwidth]{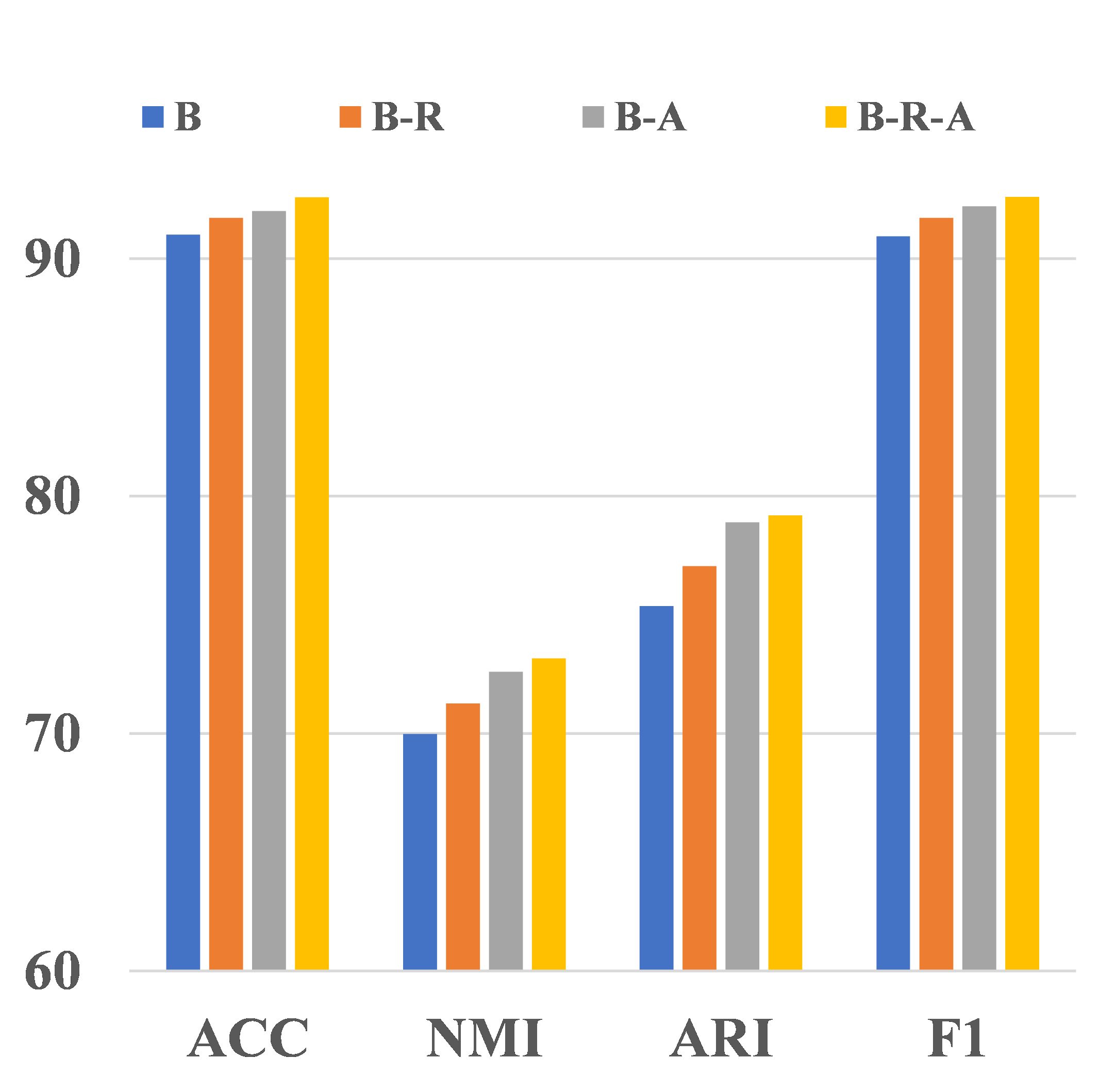}}
\vspace{3pt}
\centerline{(c) ACM}
\vspace{3pt}
\centerline{\includegraphics[width=0.9\textwidth]{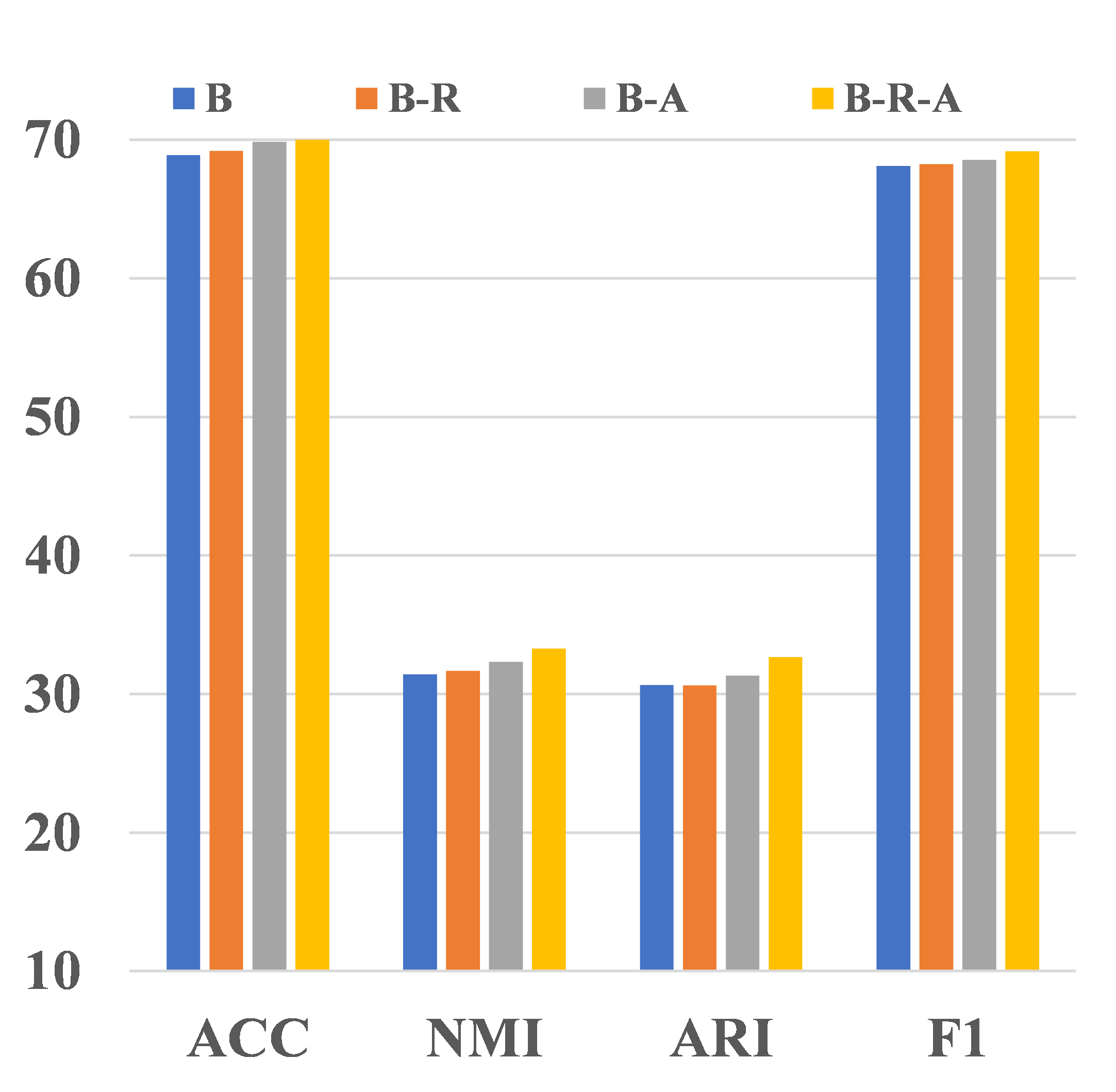}}
\vspace{3pt}
\centerline{(e) PUBMED}

\end{minipage}
\begin{minipage}{0.49\linewidth}
\centerline{\includegraphics[width=0.9\textwidth]{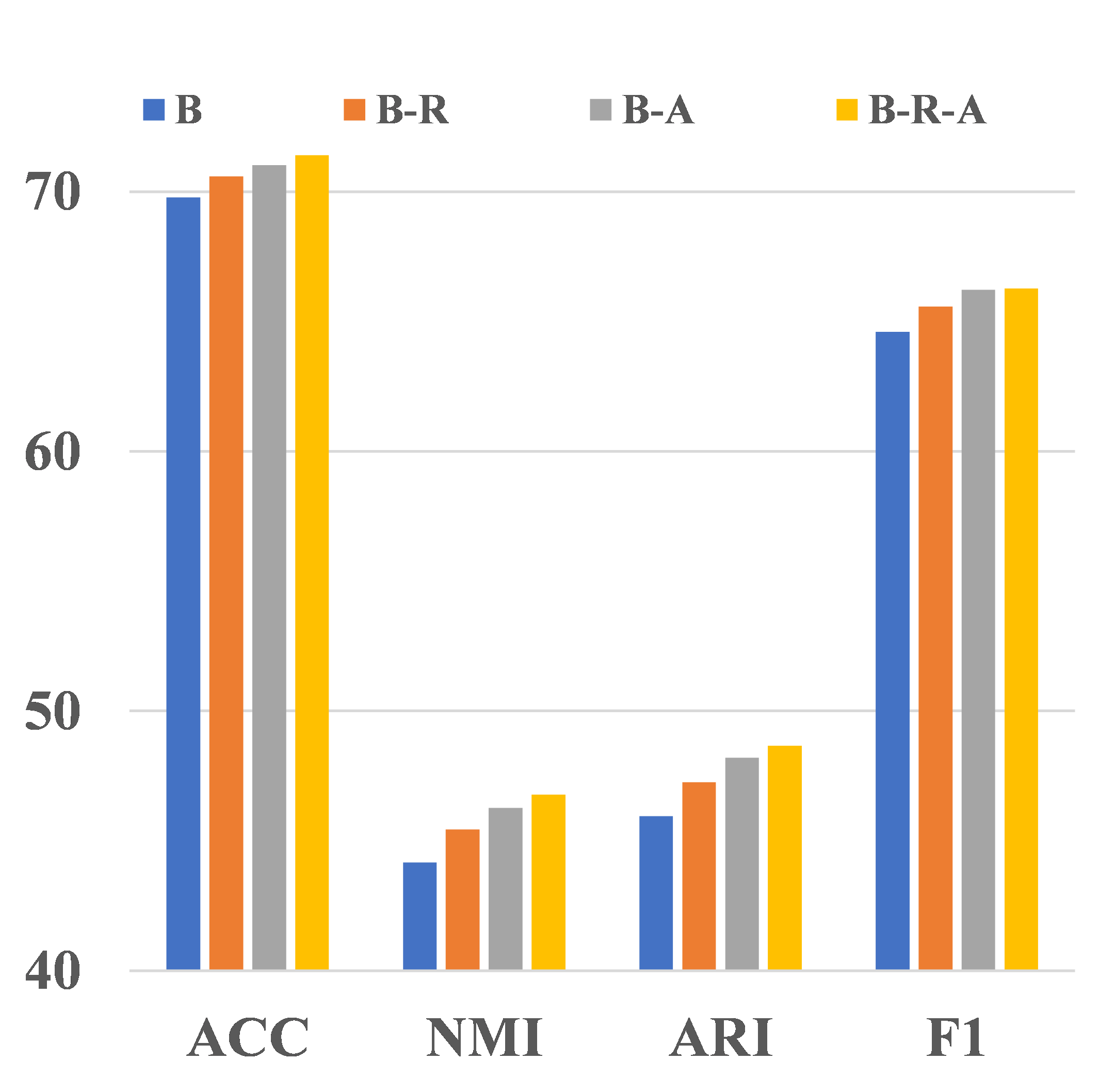}}
\vspace{3pt}
\centerline{(b) CITE}
\vspace{3pt}
\centerline{\includegraphics[width=0.9\textwidth]{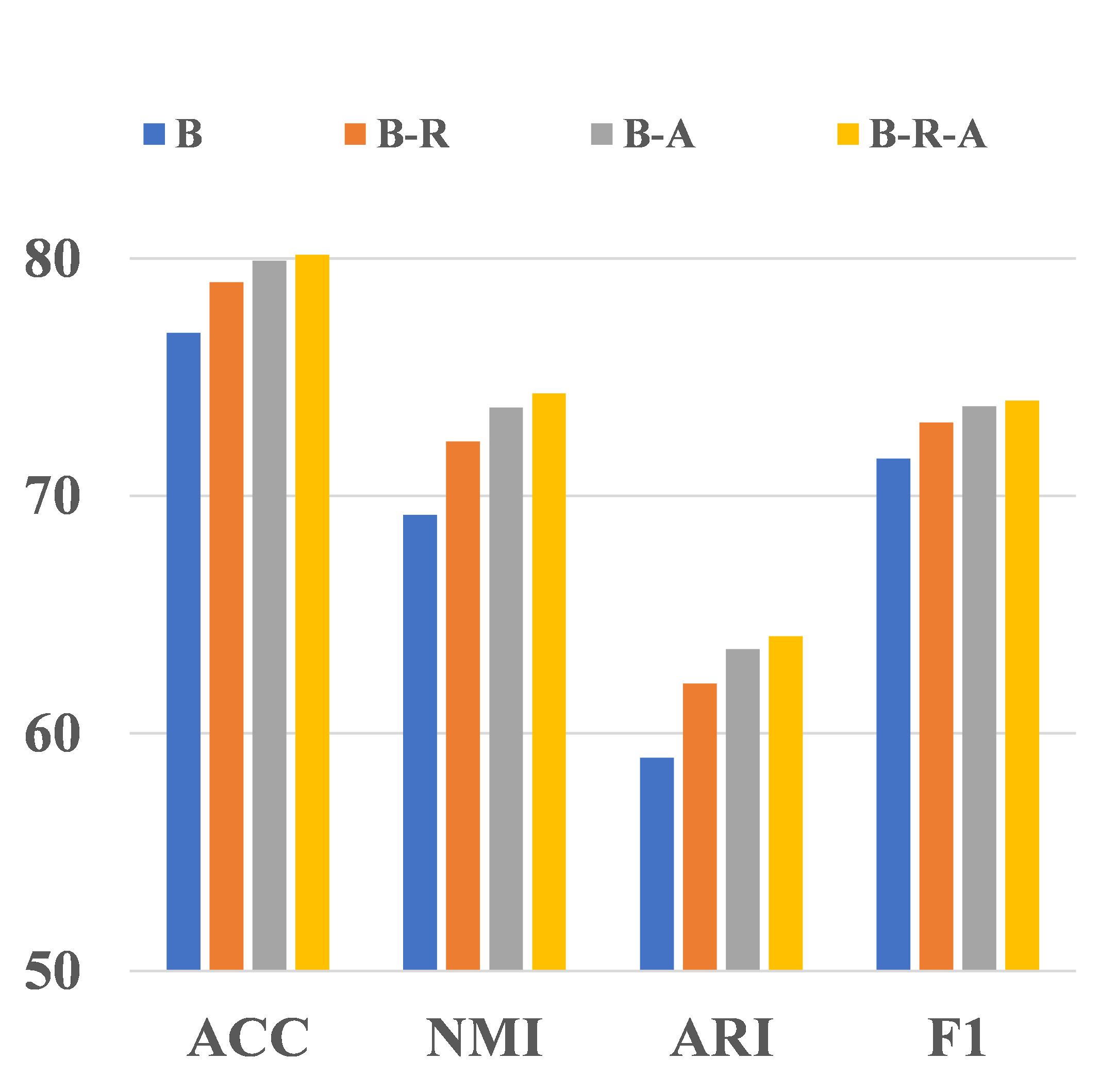}}
\vspace{3pt}
\centerline{(d) AMAP}
\vspace{3pt}
\centerline{\includegraphics[width=0.9\textwidth]{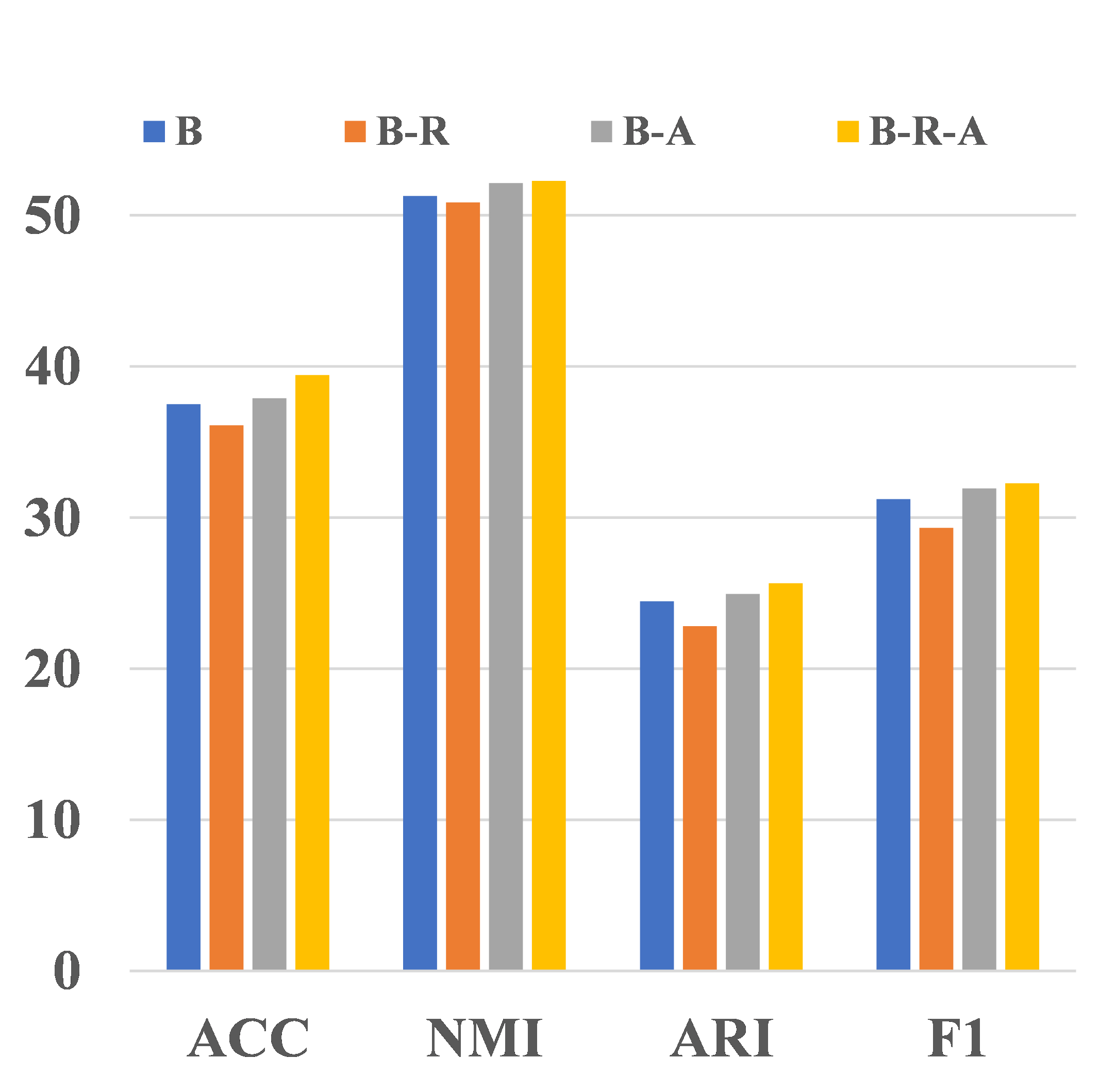}}
\vspace{3pt}
\centerline{(f) CORAFULL}
\end{minipage}
\caption{The ablation study results of the proposed two strategies, i.e., Affinity Recovery Strategy (ARS) and Redundancy Reduction Strategy (RRS).}
\label{SAMPLE_FEATURE_ABLATION}
\end{figure}

Furthermore, we carry on the ablation studies of the two proposed strategy including Affinity Recovery Strategy (ARS) and Redundancy Reduction Strategy (RRS). Here, we adopt DFCN \cite{DFCN} as our baseline. Then we denote B-R, B-A and, B-R-A as the baseline with RRS, ARS, and both, respectively. By observing the results in Fig. \ref{SAMPLE_FEATURE_ABLATION}, we have three following three conclusions. 

\begin{itemize}
\item B-R achieves better clustering performance than the baseline on four of six datasets since the learned embeddings is not robust without revealing the underlying sample distribution. 

\item The baseline with ARS significantly outperforms the baseline on six datasets. Take the results on DBLP dataset for an instance, B-A obtains 5.63\% accuracy improvement. Benefited from our proposed ARS, the network is guided to reveal the underlying sample distribution, thus enhancing the discriminative capability of the learned features. 

\item Moreover, B-R-A achieves the best clustering performance, further indicating that our proposed RRS and ARS effectively improve the discriminative capability of the learned embeddings.

\end{itemize}

\subsection{Sensitivity Analysis of Hyper-parameters}
We conduct extensive experiments to analyze the robustness of our proposed method IDCRN to the hyper-parameters. 

\subsubsection{Sensitivity Analysis of Graph Augmentation Module}
\begin{figure}[!h]
\centering
\small
\begin{minipage}{0.49\linewidth}
\centerline{\includegraphics[width=\textwidth]{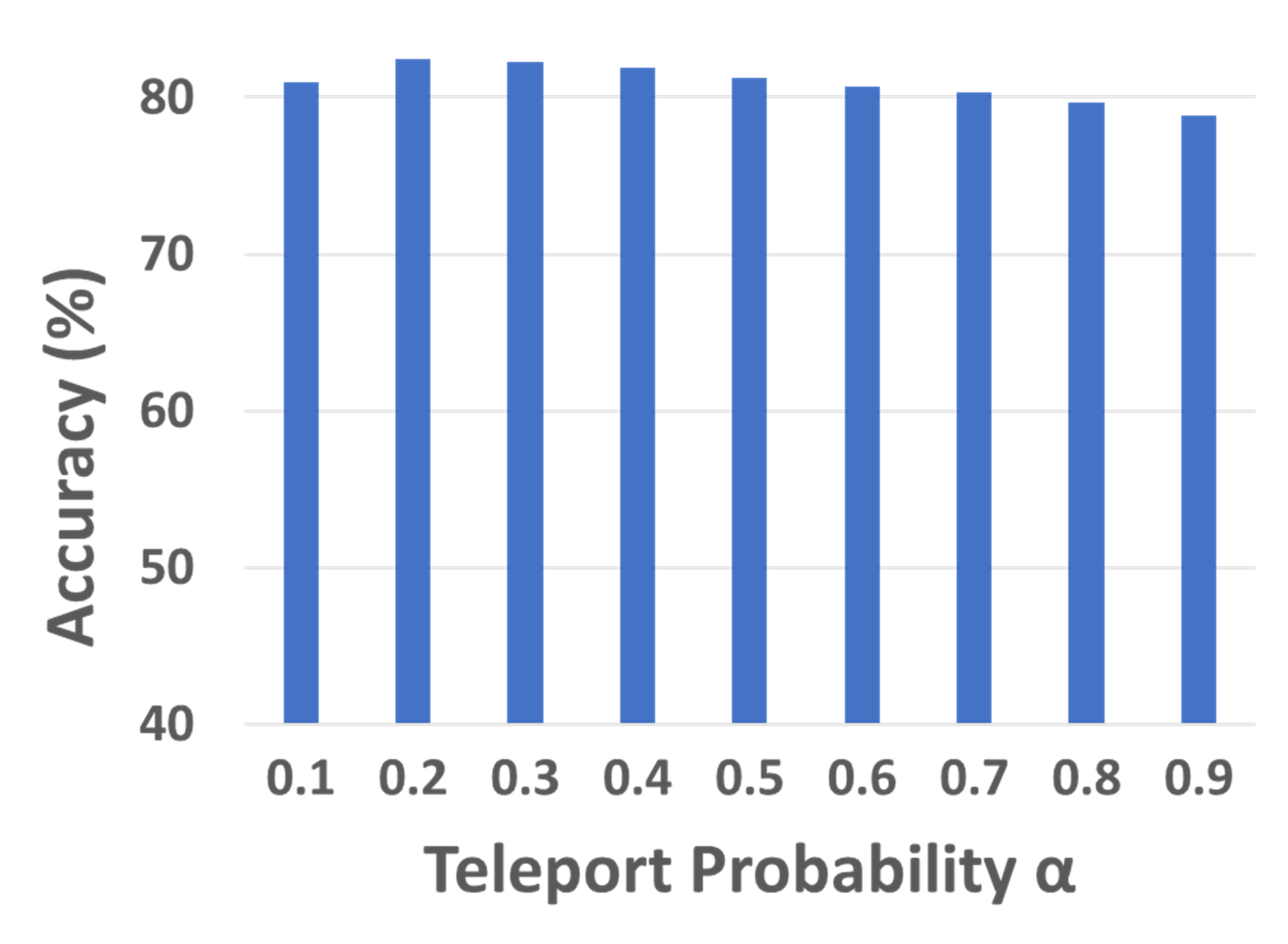}}
\centerline{(a) DBLP dataset}
\end{minipage}
\begin{minipage}{0.49\linewidth}
\centerline{\includegraphics[width=\textwidth]{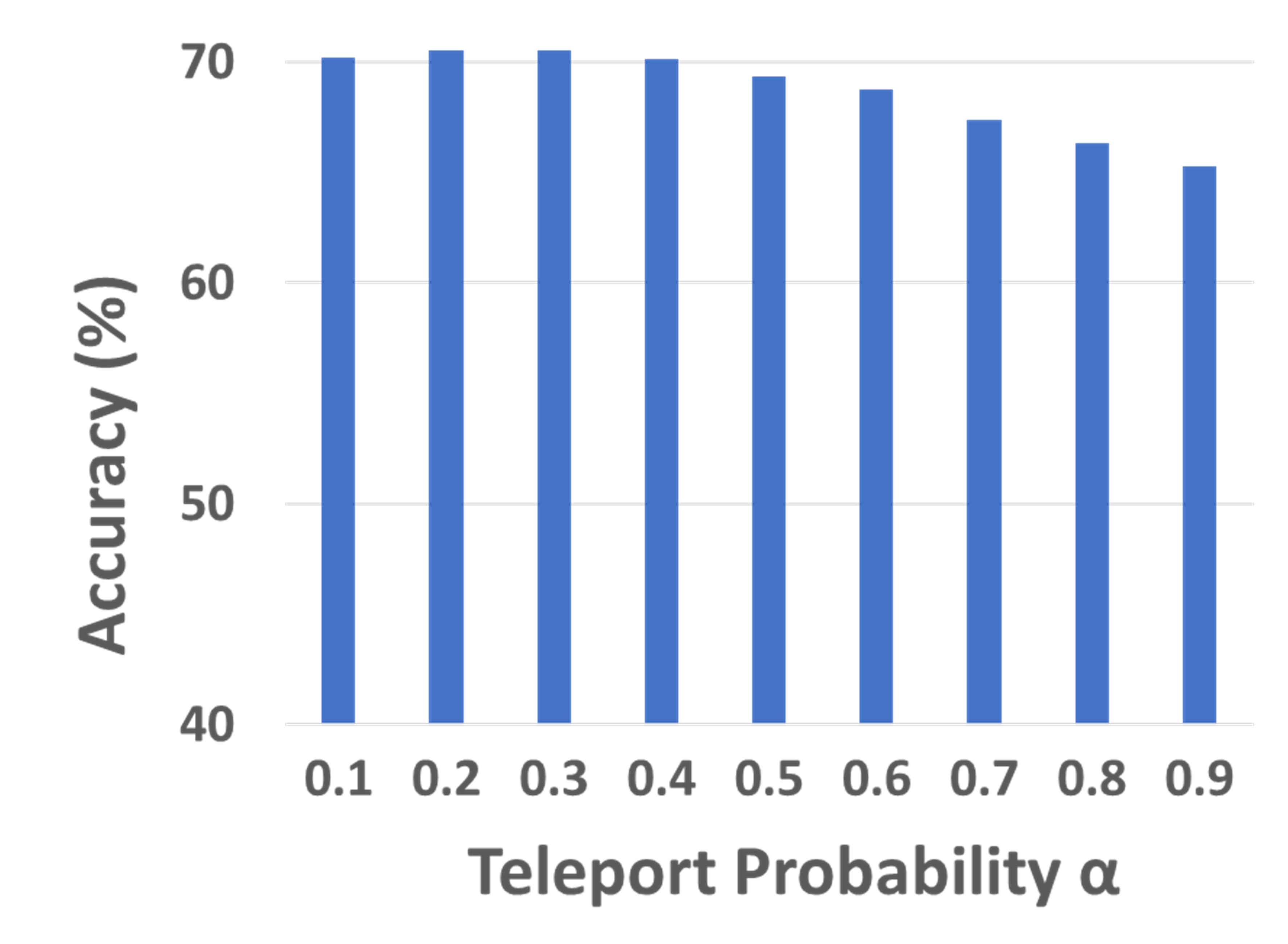}}
\centerline{(b) CITE dataset}
\end{minipage}
\begin{minipage}{0.49\linewidth}
\centerline{\includegraphics[width=\textwidth]{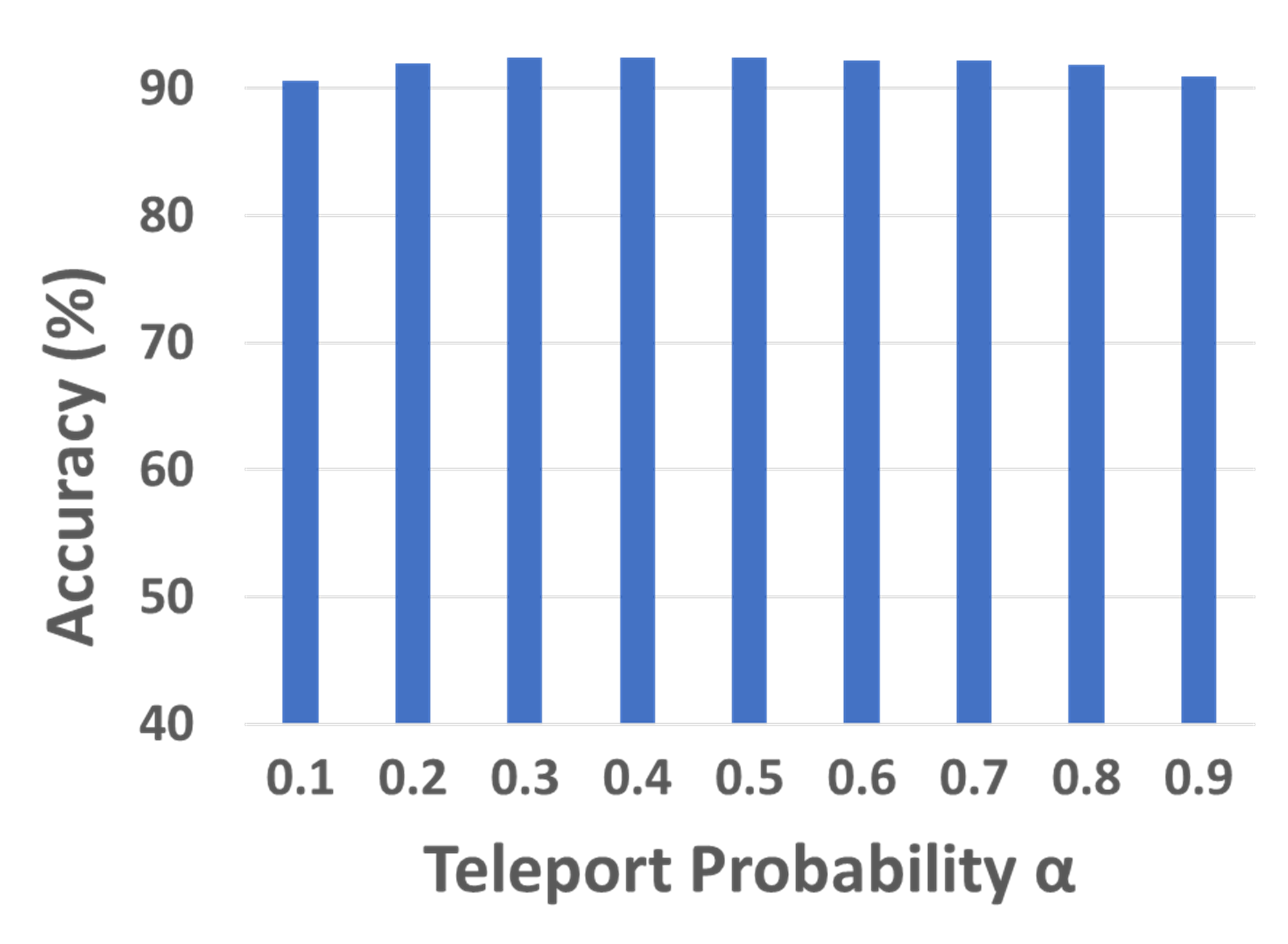}}
\centerline{(c) ACM dataset}
\end{minipage}
\begin{minipage}{0.49\linewidth}
\centerline{\includegraphics[width=\textwidth]{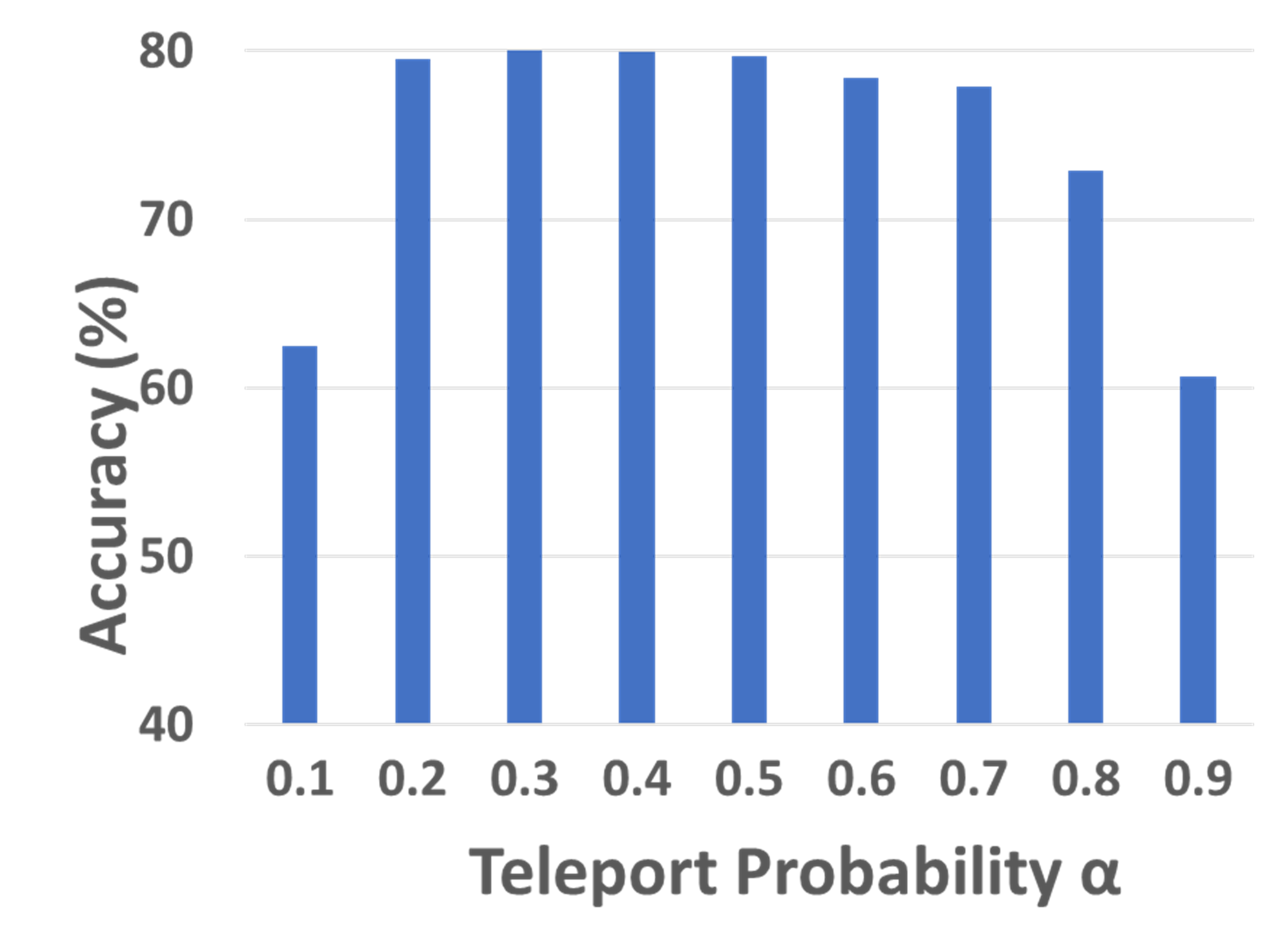}}
\centerline{(d) AMAP dataset}
\end{minipage}
\vspace{5pt}
\caption{Sensitivity Analysis of the teleport probability $\alpha$ in the graph diffusion on DBLP, CITE, ACM, and AMAP datasets.}
\vspace{5pt}
\label{TP_ABLATION}
\end{figure}

In order to investigate the influence of graph augmentation module, we first conduct an experiment about the teleport probability $\alpha$ in the graph diffusion, i.e., Personalized PageRank (PPR) \cite{PAGERANK}, on DBLP, CITE, ACM, and AMAP datasets. In Fig. \ref{TP_ABLATION}, we could observe that the accuracy firstly raises and reaches the high peak value where the teleport probability $\alpha$ is around 0.2 while the clustering performance decreases down with the larger teleport probability $\alpha$. Besides, our proposed method is robust to the teleport probability $\alpha$ when $\alpha \in (0.2, 0.8)$.

\begin{figure}[!h]
\centering
\small
\begin{minipage}{0.49\linewidth}
\centerline{\includegraphics[width=\textwidth]{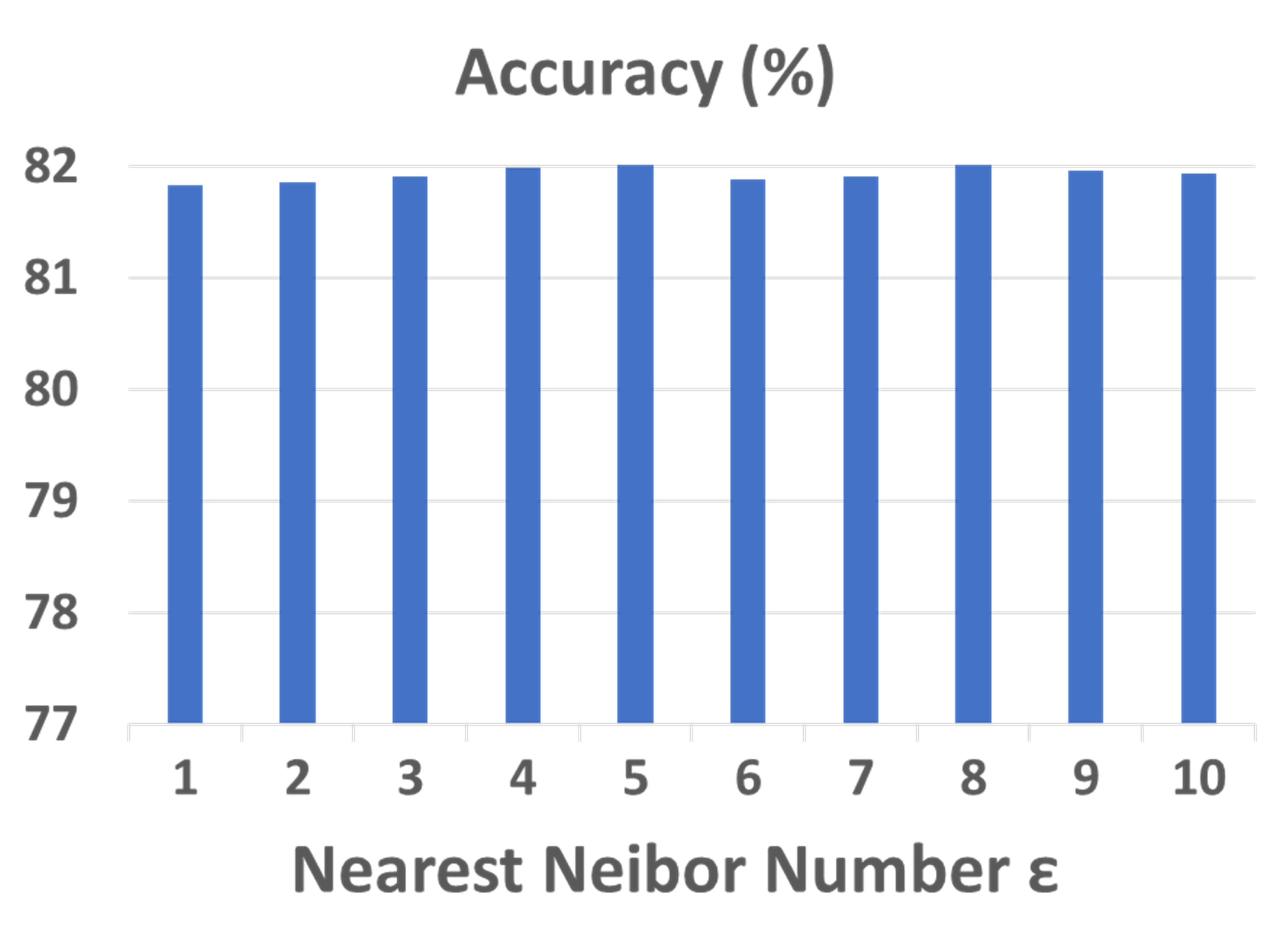}}
\centerline{(a) DBLP dataset}
\end{minipage}
\begin{minipage}{0.49\linewidth}
\centerline{\includegraphics[width=\textwidth]{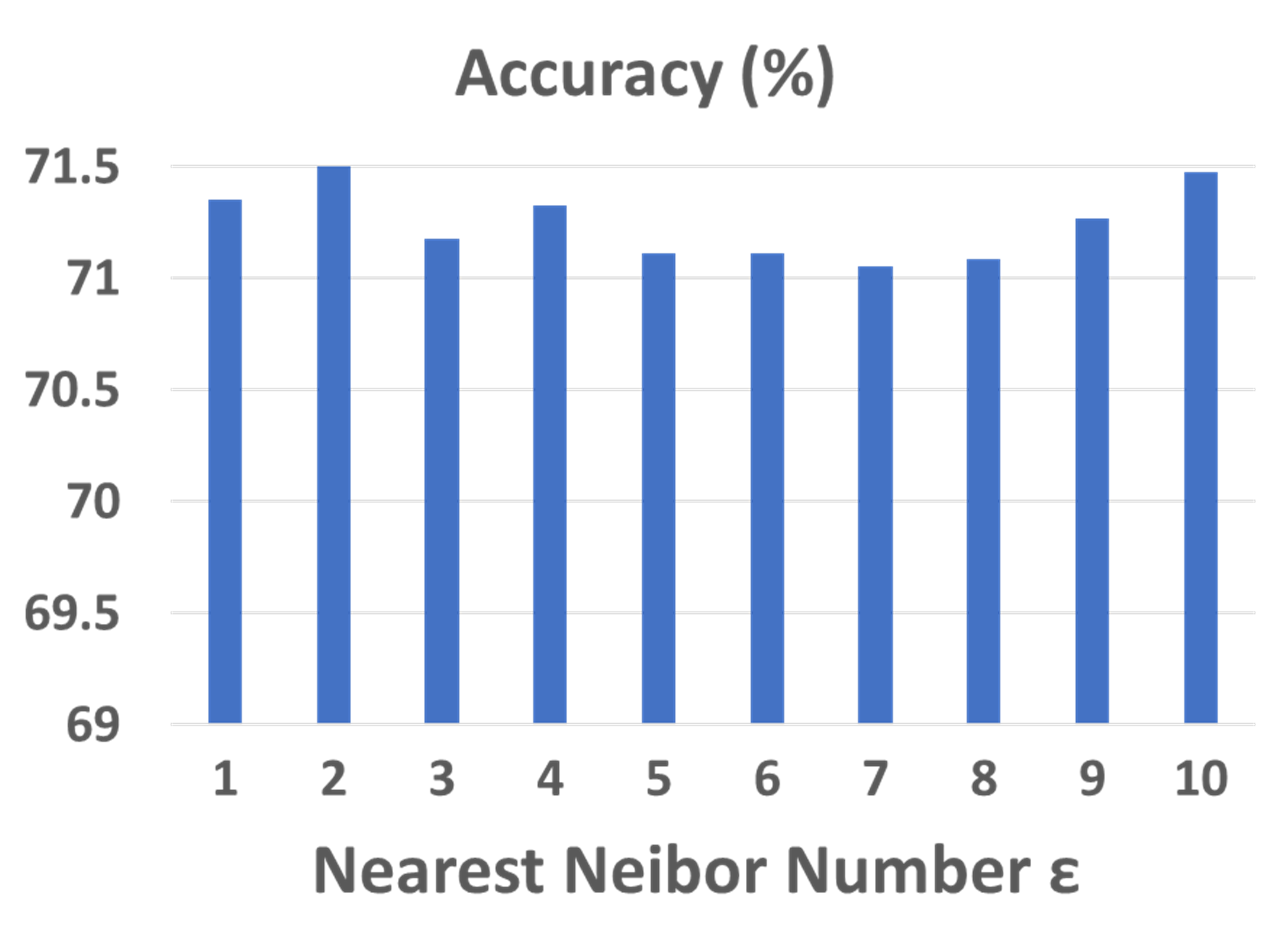}}
\centerline{(b) CITE dataset}
\end{minipage}
\begin{minipage}{0.49\linewidth}
\centerline{\includegraphics[width=\textwidth]{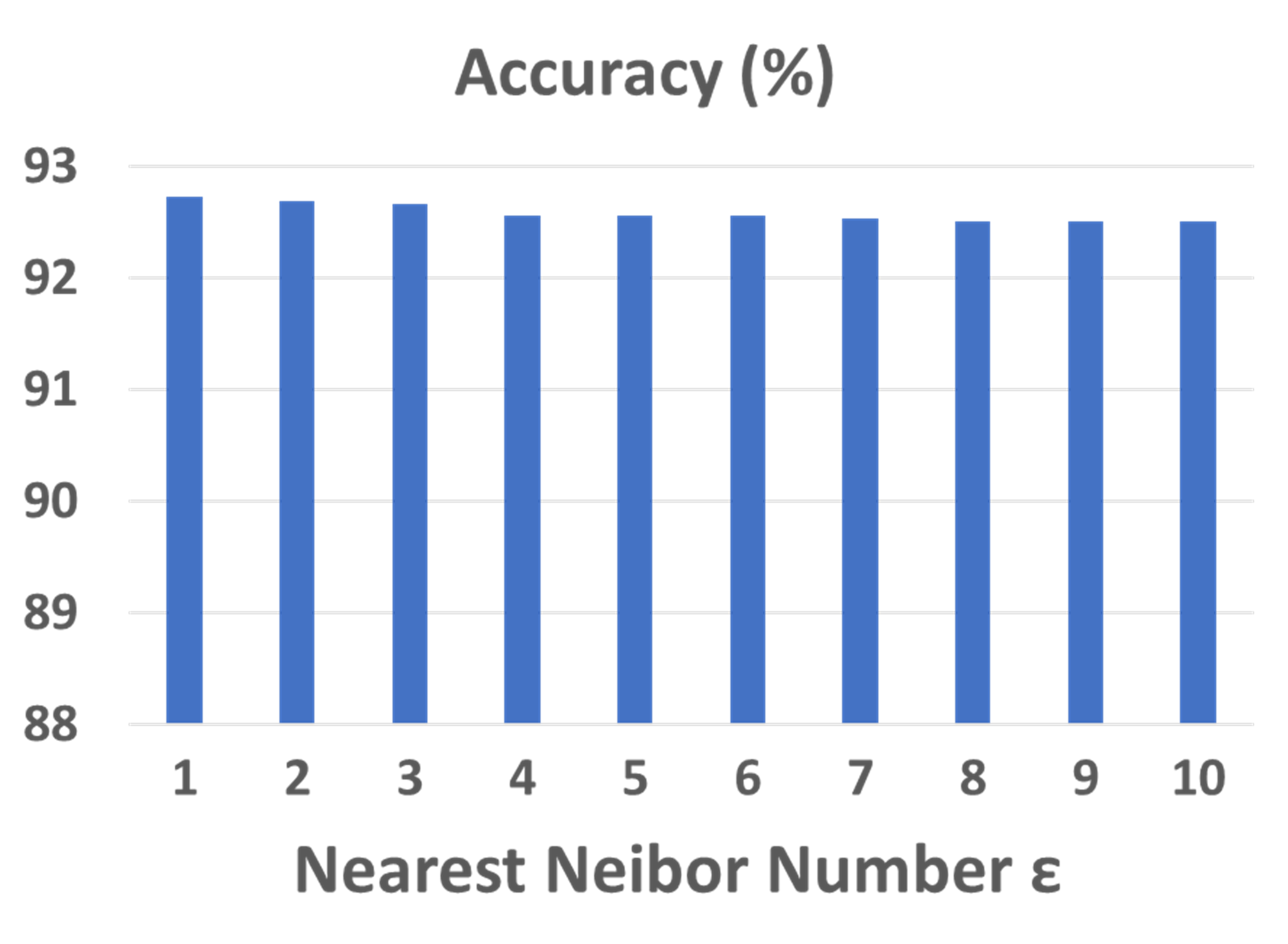}}
\centerline{(c) ACM dataset}
\end{minipage}
\begin{minipage}{0.49\linewidth}
\centerline{\includegraphics[width=\textwidth]{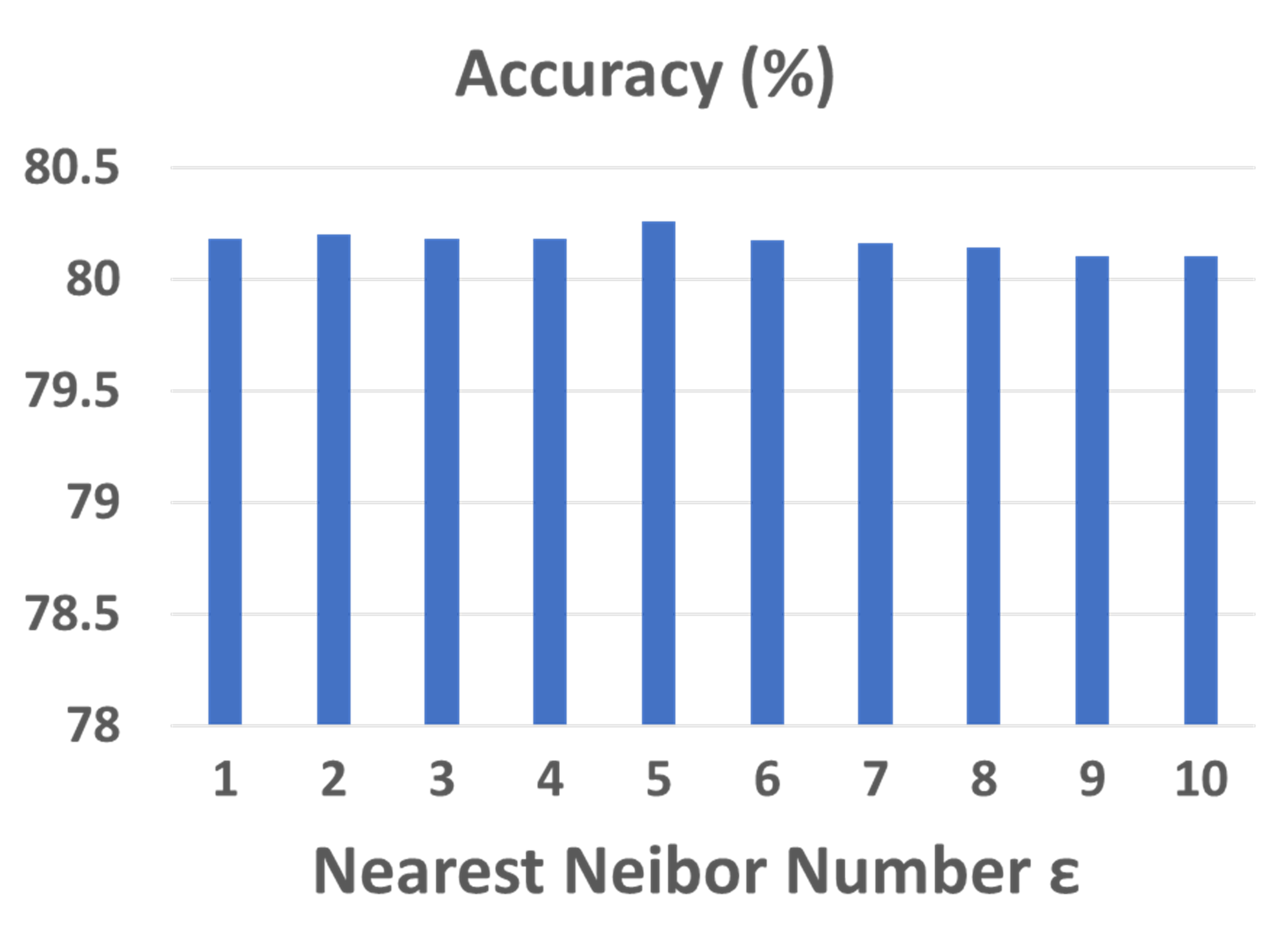}}
\centerline{(b) AMAP dataset}
\end{minipage}
\vspace{5pt}
\caption{Sensitivity Analysis of the nearest neighbor number $\epsilon$ in the KNN adjacency matrix generation on DBLP, CITE, ACM and AMAP datasets.}
\vspace{5pt}
\label{EPILON_ABLATION}
\end{figure}

In addition, we also explore the influence of the hyper-parameter $\epsilon$ in K-nearest neighbors algorithm (KNN) \cite{KNN} during the process of the KNN graph adjacency matrix generation. From Fig. \ref{EPILON_ABLATION}, we observe and conclude the ACC metric of clustering does not fluctuate significantly with the variation of $\epsilon$ so that our proposed IDCRN is insensitive to the nearest neighbor number $\epsilon$.

\begin{figure}[h]
\centering
\small
\begin{minipage}{0.49\linewidth}
\centerline{\includegraphics[width=\textwidth]{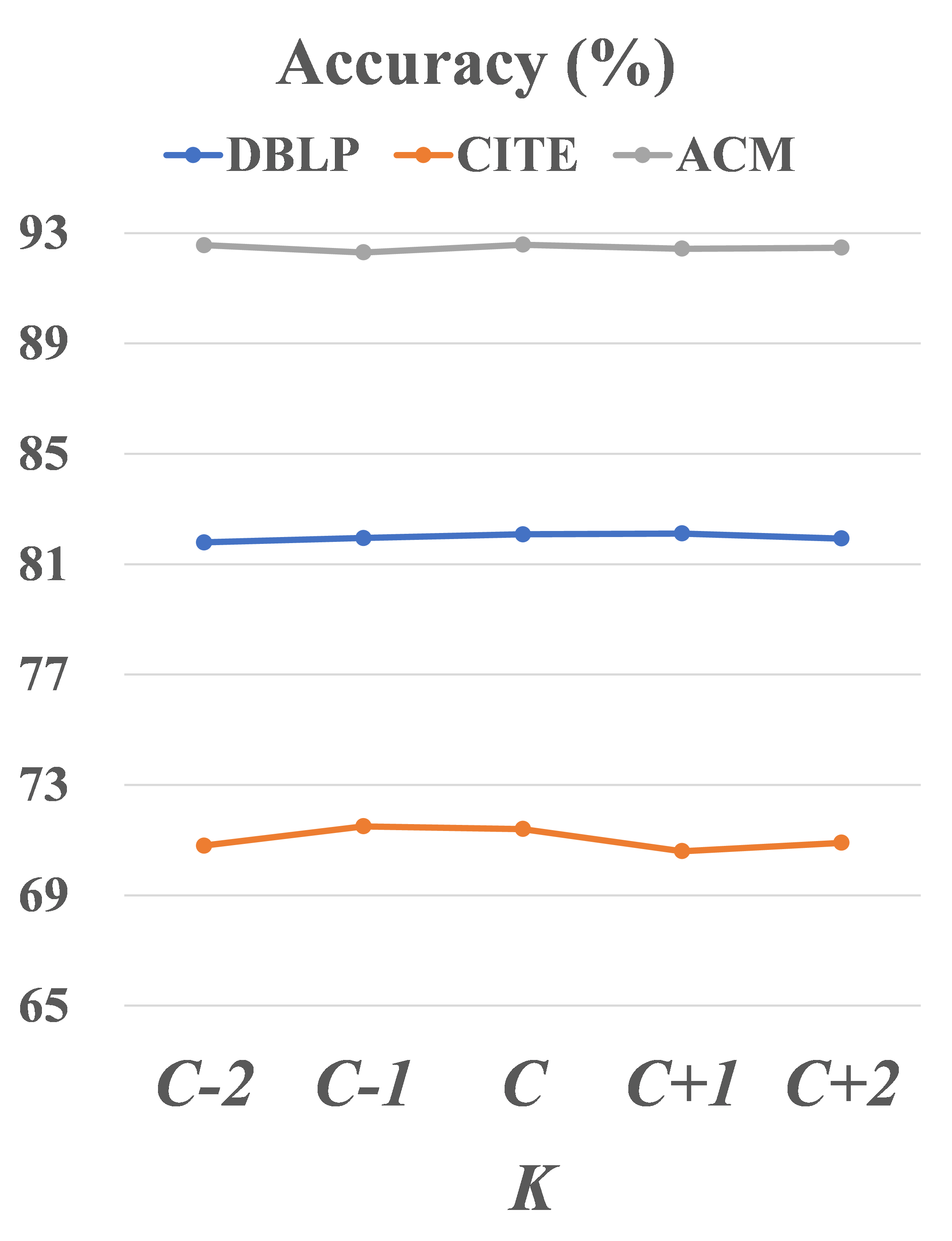}}
\centerline{(a) 
}
\end{minipage}
\begin{minipage}{0.49\linewidth}
\centerline{\includegraphics[width=\textwidth]{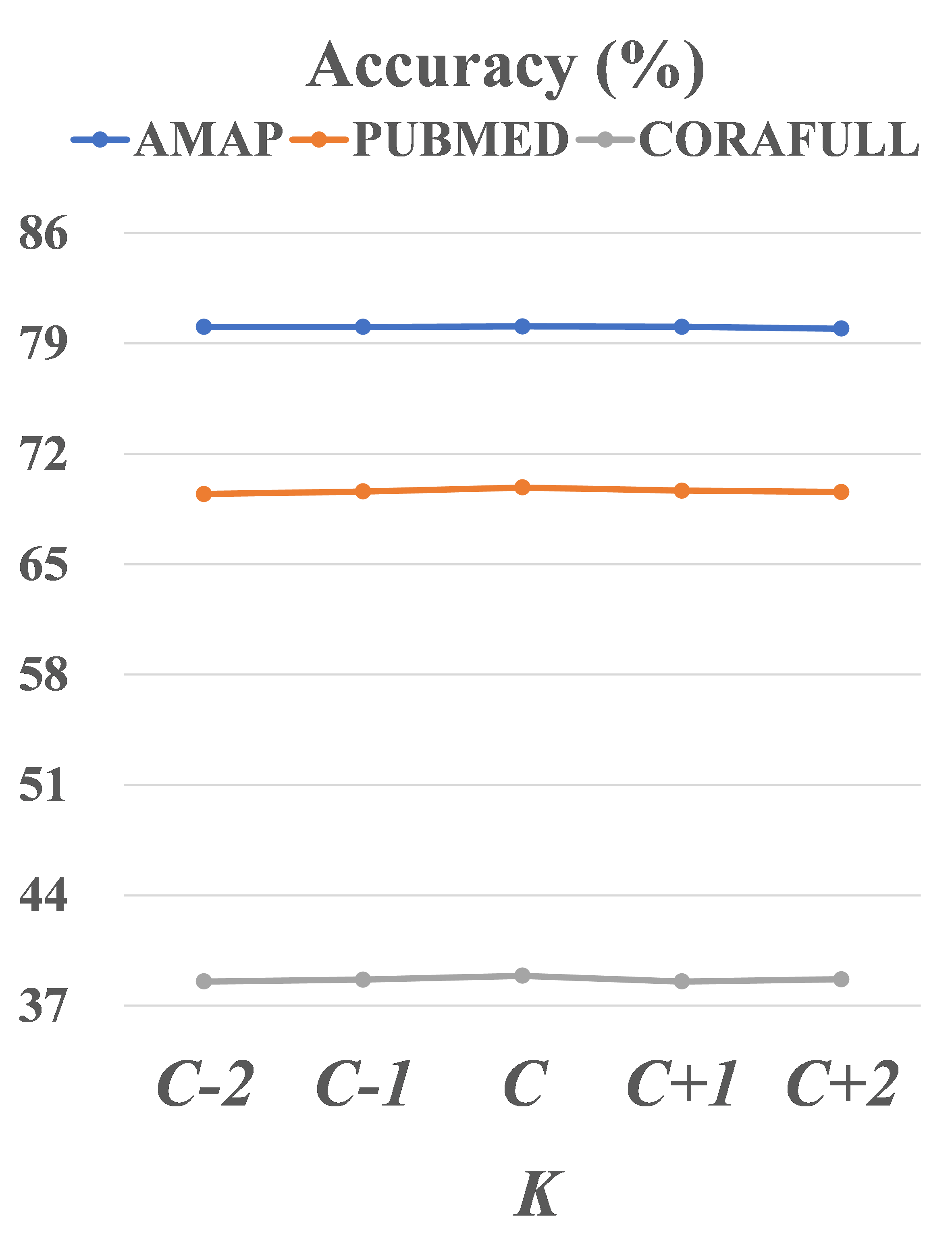}}
\centerline{(b)
}
\end{minipage}
\vspace{5pt}
\caption{Sensitivity Analysis of hyper-parameter $K$. The results on DBLP, CITE, ACM (sub-figure a) and AMAP, PUBMED, CORAFULL (sub-figure b) datasets are illustrated.}
\vspace{5pt}
\label{K_ABLATOIN}
\end{figure}

\subsubsection{Sensitivity Analysis of Hyper-parameter $K$}
The sensitivity of hyper-parameter $K$ in IDCRN is explored. Fig. \ref{K_ABLATOIN} shows that the accuracy of IDCRN firstly increases to the high peek value and then stays at it with the slight perturbation as $K$ increases. Besides, our proposed method is robust to the variation of $K$.

\subsection{GPU Memory Costs and Time Costs}
\begin{figure}[h]
\centering
\begin{minipage}{0.49\linewidth}
\centerline{\includegraphics[width=\textwidth]{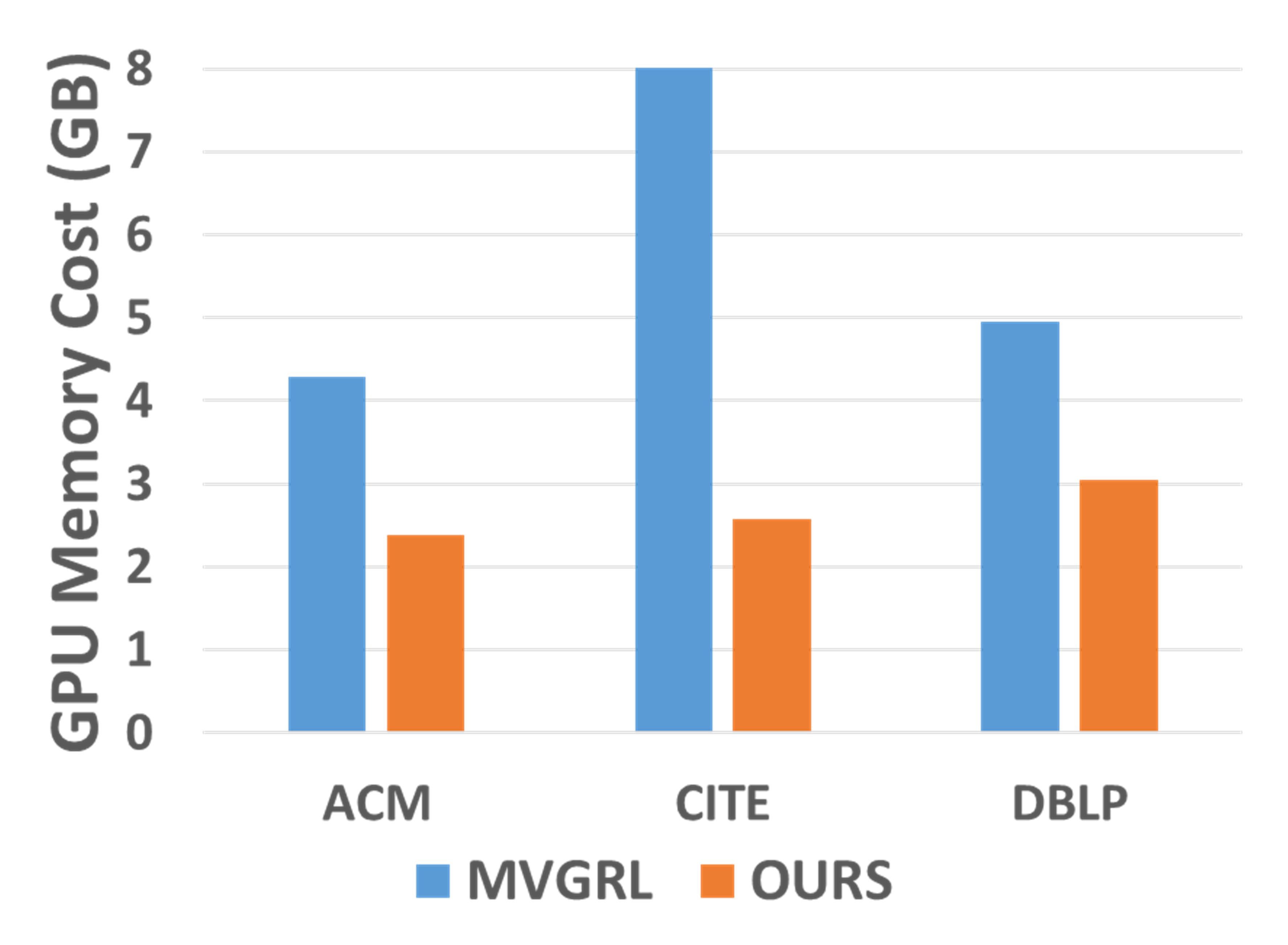}}
\centerline{(a) GPU memory costs}
\end{minipage}
\begin{minipage}{0.49\linewidth}
\centerline{\includegraphics[width=\textwidth]{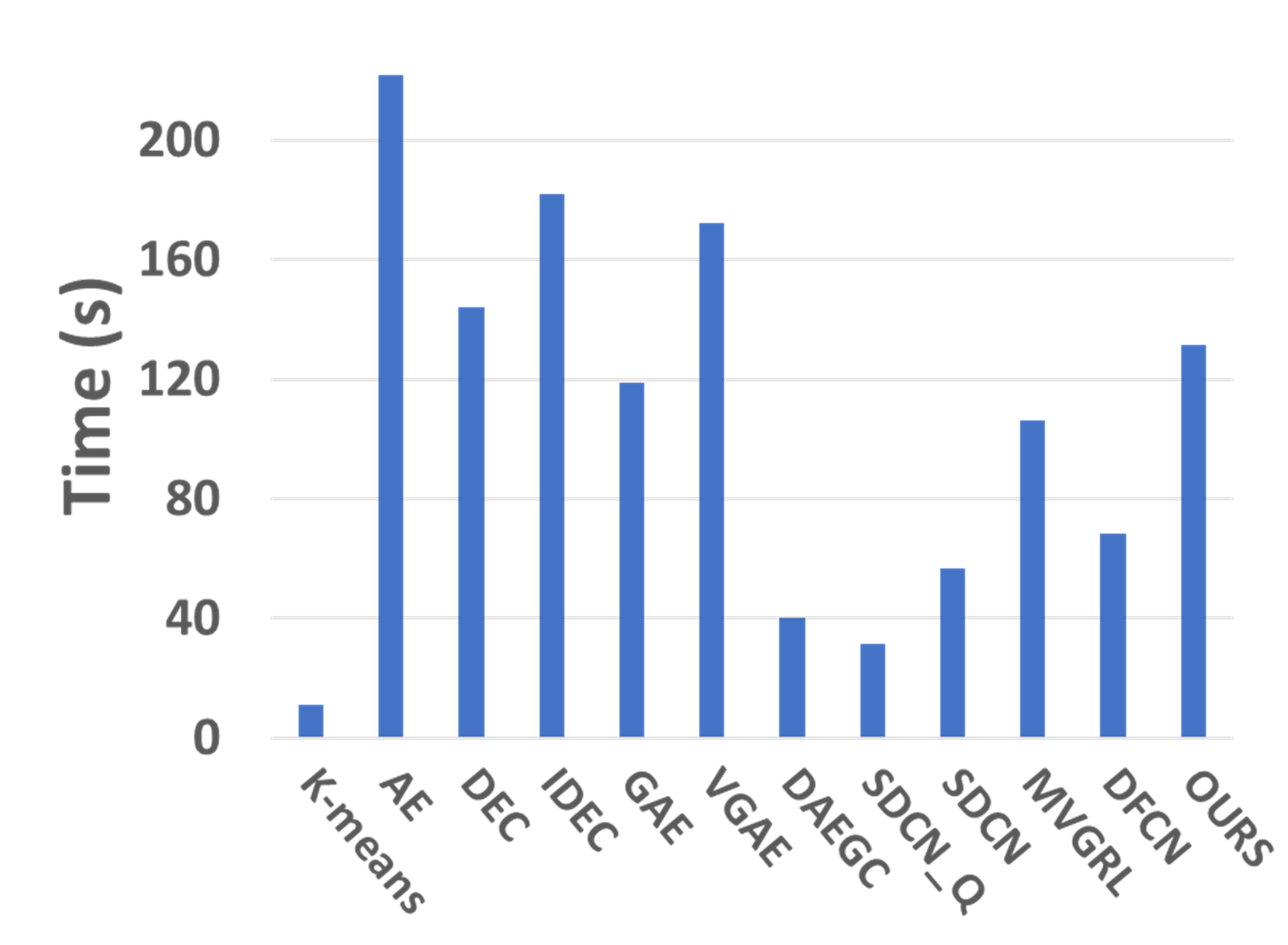}}
\centerline{(b) Time costs}
\end{minipage}
\vspace{5pt}
\caption{GPU memory (sub-figure a) and time cost (sub-figure b) comparison between our method and the state-of-the-art methods.}
\vspace{5pt}
\label{COST}
\end{figure}

GPU memory and time costs are two important indicators for the algorithm evaluation. Compared to other contrastive-learning-base methods, IDCRN could save GPU memory costs since it eliminates the space-consuming negative sample generation. In order to certify this advantage of IDCRN, we conduct experiments of GPU memory costs and report the average results on the ACM, CITE, and DBLP datasets in Fig. \ref{COST} (a). From the results, we observe that IDCRN saves about 53.55 \% GPU memory against MVGRL \cite{MVGRL} on average. Furthermore, we also test the algorithm running time of the baselines and IDCRN in Fig. \ref{COST} (b) on ACM dataset. From these results, we observe that our method has comparable time costs compared to other baselines. The running time of our proposed method could be optimized in the future. 

\subsection{Visualization Experiments}
To intuitively show the superiority of IDCRN, two visualization experiments are conducted in this section.
\subsubsection{Visualization of Node Similarity Matrices}
\begin{figure}[!t]
\small
\begin{minipage}{0.24\linewidth}
\centerline{\includegraphics[width=\textwidth]{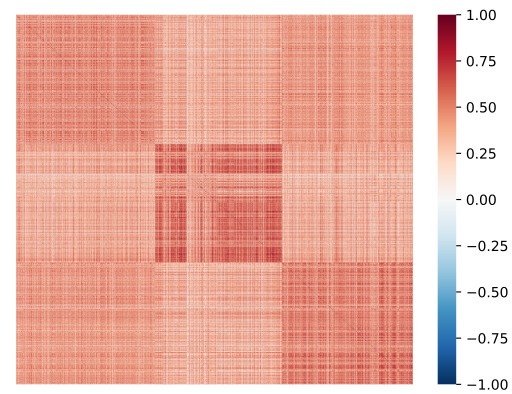}}
\vspace{5pt}
\centerline{\includegraphics[width=\textwidth]{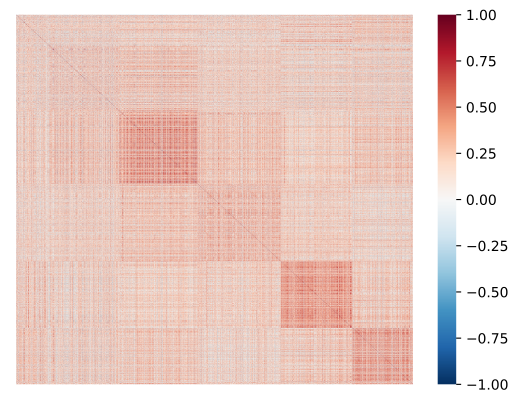}}
\vspace{5pt}
\centerline{(a) GAE}
\end{minipage}
\begin{minipage}{0.24\linewidth}
\centerline{\includegraphics[width=\textwidth]{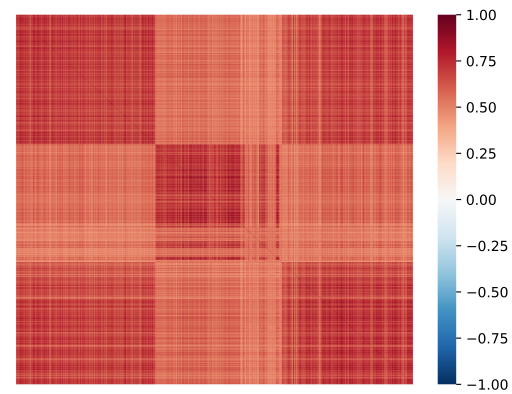}}
\vspace{5pt}
\centerline{\includegraphics[width=\textwidth]{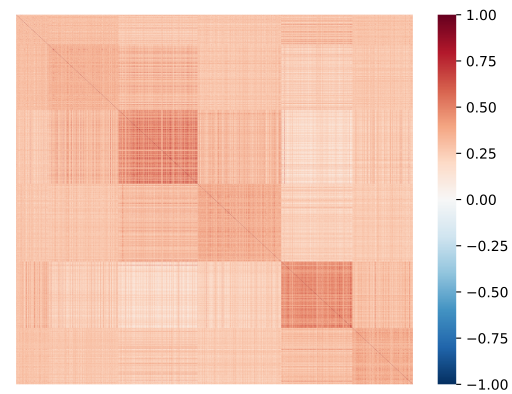}}
\vspace{5pt}
\centerline{(b) MVGRL}
\end{minipage}
\begin{minipage}{0.24\linewidth}
\centerline{\includegraphics[width=\textwidth]{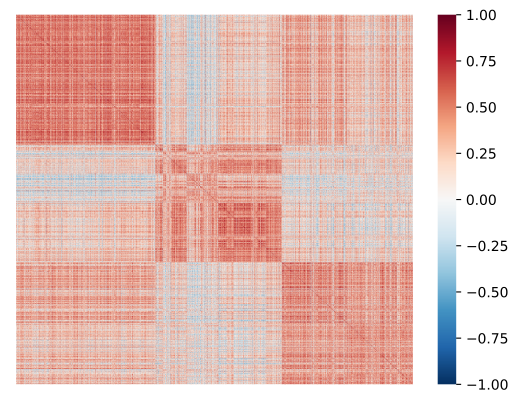}}
\vspace{5pt}
\centerline{\includegraphics[width=\textwidth]{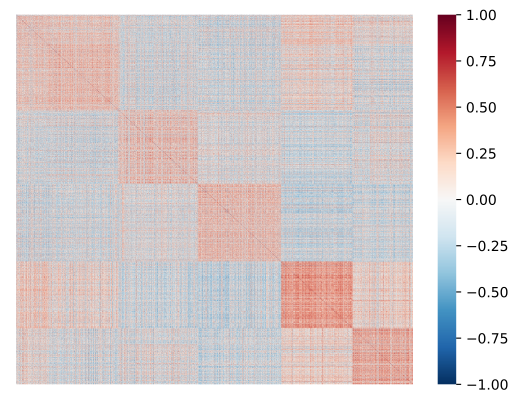}}
\vspace{5pt}
\centerline{(c)SDCN}
\end{minipage}
\begin{minipage}{0.24\linewidth}
\centerline{\includegraphics[width=\textwidth]{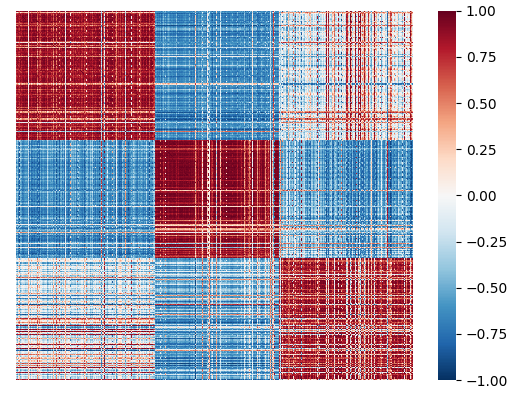}}
\vspace{5pt}
\centerline{\includegraphics[width=\textwidth]{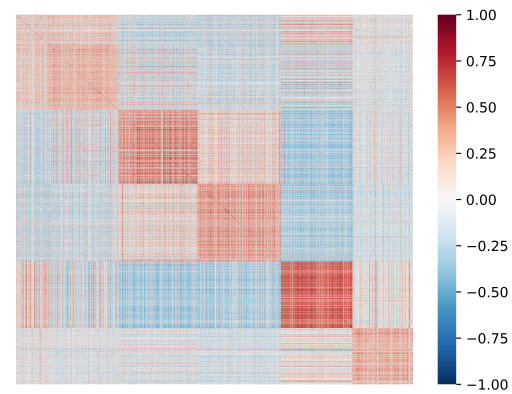}}
\vspace{5pt}
\centerline{(d) IDCRN}
\end{minipage}

\caption{Visualization of sample similarity matrices in the latent space learned by our proposed method (IDCRN), SDCN \cite{SDCN}, MVGRL \cite{MVGRL}, and GAE \cite{GAE} on two benchmarks. The first row and second row correspond to the results on ACM and CITE datasets, respectively.}
\label{SIMILARITY}  
\end{figure}

We plot the heat maps of sample similarity matrices in the learned space to intuitively show the representation collapse problem in deep clustering methods and the effectiveness of our solution to this issue. The red, blue, and white colors indicate positive correlation, negative correlation, and decorrelation, respectively. Here, we sort all samples by categories to make those from the same cluster beside each
other. As illustrated in Fig. \ref{SIMILARITY}, we observe that GAE \cite{GAE} and MVGRL \cite{MVGRL} would suffer from representation collapse during the process of node encoding. Unlike them, our proposed method learns the more discriminative latent features, thus avoiding the representation collapse.

\subsubsection{$t$-SNE Visualization of the Learned Embeddings}
\begin{figure*}[!t]
\begin{minipage}{0.139\linewidth}
\centerline{\includegraphics[width=\textwidth]{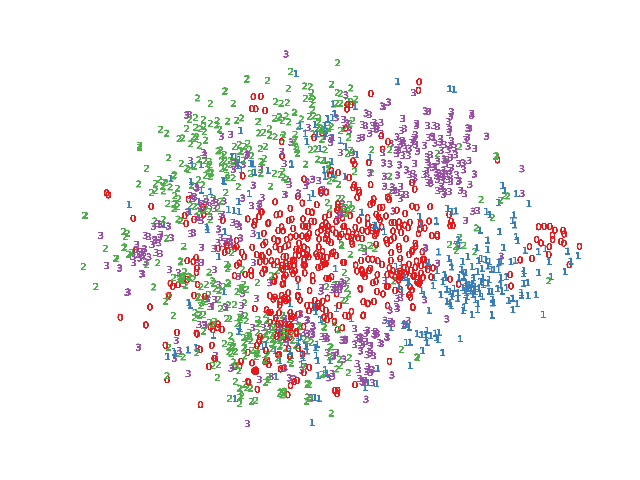}}
\vspace{5pt}
\centerline{\includegraphics[width=\textwidth]{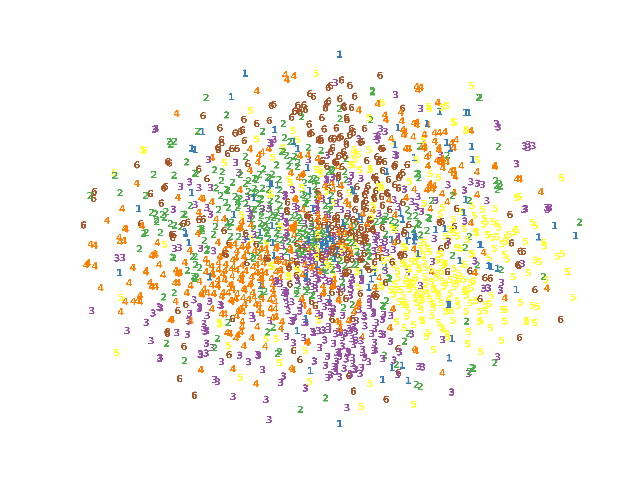}}
\vspace{5pt}
\centerline{\includegraphics[width=\textwidth]{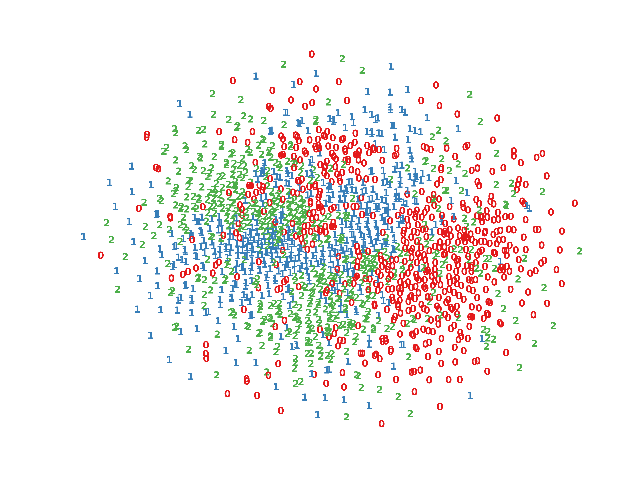}}
\vspace{5pt}
\centerline{\includegraphics[width=\textwidth]{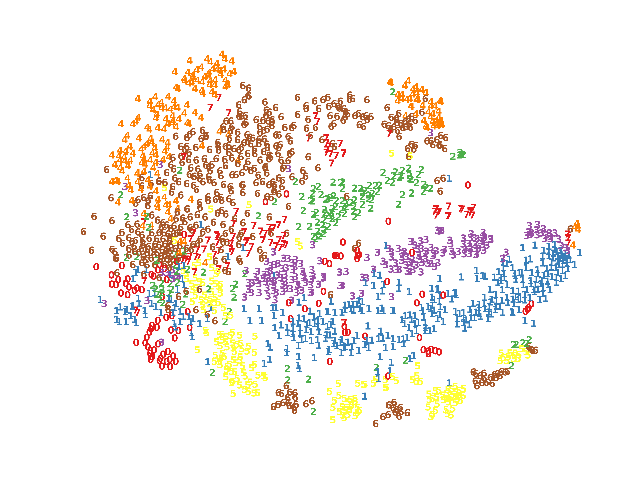}}
\vspace{5pt}
\centerline{(a) Raw Data}
\end{minipage}
\begin{minipage}{0.139\linewidth}
\centerline{\includegraphics[width=\textwidth]{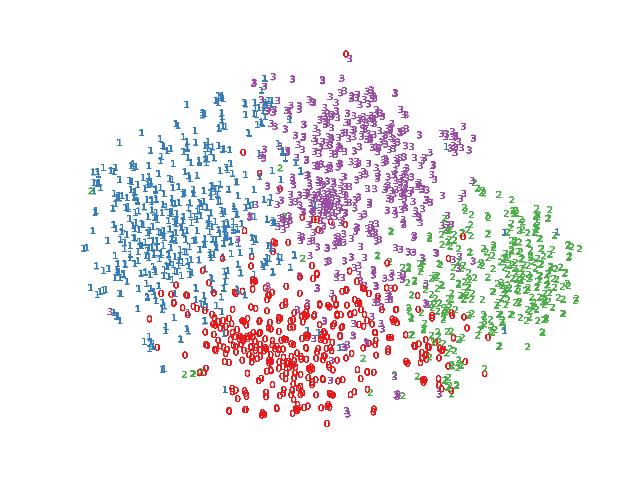}}
\vspace{5pt}
\centerline{\includegraphics[width=\textwidth]{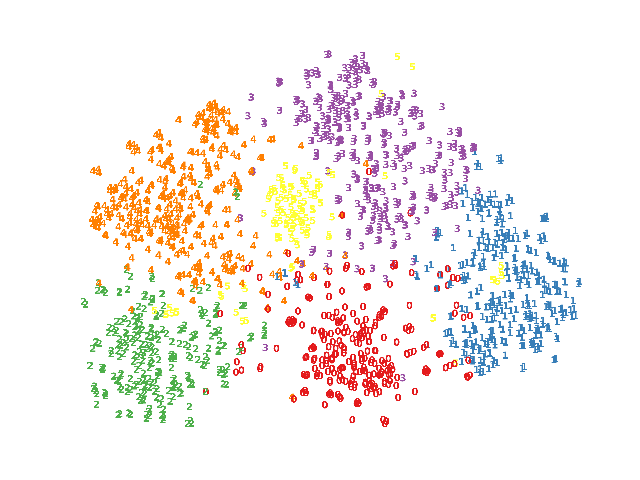}}
\vspace{5pt}
\centerline{\includegraphics[width=\textwidth]{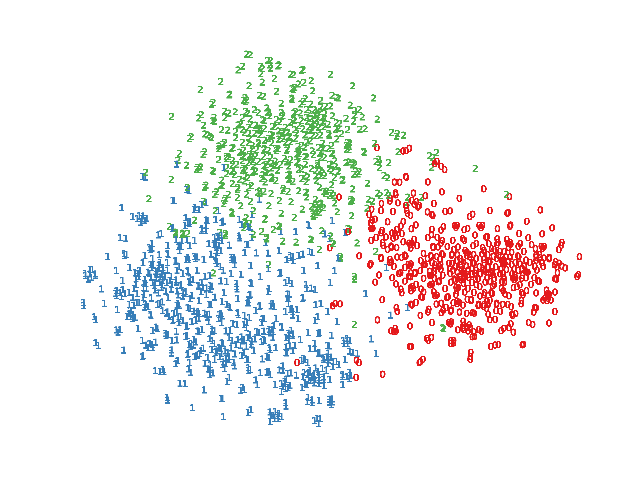}}
\vspace{5pt}
\centerline{\includegraphics[width=\textwidth]{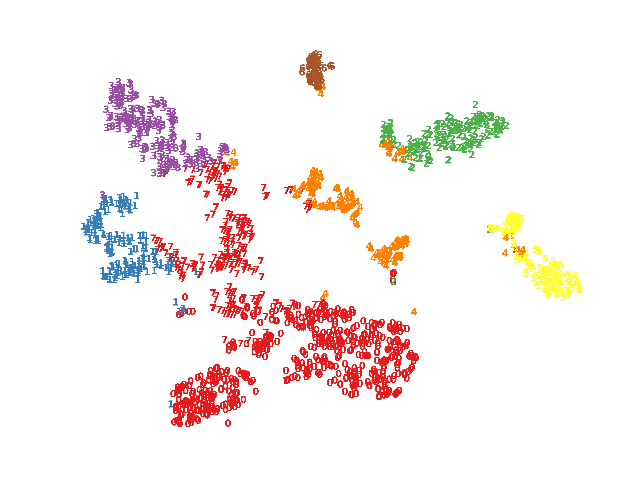}}
\vspace{5pt}
\centerline{(b) 0 Epoch}
\end{minipage}
\begin{minipage}{0.139\linewidth}
\centerline{\includegraphics[width=\textwidth]{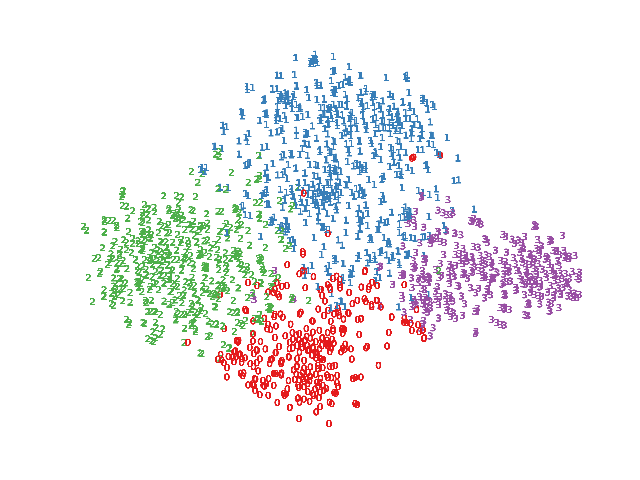}}
\vspace{5pt}
\centerline{\includegraphics[width=\textwidth]{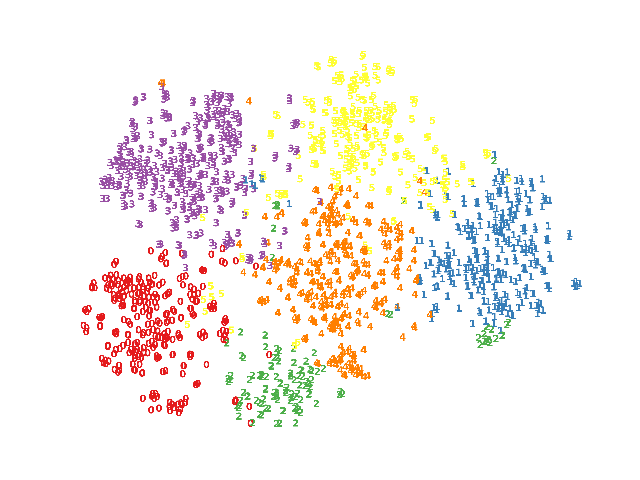}}
\vspace{5pt}
\centerline{\includegraphics[width=\textwidth]{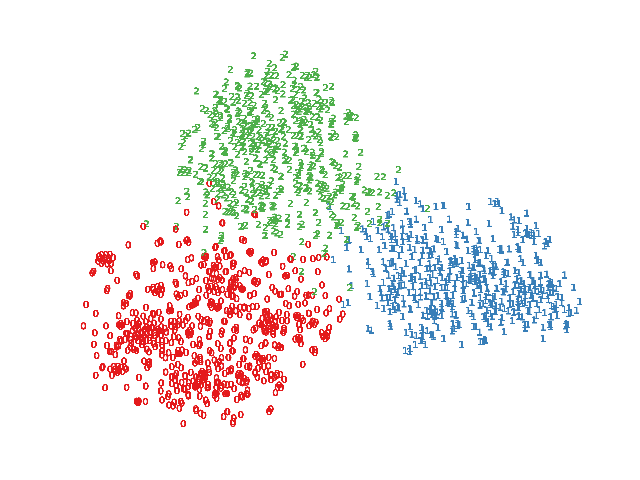}}
\vspace{5pt}
\centerline{\includegraphics[width=\textwidth]{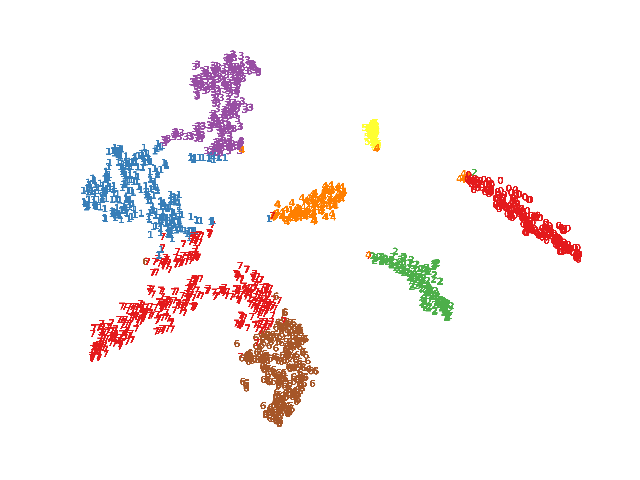}}
\vspace{5pt}
\centerline{(c) 80 Epoch}
\end{minipage}
\begin{minipage}{0.139\linewidth}
\centerline{\includegraphics[width=\textwidth]{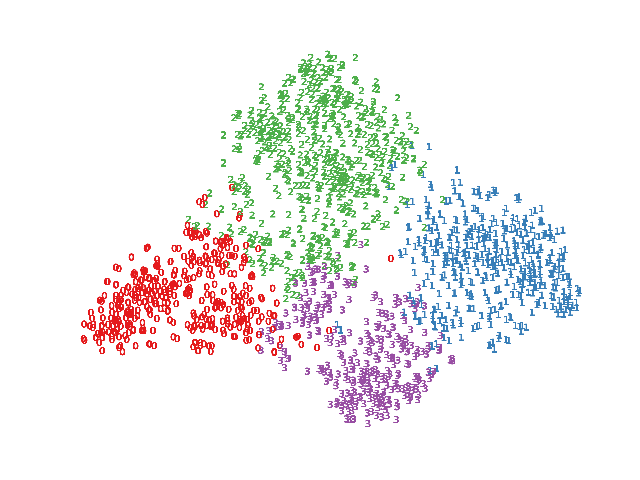}}
\vspace{5pt}
\centerline{\includegraphics[width=\textwidth]{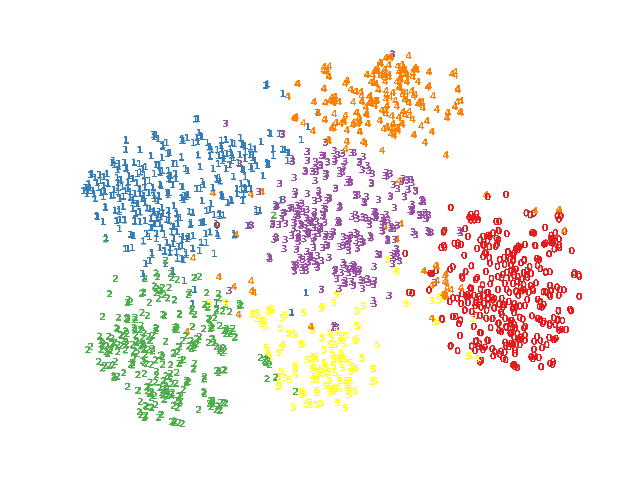}}
\vspace{5pt}
\centerline{\includegraphics[width=\textwidth]{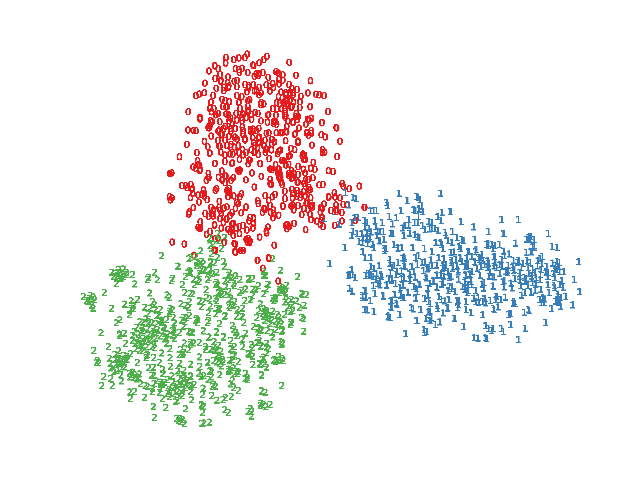}}
\vspace{5pt}
\centerline{\includegraphics[width=\textwidth]{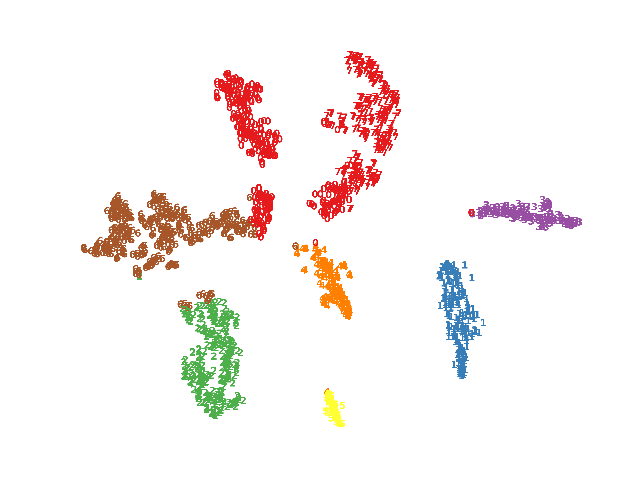}}
\vspace{5pt}
\centerline{(d) 160 Epoch}
\end{minipage}
\begin{minipage}{0.139\linewidth}
\centerline{\includegraphics[width=\textwidth]{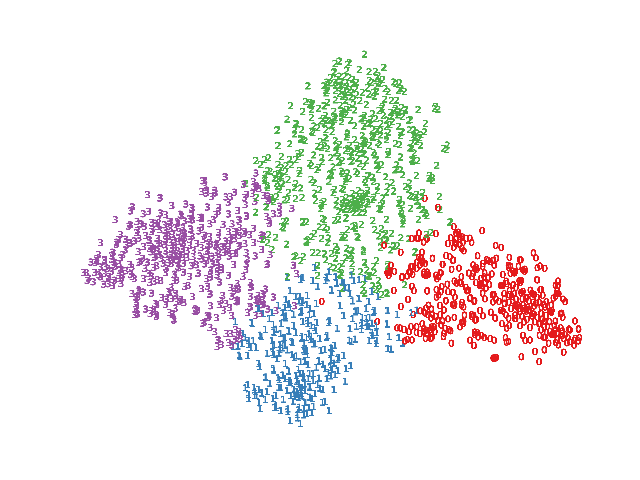}}
\vspace{5pt}
\centerline{\includegraphics[width=\textwidth]{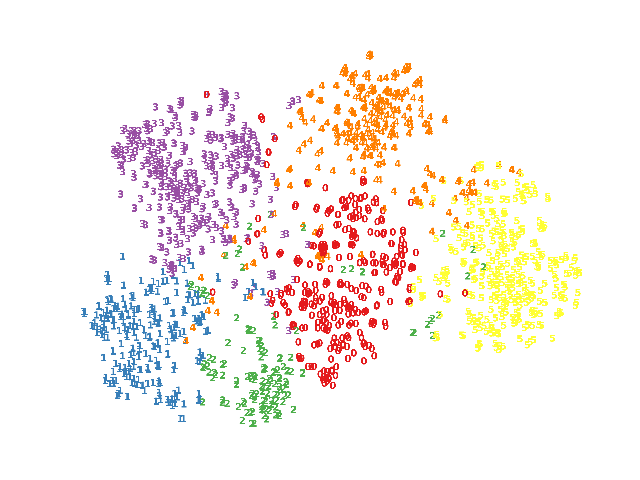}}
\vspace{5pt}
\centerline{\includegraphics[width=\textwidth]{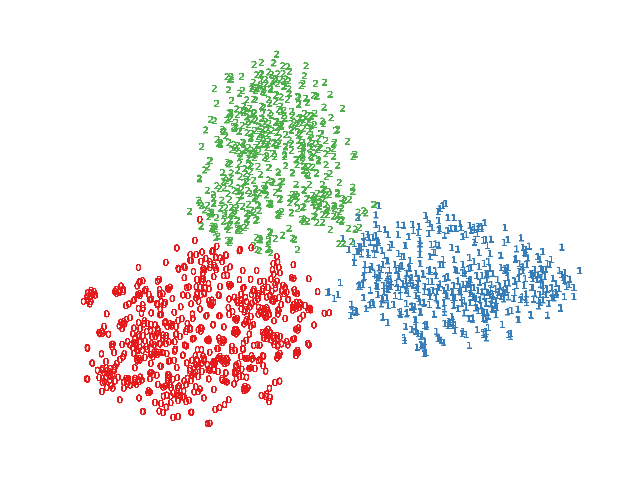}}
\vspace{5pt}
\centerline{\includegraphics[width=\textwidth]{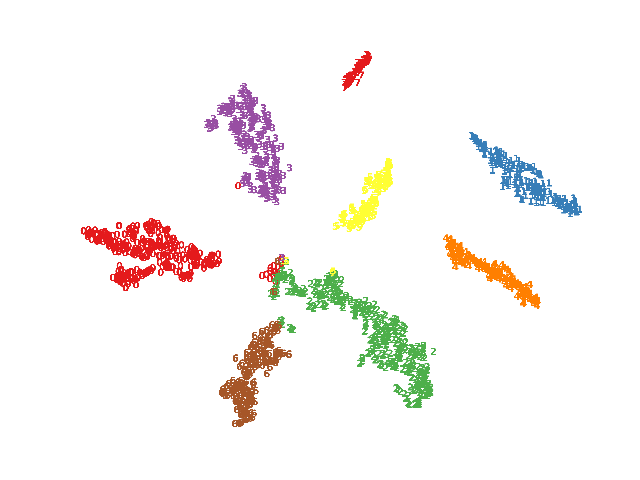}}
\vspace{5pt}
\centerline{(e) 240 Epoch}
\end{minipage}
\begin{minipage}{0.139\linewidth}
\centerline{\includegraphics[width=\textwidth]{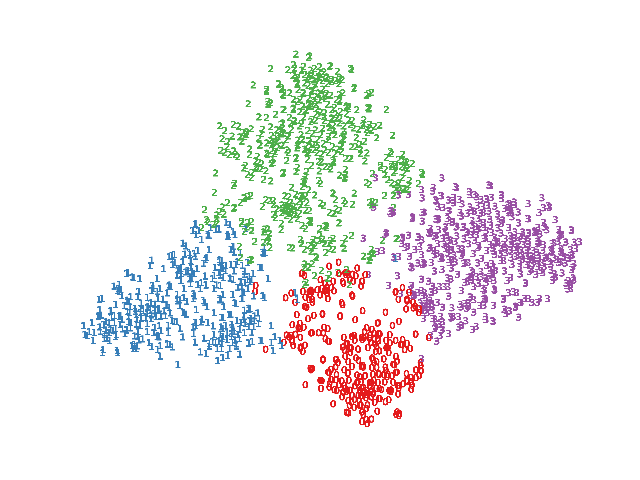}}
\vspace{5pt}
\centerline{\includegraphics[width=\textwidth]{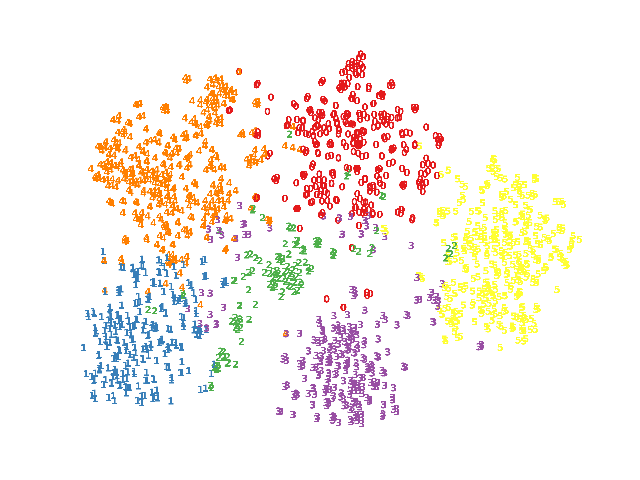}}
\vspace{5pt}
\centerline{\includegraphics[width=\textwidth]{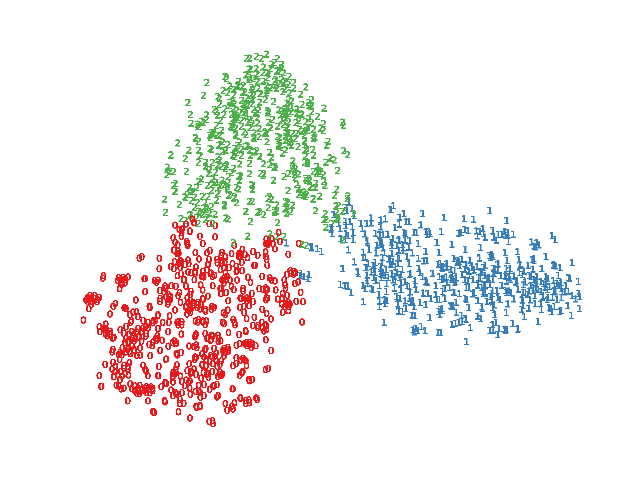}}
\vspace{5pt}
\centerline{\includegraphics[width=\textwidth]{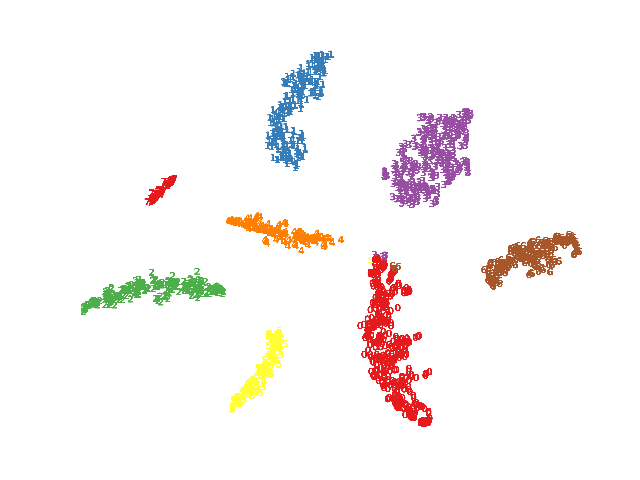}}
\vspace{5pt}
\centerline{(f) 320 Epoch}
\end{minipage}
\begin{minipage}{0.139\linewidth}
\centerline{\includegraphics[width=\textwidth]{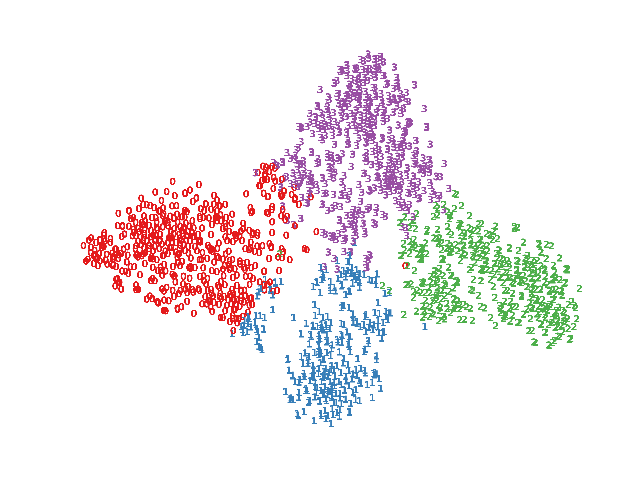}}
\vspace{5pt}
\centerline{\includegraphics[width=\textwidth]{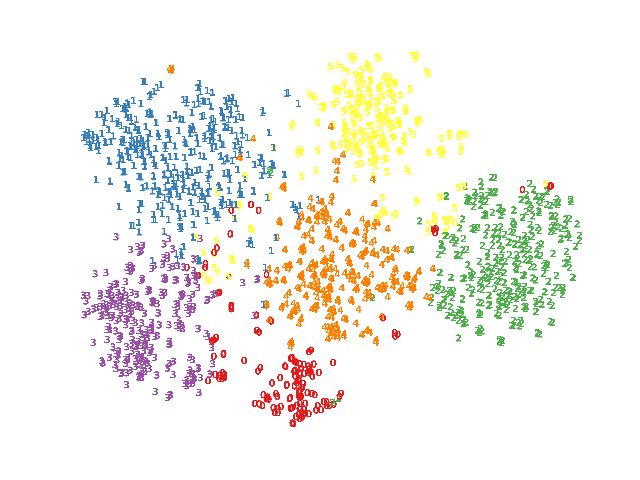}}
\vspace{5pt}
\centerline{\includegraphics[width=\textwidth]{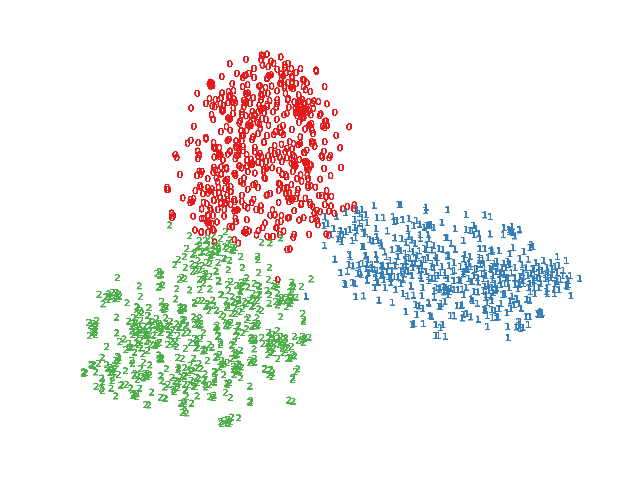}}
\vspace{5pt}
\centerline{\includegraphics[width=\textwidth]{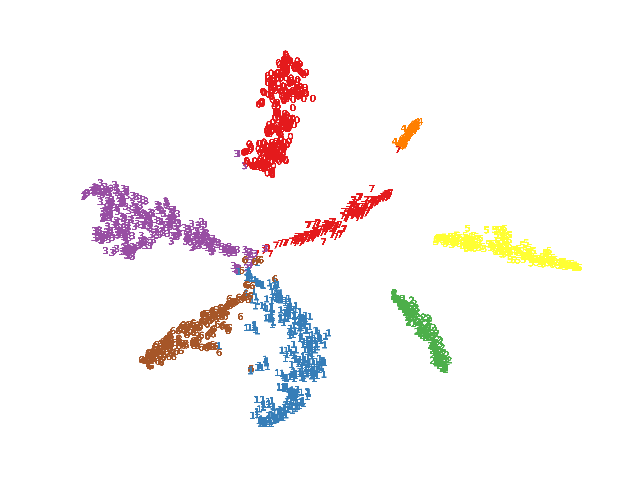}}
\vspace{5pt}
\centerline{(g) 400 Epoch}
\end{minipage}

\caption{$t$-SNE \cite{T_SNE} visualization of the raw data and the training process of our proposed method, including 0 (initialization), 80, 160, 240, 320, 400 epochs. In this figure, the first to the fourth row are the results on the DBLP, CITE, ACM, and AMAP datasets, respectively.}
\label{VIS_tranining}  
\end{figure*}

In addition, we utilize t-SNE algorithm \cite{T_SNE} to visualize the node embeddings $\textbf{Z}$ learned by AE \cite{AE_K_MEANS}, DEC \cite{DEC}, GAE \cite{GAE}, ARGA \cite{ARGA}, DFCN \cite{DFCN} and our proposed IDCRN. The t-SNE algorithm is a non-linear dimensionality reduction algorithm based on t-distribution. From the results as illustrated in Fig. \ref{VIS}, we observe that IDCRN learns a clearer structure of distribution in the latent space, thus better revealing the intrinsic clustering structure among the graph data. Besides, we further show the process of training our proposed method on DBLP, CITE, ACM, and AMAP datasets by performing t-SNE algorithm \cite{T_SNE} over the learned node embeddings $\textbf{Z}$ per 80 training epochs. From these results in Fig. \ref{VIS_tranining}, we observe that the distribute structure of learned node embeddings becomes clearer when the number of training epoch increases.

\section{Conclusion} 
In this paper, to solve the representation collapse problem, we propose a novel deep graph clustering method termed Improved Dual Correlation Reduction Network (IDCRN) by improving the discriminative capability of node embeddings in the sample and feature aspects. Specifically, to the feature aspect, we reduce the redundancy between different dimensions of the learned features by approximating the cross-view feature correlation matrix to an identity matrix, thus improving the discriminative capability of the learned space explicitly. Simultaneously, in the sample aspect, we force the cross-view sample correlation matrix to approximate the designed clustering-refined adjacency matrix. With this setting, we guide the learned latent representation to recover the affinity matrix even across views, thus improving the feature discriminative capability implicitly. Extensive experimental results on six benchmarks demonstrate the effectiveness and efficiency of IDCRN. In the future, it is worth trying to apply IDCRN to more challenging applications, such as incomplete deep graph clustering.

\ifCLASSOPTIONcompsoc
  \section*{Acknowledgments}
\else
  \section*{Acknowledgment}
\fi

This work was supported by the National Key R\&D Program of China (project no. 2020AAA0107100) and the National Natural Science Foundation of China (project no. 62006237, 61922088, 61906020 and 61773392).

\ifCLASSOPTIONcaptionsoff
  \newpage
\fi

\bibliographystyle{IEEEtran}
\bibliography{ref}

\end{document}